\documentclass{article}
\usepackage{blindtext}
\usepackage{subfiles} 
\usepackage{caption}
\usepackage[mathscr]{euscript} 
\usepackage{amsthm}
\usepackage{tocloft}
\usepackage{capt-of}
\usepackage{xpatch}

\usepackage{framed,multirow}

\usepackage{latexsym}
\usepackage{adjustbox}
\usepackage{tabu}
\usepackage{multicol}
\usepackage{soul}
\usepackage{blindtext}
\usepackage{multirow}
\usepackage{xr-hyper}
\makeatletter
\newcommand*{\addFileDependency}[1]{
\typeout{(#1)}
\@addtofilelist{#1}
\IfFileExists{#1}{}{\typeout{No file #1.}}
}\makeatother

\usepackage{xcolor}

\usepackage{PRIMEarxiv}

\usepackage[utf8]{inputenc} 
\usepackage[T1]{fontenc}    
\usepackage[hidelinks]{hyperref}    
\usepackage{url}         
\usepackage{booktabs}    
\usepackage{amsfonts}    
\usepackage{nicefrac}    
\usepackage{microtype}      
\usepackage{lipsum}
\usepackage{fancyhdr}    
\usepackage{graphicx}    
\graphicspath{{media/}}     

\usepackage{float}
\usepackage{amsmath,amssymb}
\usepackage[ruled,vlined]{algorithm2e}
\usepackage{tikz}
\usetikzlibrary{shapes.geometric, arrows}

\usepackage{grffile}

\usepackage{bbm}
\usepackage{xcolor}

\usepackage{subcaption}
\usepackage[]{graphicx} 

\title{Human limits in Machine Learning: Prediction of plant phenotypes using soil microbiome data}

\author{
  Rosa Aghdam\thanks{Joint first authors with equal contribution in alphabetical order} \\
  Wisconsin Institute for Discovery \\
  University of Wisconsin-Madison\\
  Madison, WI\\
   \\
  \And
  Xudong Tang$^*$ \\
  Department of Statistics \\
  Wisconsin Institute for Discovery \\
  University of Wisconsin-Madison\\
  Madison, WI\\
   \\
   \And
   Shan Shan \\
  Department of Plant Pathology \\
  University of Wisconsin-Madison\\
  Madison, WI\\
  \And
  Richard Lankau\\
  Department of Plant Pathology \\
  University of Wisconsin-Madison\\
  Madison, WI\\
   \And
  Claudia Sol\'{i}s-Lemus\thanks{Corresponding author: solislemus@wisc.edu} \\
  Department of Plant Pathology \\
  Wisconsin Institute for Discovery \\
  University of Wisconsin-Madison\\
  Madison, WI\\
}

\newcommand{\listofappendixfiguresname}{\normalsize List of Figures in Supplementary Material}
\newlistof{appendixfigures}{apf}{\listofappendixfiguresname}
\newcommand{\listofappendixtablesname}{\normalsize List of Tables in Supplementary Material}
\newlistof{appendixtables}{apt}{\listofappendixtablesname}

\makeatletter
\xapptocmd{\appendix}{%
  \write\@auxout{%
    \string\let\string\latex@tf@lof\string\tf@lof
    \string\let\string\tf@lof\string\tf@apf%
    \string\let\string\latex@tf@lof\string\tf@lot
    \string\let\string\tf@lot\string\tf@apt%
  }%
}{}{}
\makeatother

\begin{document}
\maketitle

\begin{abstract}
The preservation of soil health is a critical challenge in the 21st century due to its significant impact on agriculture, human health, and biodiversity.  
We provide the first deep investigation of the predictive potential of machine learning models to understand the connections between soil and biological phenotypes. 
We investigate an integrative framework performing accurate machine learning-based prediction of plant phenotypes from biological, chemical, and physical properties of the soil via two models: random forest and Bayesian neural network. 
We show that prediction is improved when incorporating environmental features like soil physicochemical properties and microbial population density into the models, in addition to the microbiome information. Exploring various data preprocessing strategies confirms the significant impact of human decisions on predictive performance. We show that the naive total sum scaling normalization that is commonly used in microbiome research is not the optimal strategy to maximize predictive power. Also, we find that accurately defined labels are more important than normalization, taxonomic level or model characteristics. In cases where humans are unable to classify samples accurately, machine learning model performance is limited. Lastly, we provide domain scientists via a full model selection decision tree to identify the human choices that optimize model prediction power.



\end{abstract}

\section*{Introduction}


It is undeniable that machine learning (ML) has revolutionized the manner in which scientific research is performed in recent years. Among the many tasks performed by ML models in our daily lives, researchers have relied on ML to assist in clinical diagnoses \cite{Khodabakhsh2023}, identification of bacterial phenotypes such as antimicrobial resistance \cite{Anahtar2021}, and even identification of objects in space \cite{Lim2021}. Recently, the vast evidence of the prediction power of ML models on a wide range of applications has launched the adoption of these models in other domains such as sustainable agriculture where soil health -- characterized by a wide range of biological, chemical, and physical properties \cite{wagg2019fungal} -- is explored as an important driver to predict plant phenotypes, such as disease susceptibility or yield.

The question of whether the improvement of soil health could result in superior crop yield and disease resistance remains open, as researchers have not been able to identify a set of indicators to accurately and robustly predict plant outcomes from soil information. Among all the candidate indicators for soil health, soil microbiome is one that continues to be understudied in its predictive potential in the productivity and resilience of agricultural ecosystems \cite{Trivedi2016Indicators}.
Aided by amplicon sequencing of highly preserved phylogenetically informative marker genes, like the 16S ribosomal RNA for bacteria and the internal transcribed spacer (ITS) for fungi, 
there have been extensive studies on the complexity and diversity of soil microbial communities over the last decade, yet little is still understood on how changes in the microbial community in soil would directly impact on plant growth and health.
While it is already well-recognized that soil microbiome is closely related to plant health and productivity \cite{Berendsen2012microbiome}, most current research efforts on the soil microbiome are limited to the change on microbial community in response to agricultural management. 


 Fortunately, the rapid development of ML models could fill this gap. Over the years, ML models have been used on high-dimensional, highly interconnected data for prediction and decision-making that does not rely on \textit{a priori} knowledge on the effects and interactions among covariates. Since we have little or no knowledge of most species among the thousands contained in the soil microbial communities, it would be beneficial to exploit the power of ML models on microbiome data as they are able to explore the unknown interactions between the microbial communities and plant phenotypes. 

There are three main challenges, however, for analyzing microbiome datasets with ML methodologies: (1) Data are inherently compositional, which means that the raw counts are normalized based on the total number of reads that were collected from the sample. Therefore, microbial abundances are not independent, and the use of traditional statistical techniques (such as correlation) might lead to increased false discovery rates \cite{guseva2022diversity}. (2) Data are highly sparse, which means that datasets include a large number of operational taxonomic units (OTUs) that are present in a small proportion of samples (or in none at all) \cite{kurtz2015sparse}. (3) Data are high-dimensional, which means that the number of OTUs are larger than the number of samples, especially in more specific taxonomic orders like Order, Family, and Genus. 

In addition to challenges on the microbiome data, prediction power also deteriorates on settings involving inaccurate and imbalanced labels. That is, when we consider the task of prediction, each microbial sample should be labeled by a plant outcome value (say, high yield or low yield). It is common that labels in these datasets are imbalanced with one class representing the majority of observed samples which is denoted as an imbalanced binary classification problem.
Furthermore, decisions on the class labels are many times not straight-forward. While diseased plant or non-diseased plant tends to be an indisputable classification, determining what constitutes high versus low yield is up for debate and depends not only on the season and other environmental factors, but also on the specific crop variety.

\begin{figure}[ht!]
\centering
\includegraphics[scale=0.2]{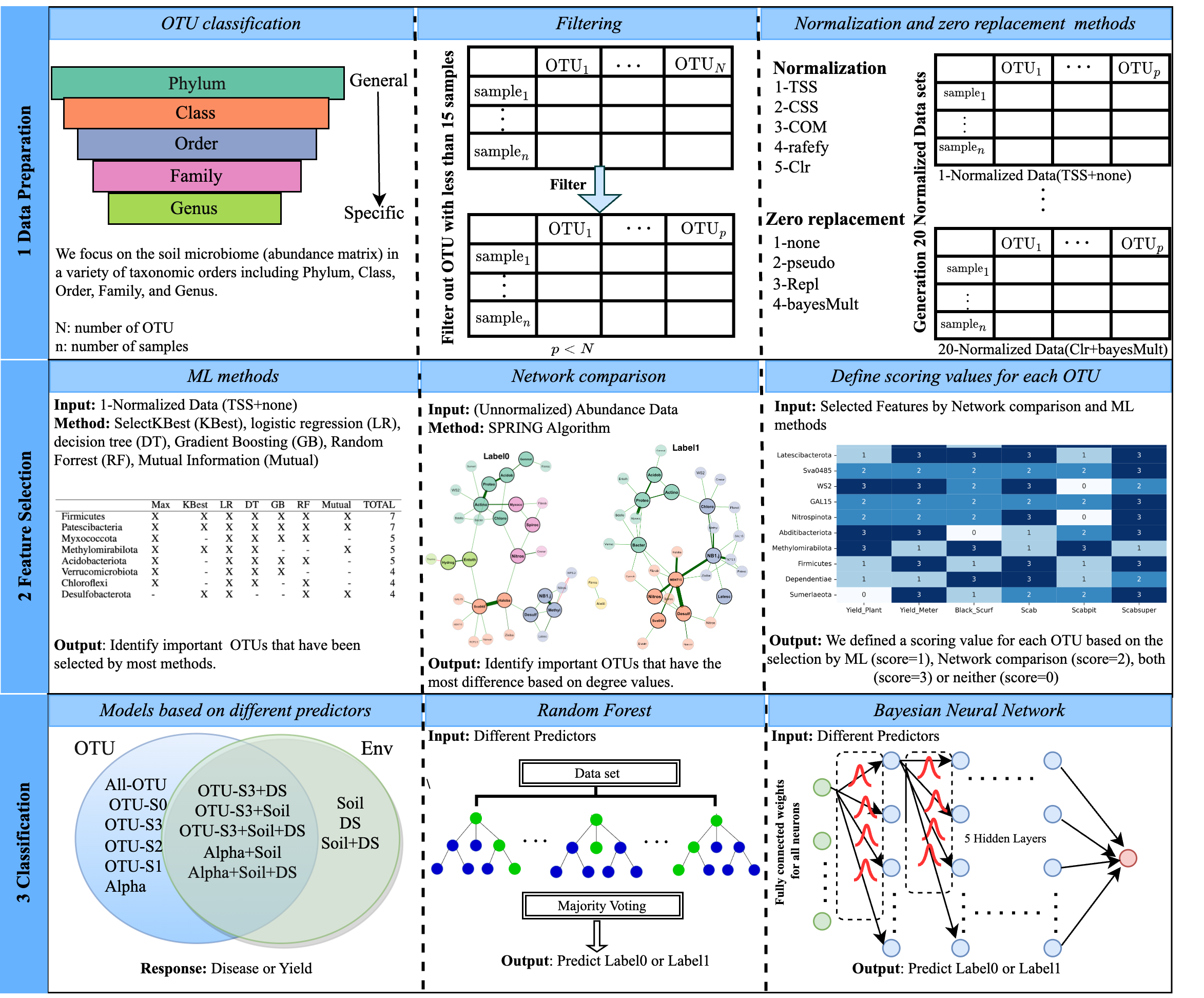}
\caption{\small{Workflow of the analyses with three main steps: (1) Data Preparation, (2) Feature Selection, and (3) Classification. In (1) Data Preparation, we consider OTUs (number of OTUs = N) at different taxonomic levels and filter by sample size (n). In addition, we perform conbination of five normalization methods and four zero replacement methods (for a total of 20 normalized datasets). In (2) Feature selection, we rank OTUs based on i) the number of times they are selected as important features by machine learning (ML) criteria, and ii) the greatest degree of difference on microbial networks reconstructed from samples of each class. We score OTUs based on whether they are selected by ML ($score=1$), by network comparison ($score=2$), both ($score=3$) or neither ($score=0$). In (3) Classification, the Venn diagram depicts the different types of predictors: microbiome (OTUs), environmental (Env), and the combination of both. The acronyms (e.g., All-OTU or OTU-S3+DS) correspond to different choices of predictors that are described in Table \ref{tableNameMethods}).
Random forest and Bayesian neural network classification models are fitted on the different input predictors.}}
\label{3-Abstract}
\end{figure}

Here, we explore the potential of ML models in the prediction of plant phenotypes of yield and disease from soil data. We utilize a real dataset from potato fields in Wisconsin and Minnesota and focus on the performance of the models while facing data challenges related to binarization, imbalance, compositionality, sparsity and high dimensionality. Furthermore, we test the impact of specific data preprocessing steps such as 
1) different normalization and zero replacement strategies to overcome the compositionality and sparsity, 2) different feature selection strategies to overcome the high dimensionality, and 3) data augmentation techniques to overcome the imbalance of the data. In addition, we are also interested in answering the question of whether soil microbiome data alone has enough predictive power to predict the disease and yield phenotypes or whether other information on the soil, such as chemical composition, is necessary for accurate prediction. The answer to this question will inform farmers on the best data collection strategies given that soil microbiome data is expensive. That is, if the soil microbiome data does not provide enough predictive power compared to other (cheaper) soil measurements, then the collection of microbiome data for prediction purposes would be futile. Figure \ref{3-Abstract} shows a graphical description of our pipeline.

For our investigation, we have chosen two distinct machine learning models. First, we employ Random Forests (RF), which have consistently demonstrated superior predictive performance across various domains \cite{belgiu2016random}. Further details about this method can be found in Section `Random Forest Model'. Second, we utilize a Bayesian Neural Network (Bayesian NN), known for its inherent protection against overfitting \cite{soumya2019bnn} (See Section `Bayesian Neural Network Model'). Given their capacity to capture complex interactions among features, neural networks are valuable, especially when dealing with unknown relationships. However, considering the small sample size and large feature space in our study, traditional back-propagation-based neural network models may exhibit a substantial bias or overfitting \cite{Hernandez2015BP}, hence the need to explore the Bayesian version.

Among the main findings, we can highlight that microbiome data alone indeed has predictive power for disease outcomes, especially for pitted scab disease, but not to predict yield.
We also find that the most powerful prediction is achieved by combination of environmental information and microbiome data. 
Among the data preprocessing strategies that we explored, we find that normalization and zero replacement strategies have a huge impact on the prediction power of the models, yet there are strong interaction effects with taxonomic levels, and thus, it is impossible to identify one strategy that outperforms the others.
In terms of the model, we found no clear differences between the RF and the Bayesian NN, yet the latter is more computationally expensive and prohibited on some datasets with a large number of predictors, which prompts us to prefer RF for most settings.
We conclude our investigation with a full model selection (FMS) decision tree \cite{sun2012full, sun2013towards} that allows us to identify combinations of normalization, zero replacement, feature selection, and model that yield the highest prediction accuracy and that can serve to provide specific strategies for other researchers with similar datasets.

\section*{Results}
\subsection*{Overall performance of predictive models: Manual binarization causes inaccurate prediction of yield}
\label{sec:boxplots}

\begin{figure}[htbp]
    \centering
    \begin{subfigure}[b]{\textwidth}
        \centering
        \includegraphics[width=\textwidth]{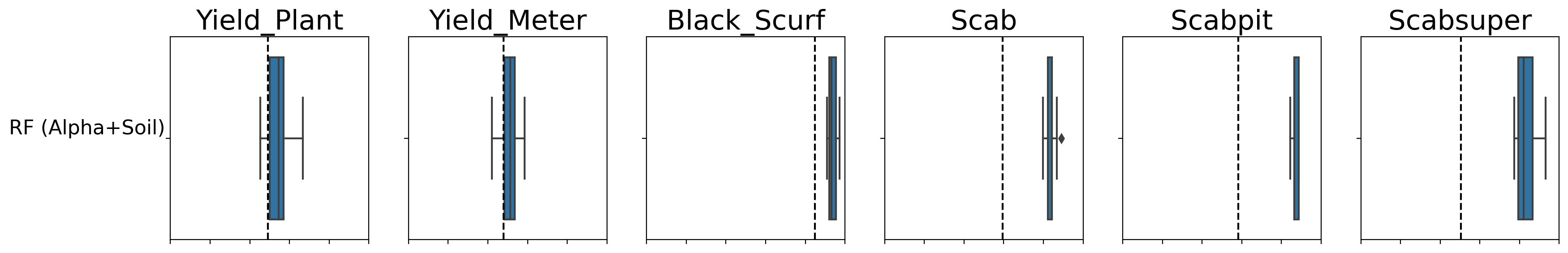}
    \end{subfigure}
    
    \begin{subfigure}[b]{\textwidth}
        \centering
        \includegraphics[width=\textwidth]{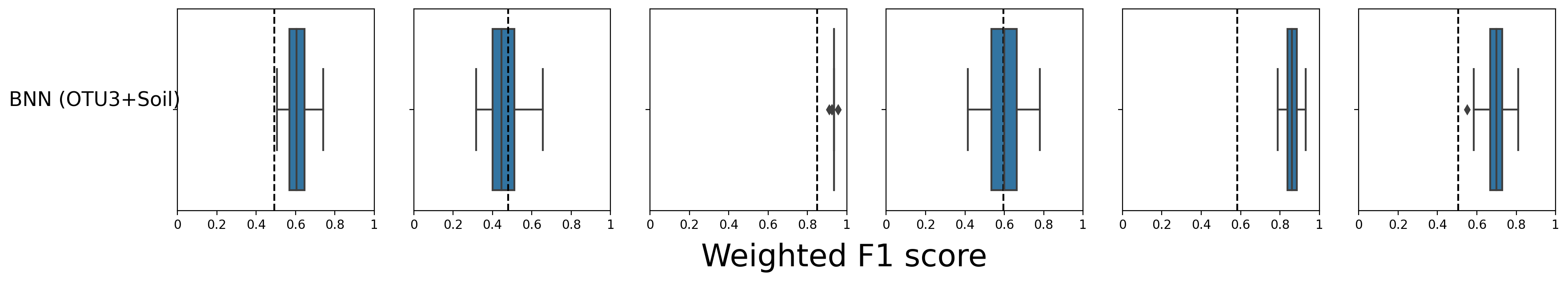}
    \end{subfigure} 
    \caption{The most accurate predictions across all outcomes are achieved using alpha diversity and soil chemistry data (Alpha+Soil) for the RF model, whereas for the Bayesian NN models, optimal performance is observed when utilizing OTUs identified as important by both machine learning and network comparison strategies, in conjunction with soil chemistry data (OTU-S3+Soil). For a detailed presentation of all results, please refer to Supplementary Figures S\ref{4-ALL-RF} (RF) and S\ref{5-ALL-BNN} (Bayesian NN).}
    \label{bestRFBN}
\end{figure}

First, we identify the outcomes (disease or yield) that are accurately predicted across models (and are thus robust to prediction regardless of model choices), as well as the models that accurately predict across outcomes (and are thus the most powerful model alternatives).
To do so, we aggregate the weighted F1 scores on data preprocessing choices such as normalization, zero replacement, and taxonomic levels for every model and every outcome.
We employed RF and Bayesian NN models across various predictive scenarios. Figure \ref{bestRFBN} showcases the best performance of RF and Bayesian NN models. In the RF model, the integration of alpha diversity and soil chemistry data (referred to as Alpha+Soil) yields the most accurate predictions across all outcomes. Conversely, optimal performance for the Bayesian NN models is achieved by OTUs identified as significant by both ML and network comparison strategies (more details on feature selection are provided in Section `Effective preservation of predictive signal with different feature selection strategies'), alongside soil chemistry data (denoted as OTU-S3+Soil). These findings highlight the importance of integrating environmental information with microbiome data to enhance predictive power for disease outcomes. Further insights into the predictive capabilities of these models are presented in Figure S\ref{4-ALL-RF} (RF) and Figure S\ref{5-ALL-BNN} (Bayesian NN), where a comprehensive analysis of all 14 models described in Table \ref{tableNameMethods} is provided, offering valuable guidance for model selection and interpretation.
Columns correspond to the six responses: four diseases and two yield outcomes. For a given panel (model in row and response in column), the boxplot corresponds to the different weighted F1 scores for every combination of normalizations/zero replacement strategies as well as different taxonomic levels (Table \ref{tablenumpredictor}). For example, the boxplots for the ALL-OTU model (first row) include weighted F1 scores of the model fit on 20 normalization/zero replacement strategies, and 5 taxonomic levels (100 different weighted F1 scores per outcome). The dashed line in each panel corresponds to the average weighted F1 score of the model when fit with all random datasets (see Section `Prediction of potato disease from microbiome data' and Figures S\ref{7-CompareToRandom} and S\ref{S-CompareToRandom} in Supplementary file). This line allows us to assess whether the real data has more predictive power than random data  \cite{aghdam2015cn}.

While these plots do not allow us to distinguish differences by taxonomic level or normalization/zero replacement strategy (more on that in the next subsections), we can identify outcomes (columns) that can be more accurately predicted across models (rows). Additionally, we can identify models that are capable of accurately predicting more outcomes.
It is readily evident, for example, that the yield outcomes cannot be accurately predicted by any model as all the weighted F1 scores fall consistently below the dashed line. It is notable that the yield responses are those for which there is not a clear binarization strategy. Since we are artificially separating samples into the two classes (low and high yield) based on whether they are above or below the variety-specific median, samples on the boundary will in fact be very similar to each other, and thus, difficult to classify. Furthermore, the poor prediction of yield is not restricted to one data type (microbiome vs environmental) which also suggests that the prediction challenges arise from the binarization process rather than the model or set of predictors.

Disease outcomes, on the contrary, display higher weighted F1 scores overall, and in particular, pitted scab displays weighted F1 scores that are consistently above the random prediction dashed line across different models.
For the case of black scurf, even when the weighted F1 scores are very high, this is a deceiving result, as this disease outcome is highly imbalanced. This means that a naive model predicting all samples to belong to the majority class will have high prediction accuracy (see dashed line above 0.8 for random data). We investigate the prediction of black scurf more carefully with the augmented data that balances the proportion of both classes (Section `Robust prediction in imbalanced datasets with data augmentation'). 
Finally, given the poor performance on yield, we focus on the fine-grained description of results for the disease outcomes only for the remaining of the manuscript.


\subsection*{Prediction of potato disease from microbiome data}

\subsubsection*{No normalization or zero replacement strategy proven superior in terms of predictive power}

One of the goals of our study is to identify ideal data preprocessing steps that are guaranteed to maximize predictive power on the ML models. Our results, however, suggest that there is not a clear pattern of superiority among the normalization or zero replacement strategies, and that the selection of the data preprocessing steps needs to be data specific (and taxonomic level specific).

\begin{figure}[ht!]
\centering
\includegraphics[scale=0.35]{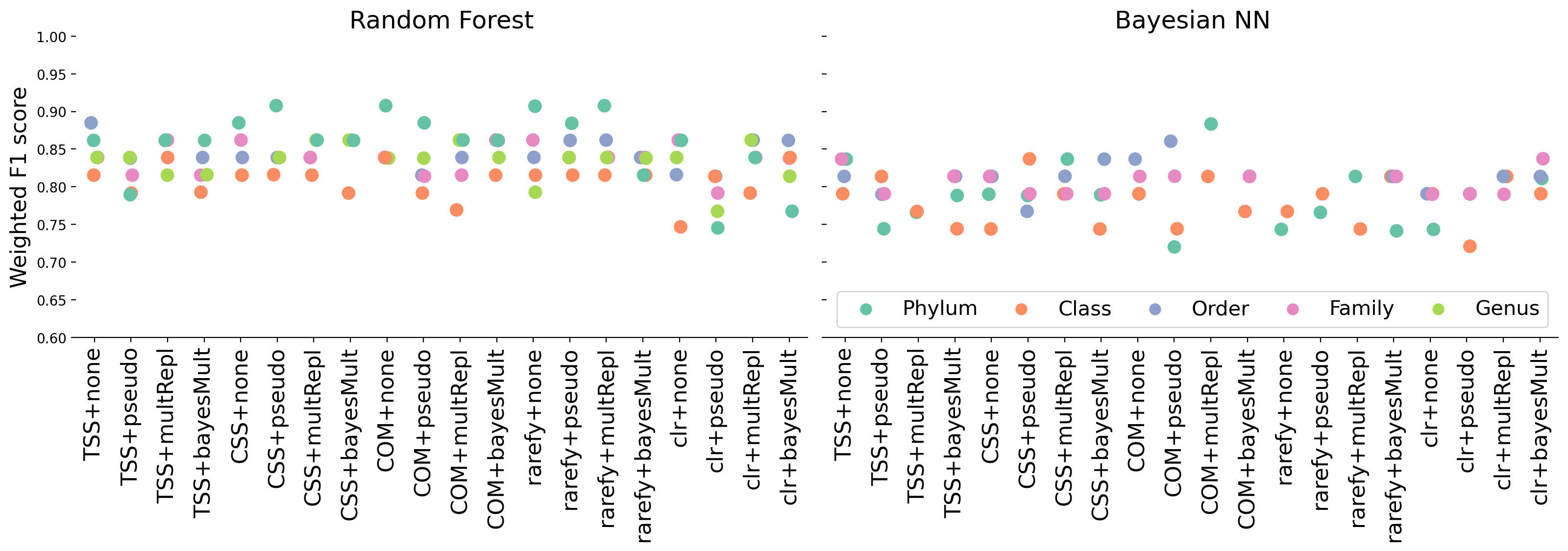}
\caption{\small{Weighted F1 scores (y-axis) for random forest and Bayesian neural network (Bayesian NN) models for the pitted scab disease under the 20 normalization/zero replacement strategies (x-axis). The lack of pattern prevents us from making recommendations of optimal strategies for microbiome data. We can conclude, however, that taxonomic levels, normalization and zero replacement strategies have an effect on the prediction accuracy of the models as evidenced by the broad range displayed by the points.}}
\label{10-RF-BNN-Normalized-Scabpit}
\end{figure}

Figure \ref{10-RF-BNN-Normalized-Scabpit} shows the weighted F1 scores for pitted scab for different combinations of normalization and zero replacement strategies (x-axis) for the two types of models (RF and Bayesian NN). These analyses include all OTUs under the five taxonomic levels (different colors). Similar plots for other diseases are presented in Figures S\ref{S-RF-BNN-Normalized-Yield-Plant} to S\ref{S-RF-BNN-Normalized-Scabsuper} in the Supplementary Material.
As mentioned, we note that there is not an apparent pattern that points at any normalization or zero replacement strategy being superior to other alternatives across all taxonomic levels. In fact, there is considerable interaction between the normalization/zero replacement method and the taxonomic level. For example, for the RF model, the best result is achieved with Phylum level and cumulative sum scaling normalization with pseudo-zero replacement strategy (CSS+pseudo)  or common sum scaling normalization without any zero replacement strategy (COM+none). Additionally, the rarefy+none and rarefy+multRepl strategies demonstrate good performance.
For Bayesian NN, however, the best results are achieved with the common sum scaling normalization with the multiplicative zero replacement (COM+multRepl) for the Phylum level.

For a given normalization/zero replacement strategy (x-axis), the variability in the scatterplot points indicates that taxonomic levels have an impact on the predictive power of the model. However, there is no discernible pattern indicating that a specific taxonomic level consistently provides more predictive signal. While the Phylum level appears to yield higher weighted F1 scores for certain normalization/zero replacement strategies, this trend is not observed across all strategies.

When we compare the range of weighted F1 scores across normalization and zero replacement strategies, we see that the effect of the strategy is not negligible. For example, for a Phylum level, the lowest weighted F1 score is around 0.75 for centered log-ratio normalization with pseudo-zero replacement strategy (clr+pseudo) to around 0.9 for cumulative sum scaling normalization with pseudo-zero replacement strategy (CSS+pseudo). This implies that for a given taxonomic level, the resulted weighted F1 score will be highly influenced by the normalization and zero replacement strategy. Traditionally, microbiome researchers use the total sum scaling normalization without any zero replacement strategy (TSS+none) on their data which has a range of 0.80 to 0.90 weighted F1 scores for the RF model (0.8 to 0.85 for the Bayesian NN model) depending on the taxonomic level.

The strong interaction effects of taxonomic level, normalization and zero replacement strategy prevent us from making recommendations about the best data preprocessing practices that can be generalizable to other datasets. We conclude by suggesting data practitioners to try different normalization and zero replacement strategies rather than just one, but see Section `Identifying general practices to predict potato disease from microbiome data using full model selection models' for more recommendations.

\subsubsection*{Effective preservation of predictive signal with different feature selection strategies}
\label{sec-feature}

\begin{figure}[ht!]
\centering
\includegraphics[scale=0.35]{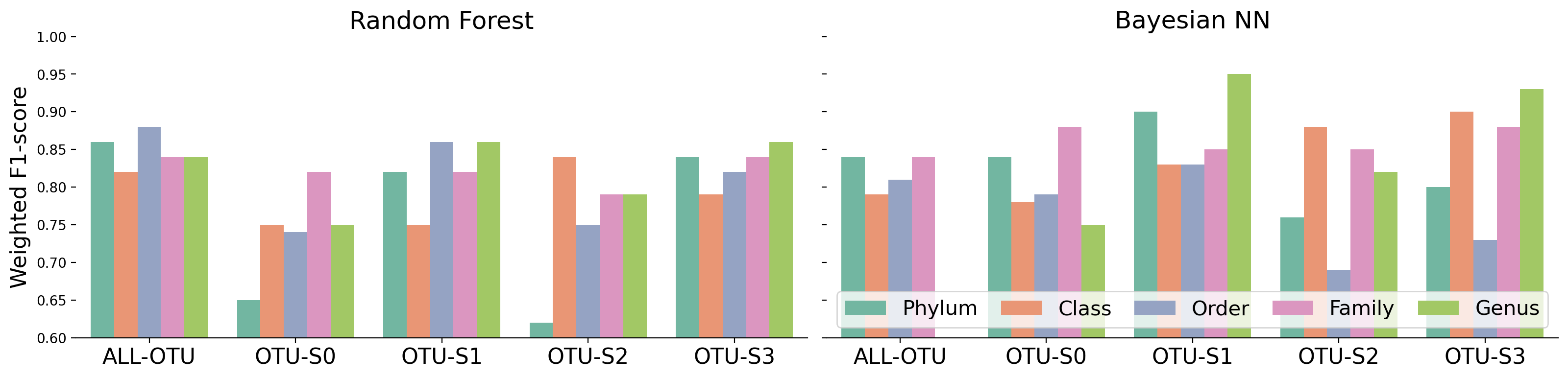}
\caption{Weighted F1 scores (y-axis) by random forest and Bayesian neural network (Bayesian NN) models for the pitted scab disease (Scabpit) by feature selection strategy (x-axis) including all OTUs (All-OTU), OTUs selected by the ML method (OTU-S1), the network comparison method (OTU-S2), both methods (OTU-S3), or neither method (OTU-S0).}
\label{11-RF-BNN-selectedfeatures}
\end{figure}

One of the standard steps in the ML pipeline is feature selection, especially for cases of high-dimensional data. We compare the ability to retain predictive signal of three feature selection strategies: standard importance score from ML methods, comparison of microbial network topologies and combination of both. More details on the feature selection strategies can be found in Methods.
Figure \ref{11-RF-BNN-selectedfeatures} shows the weighted F1 scores for the two types of models (RF and Bayesian NN) on pitted scab disease (Scapbit) under different subsets of predictors: (1) all OTUs (ALL-OTU), (2) only OTUs that were identified as important by the ML strategy (OTU-S1), (3) only OTUs that were identified as important by the network comparison strategy (OTU-S2), (4) OTUs that were identified as important by both strategies (OTU-S3), or (5) OTUs that were not identified as important by neither strategy (OTU-S0). For fair comparison, we include the same number of predictors in OTU-S0 as in OTU-S3. Similar figures for other responses are shown in Figures S\ref{S-RF-BNN-SelectedFeatures-Yield-Plant} to S\ref{S-RF-BNN-SelectedFeatures-Scabsuper} in the Supplementary Material.

Again, we perceive strong interaction between taxonomic level and feature selection strategy.
For the RF model, the highest weighted F1 score is achieved when including all OTUs (ALL-OTU) at Order level whereas for the Bayesian NN model, the highest weighted F1 score is achieved when including OTUs identified by the ML strategy (OTU-S1) at Genus level. RF models on all OTUs (ALL-OTU) have weighted F1 score above 0.8 in all taxonomic levels which suggest that this model could be a better alternative compared to Bayesian NN which is more computationally intensive. There are also smaller differences in RF models when comparing the performance on OTU-S3 (important OTUs) and ALL-OTU (all OTUs) which suggests that the feature selection strategy is sufficient to preserve the predictive signal in the data while reducing the number of predictors in the model. This is relevant for computationally intensive models such as Bayesian NN that do not allow the inclusion of all OTUs for certain taxonomic levels.



\subsubsection*{Robust prediction in imbalanced datasets with data augmentation}
\label{sec-data-aug}

High prediction power in imbalanced datasets is misleading as a naive predictor that classifies all samples as the majority class will have high accuracy. In our data, black scurf disease is highly imbalanced, and thus, the high prediction accuracy is unreliable. We confirmed, however, that after data augmentation which balanced the data, accurate prediction persisted.

\begin{figure}[ht!]
\centering
\includegraphics[scale=0.3]{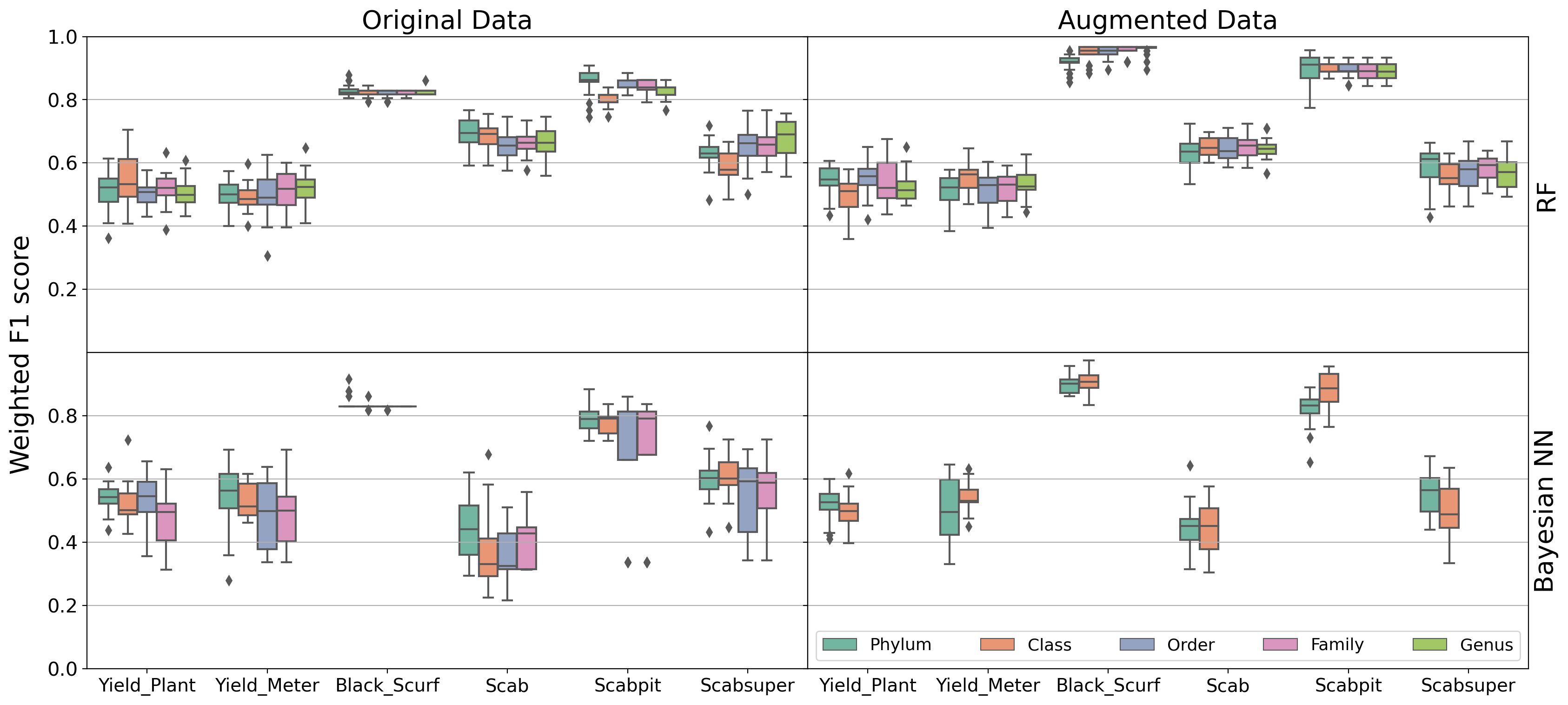}
\caption{\small{Boxplots with weighted F1 scores by model (rows): random forest  (RF) and Bayesian neural network (Bayesian NN) for original and augmented OTU predictors (columns). The x-axis represents the yield and disease outcomes. The range of each box plot depicts the weighted F1 scores for 20 normalized  datasets at each taxonomic level.}}
\label{6-Original-Aug-RF-BNN}
\end{figure}

To illustrate this, Figure \ref{6-Original-Aug-RF-BNN} depicts the weighted F1 scores on original and augmented datasets for all yield and disease outcomes (x-axis) and both models (RF and Bayesian NN). 
The range of each box plot depicts the weighted F1 scores for 20 normalized datasets at each taxonomic level. 
We observe that black scurf and pitted scab can be reliably predicted across taxonomic levels as their median weighted F1 scores for all taxonomic orders is around $0.8$ when the models are fitted on the original datasets. As mentioned before, however, black scurf is highly imbalanced, so the results on the original data are not reliable. Fortunately, the median weighted F1 scores on augmented data (which is perfectly balanced by design) increases for both diseases, such that they are around $0.9$ for all taxonomic levels. These results suggest that data augmentation, especially in cases of highly imbalanced data, is an appropriate strategy that improve the robustness of the model, in some cases even increase the accuracy. One has to be careful, however, in that augmented data can yield certain models prohibited. For example, the Bayesian NN model could not be fit on the augmented datasets for Order, Family nor Genus levels due to computational limitations.


\subsubsection*{Identifying general practices to predict potato disease from microbiome data using full model selection models}
\label{dectree}

As evidenced by our analyses, every single data and model choice has an impact on the predictive performance of our methods. The effects of different data preprocessing steps appear to strongly interact, and thus, we could not identify clear patterns on strategies to maximize prediction power.

\begin{figure}[ht!]
\centering
\includegraphics[scale=0.17]{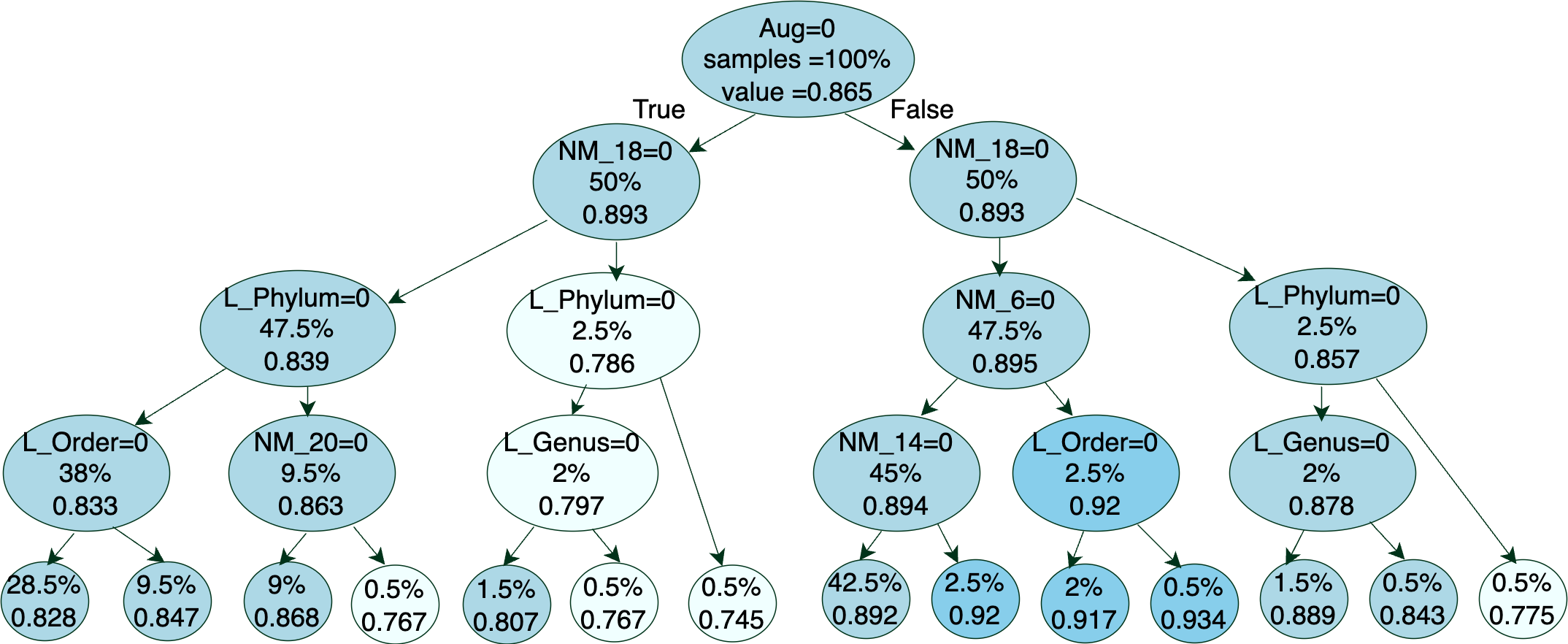}
\caption{\small{Full model selection decision tree with a maximum depth of 4 summarizing the results of random forest models on pitted scab disease. When the condition at a node is true, we follow the branch on the left, and when the condition is false, we follow the branch on the right. The percentage of the data preprocessing options and mean of weighted F1 scores are shown in each node. 
} }
\label{12-DT-RF-Scabpit}
\end{figure}

With a FMS \cite{sun2012full, sun2013towards, reyes2021full} strategy, however, we are able to 
identify the choices that yield the highest measure of performance. More details on the FMS models can be found in Methods.
Figures \ref{12-DT-RF-Scabpit} and S\ref{13-DT-BNN-Scabpit} show the FMS decision trees for the RF and Bayesian NN models on pitted scab disease, respectively. A FMS decision tree shows the different data preprocessing steps that yield different weighted F1 scores, so that practitioners can select the options that result in the highest predictive power. Here, we have five taxonomic levels,  20 normalization+zero replacement strategies, and 2 data augmentation options: no data augmentation (Aug=0) and data augmentation (Aug=1). Thus, in total, we have 200 data preprocessing options (20 normalization strategies times 5 taxonomic levels times 2 data augmentation). 

To interpret a FMS decision tree, each node corresponds to a specific step in the data preprocessing pipeline, for example, whether to perform data augmentation or not. If the condition is true, we follow the branch to the left; if the condition is false, we follow the branch to the right. At the top of the decision tree, we have the root which represents the data preprocessing step that has the greatest effect on model accuracy. At the bottom of the decision tree, we have the leaves with the average weighted F1 score of the model fitted on the data that satisfy all conditions towards the root. 
Each node also displays the percentage of data preprocessing options included in the node. 
For example, in Figure \ref{12-DT-RF-Scabpit}, the root node covers 100\% of the options with average weighted F1 scores 0.865. The condition at the root node ($\text{Aug}=0$) represents the case of "no data augmentation". Thus, "true" (left of the root) means "no data augmentation", and "false" (right of the root) means "data augmentation".
For simplicity, we denote the 20 normalization/zero replacement strategies as $\text{NM}_i$ for $i=1,\dots,20$. See Section `Data Filtering, Normalization, and Zero Replacement' for a description on each normalization/zero replacement strategy.

For the FMS decision tree for the RF model (Figure \ref{12-DT-RF-Scabpit}), the highest weighted F1 score (0.934 with 0.5\% of the data) is achieved with data augmentation, normalization/zero replacement strategy \#6 (CSS+pseudo), and Order level.
Another path of the decision tree follows data augmentation and any normalization/zero replacement strategy except \#6 (CSS+pseudo), \#14 (rarefy+pseudo), and \#18 (clr+pseudo) which yields an average weighted F1 scores of 0.892 for 42.5\% of the data preprocessing options.
If data augmentation is not an option (left of the root), the highest weighted F1 score available is 0.868 with Phylum level, and any normalization/zero replacement strategy except \#18 (clr+pseudo) or \#20 (clr+bayesMult). For the FMS decision trees on the other responses, see Figures S\ref{S-DT-RF-Yield-Plant} to S\ref{S-DT-RF-Scabsuper} in the Supplementary Material.

Similarly, in Figure S\ref{13-DT-BNN-Scabpit} for the Bayesian NN model, the highest weighted F1 score (0.896 with 15\% of the data preprocessing options) is achieved when we do data augmentation, we use any taxonomic level except Phylum, and we use any normalization/zero replacement strategy except \#10 (COM+pseudo) and \#18 (clr+pseudo). See Figures S\ref{S-DT-BNN-Yield-Plant} to S\ref{S-DT-BNN-Scabsuper} in the Supplementary Material for other responses.

While the specific recommendations on normalization, zero replacement and taxonomic level are model-specific, both models perform better with data augmentation. In terms of taxonomic level, we note that the Bayesian NN was only run on Phylum and Class, and thus, the highest accuracy is obtained with Class level (when Phylum$=0$ is true). This does not contradict the result from the RF that identified Order level as the one yielding higher accuracy. We cannot rule out that the Bayesian NN would also have higher accuracy with Order compared to Class. The results from the RF, though, seem to suggest that there is a peak at Order, and more granularity in Family and Genus does not seem to provide more predictive power. 

\subsection*{Prediction of Potato Disease from Environmental Data}

One of the questions to address in our work is whether prediction accuracy is improved by the inclusion of microbiome data, or if environmental factors (usually cheaper to collect) provide enough signal to classify potatoes in diseased or non-diseased groups.
We found that environmental factors indeed provide sufficient signals to predict pitted scab disease as illustrated in 
Figure S\ref{14-RF-BNN-Env} which shows the weighted F1 scores by RF and Bayesian NN models based on environmental (soil characteristics) data for pitted scab. The range of each boxplot corresponds to the six scaling methods described in Section `Data Filtering, Normalization, and Zero Replacement'. In contrast with the normalization methods in microbiome data, we observe here that the scaling methods do not seem to have an effect on prediction as evidenced by narrow boxplots, and that weighted F1 scores are all higher than 0.75, and therefore, comparable to the models fitted on microbiome data alone. These results suggest that environmental factors alone are powerful to predict the incidence of pitted scab in the tubers. As microbiome data is more expensive than environmental data, we suggest to prefer environmental predictors under restricted monetary budget.
See Figures S\ref{S-RF-BNN-Env-Yield-Plant} to S\ref{S-RF-BNN-Env-Scabsuper} in the Supplementary Material for other responses.

\subsection*{Leveraging Microbiome and Environmental Data in the Prediction of Potato Disease}

\begin{figure}[ht!]
\centering
\includegraphics[scale=0.35]{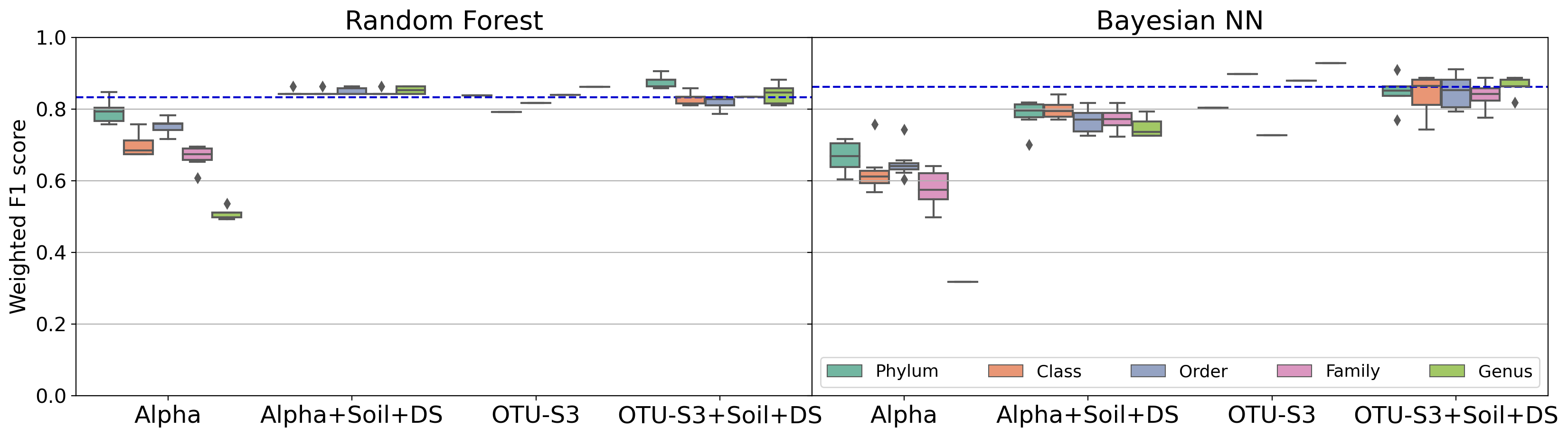}
\caption{\small{Boxplots with the weighted F1 scores (y-axis) by random forest  and Bayesian neural network (Bayesian NN) models for pitted scab disease. The range of each box plot depicts the weighted F1 scores for normalized datasets at each taxonomic level. There are 6 normalization methods for alpha diversity and environmental factors (Soil and DS), and one normalization method for microbiome (OTU-S3). See Table \ref{tableNameMethods} and \ref{tablenumpredictor} for description of models and number of normalization methods for each predictors. The models including both types of predictors outperform other models, yet models including microbiome data alone (OTU-S3) are comparable which suggests that the microbial information indeed contains signal to predict the disease outcome on its own. However, microbiome data is more expensive to collect, and perhaps not necessary, given that the model without microbiome data (Soil) performs just as accurately (blue dashed line).}}
\label{15-RF-BNN-Com}
\end{figure}

As expected, prediction accuracy is improved by the inclusion of both datatypes: microbiome and environmental.
Figure \ref{15-RF-BNN-Com} shows the weighted F1 scores by RF and Bayesian NN models based on combined datasets with environmental and microbial predictors for pitted scab. We only focus on the most accurate models identified in Section `Overall performance of predictive models: Manual binarization causes inaccurate prediction of yield'.
First, we note that a model that uses OTU abundances outperforms a model that uses alpha diversity as a predictor (comparison of Alpha with OTU-S3) for both types of models (RF and Bayesian NN). This suggests that we lose information by transforming abundances into diversity measures. Second, models including only OTU abundances (OTU-S3) perform comparably to models that include both types of predictors (OTU-S3+Soil+DS) which suggests that the microbial data indeed has substantial predictive power on its own, but adding microbiome to soil predictors may not provide much benefit for high predictive power, with the only exception of Phylum level OTU-S3+Soil+DS in a RF model (Figure \ref{15-RF-BNN-Com}). Generally, the model with only soil information (shown as a blue dashed line) performs just as accurately. Third, contrary to prior expectations that microbial communities at finer resolution would be a better choice for predicting pitted scab or other diseases, our study does not find any evidence that the prediction power increases when moving up from Phylum to Genus level.
Particularly, the prediction power of OTU-S3 in RF model increases from Class to Genus, and this pattern is not preserved when diversity is used instead of OTU abundances. For example, a model with only alpha diversity as the predictor (Alpha) shows decreasing weighted F1 score as we move from Phylum to Genus level. Both models (RF and Bayesian NN) when including all types of predictors (OTU-S3+Soil+DS) result in the similar weighted F1 score regardless of taxonomic level.
See Figures S\ref{S-RF-BNN-Com-Yield-Plant} to S\ref{S-RF-BNN-Com-Scabsuper} in the Supplementary Material for other responses. 

\section*{Discussion}


Soil microbiome represents the most complex and least understood aspect of soil health. In this study, we use ML techniques such as random forest  (RF) and Bayesian neural network (Bayesian NN) to determine whether soil microbial information has any predictive power for plant outcomes such as yield and disease. Our results show that the microbiome plays a role in predicting potato disease, but the most accurate prediction originates from a combination of both datatypes: microbiome and environmental factors. Given that prediction with environmental factors alone was sufficiently powerful, it is uncertain whether the extra expense to sequence microbiome data is worth the cost.

\paragraph{Can machines classify what humans cannot? The importance of accurate labels.}
The prediction power of microbiome data varies depending on the outcome we want to predict.  For example, among all models that predict diseases, models for the pitted scab disease receive very high weighted F1 scores compared to other diseases. We further confirm this predictive power by comparing the performance of models trained on the real microbiome data to models trained on randomly generated data (see Figures \ref{7-CompareToRandom} and \ref{S-CompareToRandom} in the Supplementary Material). Given that the prediction of the pitted scab disease is far from random, we can confidently conclude that this disease can be accurately predicted from microbiome data. It is noteworthy, however, that pitted scab disease is precisely one of the diseases that are easier to be visually detected, and thus, there is a reliable separation among the two classes (diseased and non-diseased) which is aiding in prediction by ML models. Other diseases, and more so yield, do not have such clear distinction between classes which results in lower predictive power. That is, we believe that the lack of prediction accuracy in yield is not driven by a lack of a biological connection between soil microbiome and yield, but on the lack of accurate labels that distinguish the two classes (e.g. low and high yield). If humans cannot distinguish what is low yield vs high yield, then that ambivalence will propagate into the ML classification. This conclusion seems to be confirmed when we notice that yield cannot be accurately predicted by any of 14 models in consideration (Table \ref{tableNameMethods}). We conclude that one of the main challenges when applying ML methods in biological applications is the artificial binarization of phenotypes. More work is needed to improve the performance of regression models that can predict continuous phenotypes when faced with limited sample sizes that are common in biological domains. 

\paragraph{Data preprocessing has a substantial impact in prediction performance.}
We show that the normalization and zero replacement strategies for microbiome data sets have a huge impact on results, and unfortunately, we could not find any pattern to suggest that one strategy is superior to others. The FMS decision tree, however, suggests that certain normalization methods such as centered log-ratio normalization with pseudo-zero replacement (clr+pseudo) or centered log-ratio normalization with Bayesian-multiplicative treatment (clr+bayesMult) should not be used for disease prediction for future studies. We recommend domain scientists to perform different types of normalization and zero replacement strategies in their data prior to be used as input for prediction. See also \cite{badri2020shrinkage, Pereira2018, callahan2017exact} for other performance comparisons on types of normalizations.

In addition, the data augmentation proved beneficial for our models in various ways. First, it added credibility for imbalanced responses such as the black scurf disease, as adding augmented data to the training set enables the model to train on both response labels of equal size, preventing the model from making biased decisions towards one label. Second, the model offers slightly better performance when training on the augmented data for all disease responses. 
Finally, data augmentation yields a more robust model, which lowers the chance of overfitting on the current data and should offer better performance on future data if we want to make decisions based on our current models. We advice practitioners, however, to be careful with data augmentation techniques to prevent data leakage. Unlike data augmentation in imaging data which is straight-forward, data augmentation of relative abundances can artificially boost accuracy by having the original sample in the training set and the augmented sample in the testing set. More work is needed to produce reliable data augmentation techniques for biological data that are not images beyond the standard Gaussian noise incorporation. 
\paragraph{Feature selection effectively preserves the predictive signal on lower dimensions.}
In terms of feature selection, we considered different strategies to select important features (ML, network comparison, and intersection of both).  There is significant overlap between important OTUs by two methods and the inclusion of this subset of predictors allowed us to build less complex models with comparable good performance (see, for example, Figure S\ref{11-RF-BNN-selectedfeatures} for pitted scab disease). We did not observe one feature selection strategy that outperformed the others. For example, the weighted F1 scores obtained when using the OTUs identified as important by the ML methods are comparable to the weighted F1 scores when including OTUs identified as important by network comparison. It is worth mentioning, however, that the performance is also comparable to that obtained when including all OTUs which shows that the feature selection strategies work at preserving the OTUs that have predictive signal while simultaneously allowing computationally expensive models (like Bayesian NN) to be applied. Tables S\ref{FSmethod} and S\ref{S-FSNetcomi} in the Supplementary Material contain the important OTUs identified by the two feature selection methods for the Phylum level associated with pitted scab disease.

\paragraph{Finer taxonomic levels do not provide higher predicting power in all scenarios.}
In our study, we could not identify an overall prediction superiority on certain taxonomic levels compared to others. 
While it is intuitive to expect that finer taxonomic levels would provide more power to predict potato outcomes, this is only true in some cases such as the OTU-S3 model in RF and the OTU-S1 model in Bayesian NN (Figure S\ref{11-RF-BNN-selectedfeatures}) where Genus level indeed outperforms the other taxonomic levels. However, we cannot find evidence to support this expectation in all other cases.
Even though this finding disagrees with common expectations in microbiome research, we believe that it is a positive result from the ML perspective. If less refined taxonomic levels perform just as well as finer taxonomic levels, then the model could be fitted with fewer features, and would thus be less computational expensive and more efficient. Furthermore, the more features in the data, the more samples are needed to train the models. So, ML models that are trained on limited samples sizes (as the ones we have in this study) would benefit greatly by having fewer features.


\paragraph{Limited predictive power in soil microbiome compared to environment.}
When including environmental features such as soil physicochemical properties and microbial population density of soil in the model, we achieved higher weighted F1 score values For \textit{Scabpit}, utilizing alpha diversity with the RF model yields a median weighted F1 score of approximately 0.75. When combined with soil population density information, this score increases to 0.85. Similarly, for the Bayesian NN model, the score improves from 0.6 to 0.8 with the addition of soil population density information. For RF, employing OTU3 results in a median weighted F1 score of about 0.8, which further increases to 0.9 when supplemented with soil population density data. Similarly, for BNN, the median score reaches approximately 0.9. However, incorporating the population density of soil information leads to a narrower range for the box plot.
In general, we investigated 14 different models for yield and disease prediction with different combinations of microbiome and environmental data. Results show poor performance in predicting yield across different models. The best results for pitted scab are achieved by combining alpha diversity and soil chemistry (Alpha+Soil) for RF and important OTUs and soil data (OTU-S3+Soil) for Bayesian NN. The median weighted F1 scores for predicting diseases range from 0.8 to 0.9 for RF and from 0.6 to 0.9 for Bayesian BNN models (refer to Figure \ref{bestRFBN}).
Although the best-performing models include microbiome predictors (Alpha and OTU-S3), it is important to note that the models without microbiome data are comparably powerful as those including microbiome data. Specifically, for the RF method, the F1 score exceeds 0.75 for \textit{Scabpit} and \textit{Black\_Scurf}, and surpasses 0.6 for \textit{Scab} and \textit{Scabsuper}. While the models trained with microbiome data alone show that microbiome can effectively predict pitted scab disease, the fact that models without this type of data continue to perform well provides evidence that microbiome data may not be necessary to achieve reasonable prediction. In fact, when the collection of microbiome data requires much higher cost investment, the increased prediction accuracy by including microbiome predictors may not be enough to justify the extra cost.

The RF and Bayesian NN methods offer similar performance in all models, which validates the results of both methods. However, in this study, Bayesian NN take far longer to train due to weight sampling and approximation, especially on more refined taxonomic level such as Order, Family, and Genus. Thus, we believe that based on the current dataset, RF is the preferred decision-making model with high prediction potential for disease outcomes when including a combination of microbiome and environmental predictors. 

\section*{Methods}

Figure \ref{3-Abstract} shows a graphical representation of our pipeline with three major steps: (1) data preparation, (2) feature selection, and (3) classification based on random forest  (RF) and Bayesian NN models with various types of predictors, including microbiome data (OTUs), environmental, and a combination of both.
We describe each step in the pipeline in the next subsections.


\subsection*{Data description}

In this study, we focus on the soil microbiome (matrix of abundances) in a variety of taxonomic orders, including Phylum, Class, Order, Family, and Genus as well as other environmental information from soil samples acquired from potato fields in Wisconsin and Minnesota.
The dataset consists of measurements related to soil health, potato yield and soil quality information. The soil health data are collected from pre-plant fields and include soil physicochemical properties, soil microbiome composition, soil microbiome diversity, and soil pathogen abundance. The potato yield and quality assessments are taken from the following growing season with indicators including tuber yield and disease severity. Overall, we have 256 samples, 108 of which are taken from fields in Minnesota, and 148 of which are taken from fields in Wisconsin. We list all measurements in Table S\ref{DS1} in the Supplementary Material.
\subsubsection*{Binarization of Response Variables}
We have six phenotypes (response variables) of interest, four of them correspond to diseases and two of them correspond to yield (Table \ref{DS2}).
\begin{table}[!h]
\centering
\scalebox{0.9}{
\begin{tabular}{ll}
\hline
Response & Description  \\\hline
Yield\_Meter     & Weight of potatoes (grams) per one meter yielded at harvest time.                   \\
Yield\_Plant     & Weight of potatoes (grams) per plant yielded at harvest time\\

Scab       & Number of tubers that carry scab (superficial+pitted) disease on the sampled plant.   \\
Scabpit    & Number of tubers that carry pitted scab disease on the sampled plant.               \\
Scabsuper  & Number of tubers that carry superficial scab disease on the sampled plant. \\
Black\_Scurf    & Percentage coverage of black scurf disease on the sampled plant \\
\hline
\end{tabular}}
\caption{\small{Description of responses of interest in this study.} }
\label{DS2}
\end{table}
All six responses in the dataset are continuous, so we need to binarize them to fit the classification models. 
For the disease-related responses, we simply make the binary label $0$ if there is no presence of disease, and $1$ if there was detection of disease (that is, if the continuous response is greater than $0.0$).
Binarizing the yield responses variables is harder as there is not a universal standard to classify potato yield to be low or high. Furthermore, yield values are highly dependent on the type of potato variety. We assign a label of $0$ (low yield) to samples with a yield less than the variety-specific median. Similarly, we assign a label of $1$ (high yield) to samples with yield greater than the variety-specific median. We illustrate this approach in Figure  S\ref{flowchart-label} in the Supplementary Material. 

After binarization, we note that pitted scab disease (denoted \textit{Scabpit} in the figures), superficial scab disease (denoted \textit{Scabsuper} in the figures), and both yield responses are balanced, whereas other responses are highly imbalanced: scab disease (denoted \textit{Scab} in the figures) has 80\% of samples labeled as $1$, and black scurf disease (denoted \textit{Black\_Scurf} in the figures) has only six samples labeled as $1$. We use these imbalanced cases to assess the performance of the methods under imbalance settings and data augmentation strategies.

\subsubsection*{Data Filtering, Normalization, and Zero Replacement}
\label{sec:norm}

The input data is a matrix with non-negative read counts that were generated by a sequencing procedure. Let $w^{(k)}= [w_1^{(k)},...,w_p^{(k)}]$ be the total read counts of sample $k$ containing $p$ OTUs, where $w^{(k)}$ is a composition that adds up to a fixed value of $m^{(k)}=\sum_{i=1}^p w_i^{(k)}$. This value $m^{(k)}$ is the sequencing depth, which varies across samples and is predetermined by technical factors resulting in highly sparse data. It is reasonable to filter out a certain set of OTUs as first data preparation step. For filtering, we only include OTUs that appear in at least 15 samples.  
Table S\ref{NOTU} displays the number of features (OTUs) before and after filtering for different taxonomic levels. 


As mentioned, the input data is compositional and highly sparse. It is known that ML methods do not perform well with unnormalized data \cite{mcmurdie2014waste} and with sparse data \cite{xia2018statistical}. Therefore, we explore the effect of four zero replacement strategies (to overcome sparsity) and five normalization strategies (to overcome compositionality). All strategies are implemented in the \texttt{NetCoMi} R package \cite{peschel2021netcomi}. 

In particular, we consider the four zero replacement strategies: (1) the original dataset which included zeros (denoted \textit{none} in the figures), (2) pseudo-zero replacement which replaces zero counts by a predefined pseudo count (denoted \textit{pseudo} in the figures), (3) multiplicative zero replacement which imputes left-censored compositional values by a given fraction and applies a multiplicative adjustment to preserve the multivariate compositional properties of the samples (denoted \textit{multRepl} in the figures) \cite{martin2003dealing}, and (4) Bayesian-multiplicative treatment which imputes zero counts by posterior estimates of the multinomial probabilities generating the counts, assuming a Dirichlet prior distribution (denoted \textit{bayesMult} in the figures) \cite{martin2015bayesian}.

Next, we use five normalization methods: (1) Total sum scaling which simply converts counts to appropriately scaled ratios (denoted \textit{TSS} in the figures) \cite{badri2020shrinkage}, (2) Cumulative sum scaling which rescales the samples based on a subset (quartile) of lower abundant taxa, thereby excluding the impact of highly abundant taxa (denoted \textit{CSS} in the figures) \cite{badri2020shrinkage}, (3) Common sum scaling in which counts are scaled to the minimum depth of each sample (denoted \textit{COM} in the figures) \cite{mcmurdie2014waste}, (4) Rarefying which random samples without replacement after a minimum count threshold has been applied (denoted \textit{rarefy} in the figures) \cite{gotelli2001quantifying}, and (5) Centered Log-ratio which transforms the data using the geometric mean as the reference (denoted \textit{clr} in the figures) \cite{aitchison1982statistical}. 

With four zero replacement methods and five normalization methods, we create 20 datasets by the combination of zero replacement and normalization techniques. Our goal is to study the effect of the zero replacement and normalization choice in the performance of the deep learning methods. Namely, we have the following 20 combinations,
$\text{NM}_1$: TSS+none,
$\text{NM}_2$: TSS+pseudo,
$\text{NM}_3$: TSS+multRepl,
$\text{NM}_4$: TSS+bayesMult,
$\text{NM}_5$: CSS+none,
$\text{NM}_6$: CSS+pseudo, 
$\text{NM}_7$: CSS+multRepl,
$\text{NM}_8$: CSS+bayesMult,
$\text{NM}_9$: COM+none,
$\text{NM}_{10}$: COM+pseudo,
$\text{NM}_{11}$: COM+multRepl,
$\text{NM}_{12}$: COM+bayesMult,
$\text{NM}_{13}$: rarefy+none,
$\text{NM}_{14}$: rarefy+pseudo,
$\text{NM}_{15}$: rarefy+multRepl,
$\text{NM}_{16}$: rarefy+bayesMult,
$\text{NM}_{17}$: clr+none,
$\text{NM}_{18}$: clr+pseudo,,
$\text{NM}_{19}$: clr+multRepl, and
$\text{NM}_{20}$: clr+bayesMult.
For convenience, we use the notation $\text{NM}_i$ (Normalization Method) for $i=1,\dots,20$ in the Full Model Selection section (See Section `Identifying general practices to predict potato disease from microbiome data using full model selection models').

For the environmental predictors of soil chemistry and microbial population density in the soil, we apply six scaling methods:
(1) standardize features by subtracting the mean and scaling to unit variance \cite{chan1983algorithms};
(2) scale each feature to a $[0,1]$ range;
(3) scale each feature by its maximum absolute value;
(4) scale features by subtracting the median and scaling to the interquantile range \cite{yeo2000new};
(5) transform the features to follow a normal distribution \cite{scikit-learn};
(6) normalize samples individually to the unit norm. 
After normalization, the datasets are split into training, validation,
and testing sets with 10-fold cross-validation. We used 80\% of samples
for training and validation, and 20\% for testing.
\subsubsection*{Data Augmentation}

There are three main goals that we wish to achieve with data augmentation: (1) improve the model's prediction performance with more artificial samples; (2) balance the number of labels with artificial samples, and (3) make the model more robust and avoid overfitting with unseen (artificial) data. We note that augmenting the whole dataset and then splitting into training and testing sets would result in \textit{data leakage}. For example, when the original sample is in the testing set and the augmented sample from this sample is in the training set, the model is essentially training and testing on the same sample since the normalized values of OTUs are very close. Thus, we split the data into training and testing sets first and only augment the training set. This strategy also allows us to have a fair performance comparison for augmented and non-augmented sets with the same testing data. 

Regarding the data augmentation procedure,
instead of simply adding a randomly generated noise to the original sample, we subset the data by variety and label, compute the mean (and standard deviation) abundance value for this subset, and create a new sample that includes the original data plus a Gaussian error with mean $\mu /100$ and standard deviation $\sigma /100$ where $\mu, \sigma$ are the subset-specific mean and standard deviation, respectively. This approach is illustrated in Figure S\ref{flowchart-aug} in the Supplementary Material.
By the end of this procedure we would have a balanced augmented training set with 400 samples per label for each of the five taxonomic levels (Phylum, Class, Order, Family, and Genus).

\subsection*{Feature Selection}

Feature selection involves the identification of important features (or covariates) that have high predictive power.
Given the high-dimensionality of the data (e.g. 256 original samples for 485 OTUs at the Genus level), feature selection is necessary, especially for Bayesian NN models that are computationally expensive. 

We pursue two approaches for feature selection: (1) using ML models to assess variable importance, and (2) using network analyses. To focus exclusively on the effect of feature selection, we only consider one type of normalization and zero replacement strategy in this investigation, namely, total sum scaling normalization without zero replacement ($\text{NM}_1$: TSS+none).

\subsubsection*{Using ML Models for Feature Selection}

To identify important OTUs, we use six ML strategies implemented in \texttt{scikit-learn} \cite{scikit-learn}: (1) "SelectKBest" method selects features based on the $k$ highest analysis of variance F-value scores, (2) select the top $k$ features based on the mutual information statistic, (3) recursive feature elimination (RFE) with logistic regression, (4) RFE with decision tree, (5) RFE with gradient boosting, and (6) RFE with RF.
In addition to the six ML strategies, we consider a 7th strategy which consists in including OTUs in the model if their maximum value is within the top 30\%.

After running all seven strategies, we assign a value ("TOTAL") to each OTU based on the number of times the OTU is selected as an important feature under the seven criteria. That is, an OTU that is selected as important by all seven strategies will have a value of 7. The OTUs are sorted based on "TOTAL" column and top 30\% of them are selected as important features. Thus, 30, 36, 75, 85, and 162 OTUs are selected for Phylum, Class, Order, Family and Genus levels, respectively.
For example, Table S\ref{FSmethod} in the Supplementary Material shows the top 30 OTUs at Phylum level and by which strategies they are identified as important features to predict the pitted scab response.

\subsubsection*{Using Network Comparison for Feature Selection }
Next, we identify important OTUs by comparing their interactions in microbial networks when the network is constructed with samples from one class (say, low yield) versus with samples from the other class (say, high yield).
Indeed, evidence shows that microbial interactions can help differentiate between crop disease states \cite{peschel2021netcomi}. We identify the OTUs that interact differently on the samples corresponding to one class versus another class as relevant OTUs that contain information about the outcome.

We use the \texttt{SPRING} method (semi-parametric rank-based approach for inference in graphical model) \cite{yoon2019microbial} in the \texttt{NetComi} package to construct two microbial networks: one corresponding to label $0$ and one corresponding to label $1$. To build these networks, the estimated partial correlations are transformed into dissimilarities via the signed distance metric and the corresponding similarities are then used as edge weights. We then compare these networks in order to identify OTUs that have the greatest difference based on degree values. Figure S\ref{1-NetComi} shows an example of two microbial networks with different graphical structures, and Table S\ref{S-FSNetcomi} in the Supplementary Material lists OTUs ranked by the difference in degree in the two microbial networks for diseased and non-diseased classes of pitted scab response. 
The OTUs are sorted based on the values of degree difference column and similar to ML strategy top 30\% of them are selected as important features.



\subsubsection*{Combination of ML and Network Strategies in Feature Selection}

We define a scoring value for each OTU based on whether they are identified as important by ML strategy ($score=1$), by the network comparison ($score=2$), or both ($score=3$). If the OTU is not identified as important by any strategy, it is denoted $score=0$. Figure S\ref{2-MLNetComi} displays the scoring values of all OTUs at the Phylum level for all responses.


\subsection*{Model descriptions: random forest and Bayesian neural network}

We apply two types of classification models to predict potato disease and yield: RF and Bayesian NN. 
We consider separately three types of predictors: OTU abundances (5 taxonomic levels and 20 normalization/zero replacement strategies described in Section `Data Filtering, Normalization, and Zero Replacement'), environmental predictors such as soil characteristics and microbial population density, and a combination of both types.
Table \ref{tableNameMethods} lists all the models we consider in this study.
\begin{table}[h]
\centering
\scalebox{0.9}{
\begin{tabular}{lll}
\hline
 Type of predictors & Name of the model & Predictors included in the model  \\
    \hline
    \multirow{2}{*}{OTU}&ALL-OTU&\small{OTU abundances}\\
    &OTU-S0& \small{OTUs with a score of zero (not selected by ML or network comparison} \\
    & & \small{feature selection strategies)} \\
    &OTU-S1&\small{OTUs selected by the ML feature selection strategy (score of one)}\\
    &OTU-S2&\small{OTUs selected by the network comparison feature selection strategy} \\
    & & \small{(score of two)}\\
    &OTU-S3&\small{OTUs selected by the ML and network comparison feature selection}\\
    & &\small{strategies (score of three)}\\
    &Alpha&\small{Alpha diversity}\\
    \hline
    \multirow{2}{*}{Environmental}&Soil & \small{Soil chemistry}\\
    &DS&\small{Microbial population density in soil}  \\
    &Soil+DS&\small{Combination of soil chemistry and microbial population density in soil} \\
    \hline
        \multirow{2}{*}{Combination}&Alpha+Soil&\small{Combination of alpha diversity and soil chemistry} \\
    &Alpha+Soil+DS&\small{Combination of alpha diversity, soil chemistry, and} \\
    && \small{microbial population density in soil} \\
    &OTU-S3+Soil&\small{Combination of OTUs with score of three and soil chemistry}\\
    &OTU-S3+DS&\small{Combination of OTUs with score of three and microbial population} \\ && \small{density in soil}\\
     &OTU-S3+Soil+DS&\small{Combination of OTUs with score of three, soil chemistry, and}\\ 
     & &\small{microbial population density in soil}\\ 
    \hline
\end{tabular}
}
\caption{\small{Models under study for three types of predictors (first column): OTUs, environmental and both. The name of each model (second column) is used in figures and tables in the text.} }
\label{tableNameMethods} 
\end{table}

Table S\ref{tablenumpredictor} shows the number of predictors included in each model as well as the data choices related to taxonomic level or normalization and zero replacement strategies.
For example, using all OTUs (first row), we have five taxonomic levels and 20 normalization+zero replacement strategies each, so in total, we have 100 (20 normalization strategies times 5 taxonomic levels) different OTU datasets. For each of these 100 datasets, the number of predictors (i.e., OTUs) would depend on the taxonomic level being analyzed. For example, at the Phylum level, there might be 42 predictors, while at the Genus level, there could be up to 485 predictors.

    
\subsubsection*{Random Forest Model}
\label{sec:Random Forest Model}


The RF classifier is a powerful ML technique that has gained significant popularity in the last two decades because of its accuracy and speed. RF randomly creates an ensemble of decision trees. Each tree picks a random set of samples (bagging) from the data and models the samples independently from other trees. Instead of relying on a single learning model, RF builds a collection of decision models, and the final decision is based on the output of all the trees in the model. The bagging approach promotes the generation of uncorrelated trees which reduces the risk of overfitting. 


Each decision tree is generated individually without any pruning and
each node is split using a user-defined number of features. By expanding the forest to a user-specified size, the technique generates trees with a high variance and low bias. The final classification choice is determined by summing the class-assignment probability obtained by each tree. A new unlabeled data in testing set input is thus compared to all decision trees formed in the ensemble, with each tree voting for class membership and the membership category with the most votes will be picked.

The RF has several hyper parameters to be determined by the user, such as the number of decision trees to be generated, the number of variables to be selected and tested
for the best split when growing the trees, the maximum depth of the tree, the minimum number of samples required to split an internal node, among others.
Generally, a grid search is combined with K-fold cross validation to select the best hyper parameters \cite{yan2022prediction}. \texttt{GridsearchCV} is a well-known search  method which is available in \texttt{scikit-learn} \cite{scikit-learn} and it evaluates all possible parameter combinations to determine optimal values. 

Here, we set different values for parameters (see Table S\ref{RFparameter} in the Supplementary Material) and tune them using \texttt{GridsearchCV} to find the optimal values for the RF classifier. \texttt{GridSearchCV} uses a ``score'' method for evaluating the performance of the cross-validated model on the test set. We use the F1 score to evaluate the performance of the model. The F1 score is the harmonic mean of precision and recall, where the proportional contributions of precision and recall to the F1 score are equal. A F1 score of 1 is the best, while a score of 0 is the worst. Finally, when the parameters for the RF model are tuned, we use 20\% of samples to report the performance of the final model. We use the weighted average F1 score to evaluate the performance of the models, which is computed by averaging all the per-class F1 scores while accounting for the number of samples in each class.

\subsubsection*{Bayesian Neural Network Model}
\label{BNN}
Given that neural network models are intrinsically data-hungry and microbiome applications have relatively small sample sizes, Bayesian NN models provide a suitable alternative for small sample sizes as they provide natural protection against overfitting. This is due to the fact that distributions are considered for the parameters in the model which allow us to marginalize them so that the prediction is based on data points alone \cite{soumya2019bnn}.

We provide details on the Bayesian NN model formulation next.
Let $\mathcal{D} = \{x_i, y_i\}_{i=1}^{n}$ be the data with $n$ samples where $x_i$ is a $N$-dimensional feature vector in $\mathbb{R}^N$ and $y_i \in \mathbb{R}$ is the target response. A Bayesian NN model has $L$ layers with $K_l$ neurons in the $l^{th}$ layer. The weight parameters are defined in a set of matrices $\mathcal{W} = \{W_l\}_{l=1}^{L}$ where each $W_l$ corresponds to the weights for the $l^{th}$ layer, and is of size $(K_{l-1}+1)\times K_l$. For an input $x_i\in \mathbb{R}^N$, the neural network maps it to a response $f(\mathcal{W}, \textbf{x})$ by multiplying the input of each layer by the weights and then transforming via an activation function such as rectified linear unit (ReLU): $h(a) = \max(0,a)$.

Unlike regular neural network models, a Bayesian NN imposes 
a prior distribution on the weights $\mathcal{W}\sim p(\mathcal{W})$ to capture their uncertainty. The posterior distribution of the weights given the data is then given by: 
$    p(\mathcal{W}|\mathcal{D}) = \frac{p(\mathcal{W})\prod_{i=1}^{n}p(y_i|f(\mathcal{W}, x_i))}{p(\mathcal{D})}$,
which can be used to predict the response $\textbf{y*}$ of unseen data $\textbf{x*}$ via
    $p(\textbf{y*}|\textbf{x*}) = \int p(\textbf{y*}|f(\mathcal{W}, \textbf{x*}))p(\mathcal{W}|\mathcal{D}) d\mathcal{W}$.

We use the original implementation of Bayesian NNs denoted ``software for flexible Bayesian modeling and Markov chain sampling" \cite{neal2012bayesian}
for training and testing on our data. 
For each Bayesian NN model, we use five hidden layers, one input layer and one output layer. The input layer has $n$ input neurons with $n$ being the number of predictors for the model. This number varies depending on the predictors to include. For example, for the ALL-OTU model for the Phylum level, we have 42 predictors (Table \ref{NOTU}), and thus the input layer has $N=42$ neurons. For each hidden layer, we have $3N$ hidden neurons and one bias node.
For the output layer, we only have one neuron with a binary value for the response. We use the hyperbolic tangent as the activation function: $f(x) = tanh(x)$ as opposed to the sigmoid function, also built in the software, as we found in practice that tanh performs much better.

We choose a hierarchical zero-mean Gaussian prior on all weights and on bias nodes. The prior for the weight from node $i$ of one layer to node $j$ of another layer is formulated as follows \cite{neal2012bayesian}: $\mathcal{P}(w_{ij} | \sigma_{w_{i}}, \sigma_{a_{j}})=\frac{1}{\sqrt{2\pi}\sigma_{w_i}\sigma_{a_j}}\exp(-\frac{w_{ij}^2}{2\sigma_{w_i}^2\sigma_{a_j}^2})$, 
where $\sigma_{w_i}$ are the 
hyperparameters for the standard deviation of the priors for node $i$ of layer $w$, and $\sigma_{a_j}$ is the hyper parameter for the standard deviation of the priors for node $j$ of layer $a$.
Let $\tau_{w_i} = \sigma_{w_i}^{-2}$, then the hyper prior is given by
\begin{equation}
    \mathcal{P}(\tau_{w_i}|\tau_w) = \frac{(\alpha_{w_i}/2\tau_w)^{\alpha_{w_i}/2-1}}{\Gamma(\alpha_{w_i}/2)}\tau_{w_i}^{\alpha_{w_i}/2-1}\exp(-\frac{\tau_{w_i}\alpha_{w_i}}{2\tau_{w}})
    \label{invgam}
\end{equation}
where $\alpha_{w_i}$ is the shape hyper parameter, and $\tau_w$ is the scale hyper parameter. For weights between hidden layers and bias nodes, we set $\alpha_{w_i} = 2$ and $\tau_w = 0.5$. For weights from the input layer to the first hidden layer, we set a prior for $\tau_w$ as an inverse gamma distribution (same as Equation \ref{invgam}) with shape hyper parameter $\alpha = 2$ and scale hyper parameter $\tau = 0.5$. This extra hyper parameter diminishes the effect of input nodes that have insufficient predictive power by setting weights arbitrarily close to zero.

The training and the approximation of the posterior distribution are done via a Hamiltonian (Hybrid) Monte Carlo (HMC) implemented in  \cite{neal2012bayesian}. We select different leap frog lengths and step sizes for different models. Leapfrog lengths begin at 100 as starting point and are gradually decreased as a leapfrog length of 100 is expected to deal optimally with challenging estimation problems. Step size, on the other hand, is set at 0.1 as a starting point and it is decreased and increased in search for an average rejection rate smaller than 0.3 \cite{neal2011hmc}. Burn-in is selected as the first half of the chain so that weights are sampled on the second half of the converged chain.

Given that HMC does not scale well for high dimensional parameter spaces and large datasets \cite{yao2019BNNcoomparison}, we did not fit the Bayesian NN on all the models described in Table \ref{tableNameMethods}. In particular, we do not fit a Bayesian NN model on the Genus level for the ALL-OTUs as this model would involve over 9 million weights in the network with 485 input neurons.

\subsection*{Full Model Selection}
\label{fms}

FMS \cite{sun2012full, sun2013towards, reyes2021full} involves the process of listing all data preprocessing steps, model options and selection of predictors, and using a decision tree model to identify the choices that yield the highest measure of performance. Here, we fit a FMS strategy with the following options: 1) type of normalization, 2) type of zero replacement, 3) taxonomic level, and 4) data augmentation. We combine the type of normalization and type of zero replacement strategy into one variable (denoted $\text{NM}_i$ for $i=1,\dots,20$). We focus on the weighted F1 score as measure of performance, and we include all OTU predictors (that is, we do not consider feature selection as one of the options to compare).
We build the regression decision tree by using the \texttt{DecisionTreeRegressor} 
which is available in \texttt{scikit-learn} \cite{scikit-learn}. We use the default parameters in the \texttt{DecisionTreeRegressor} such as "squared error" as the criterion to measure the quality of a split, a minimum number of 2 for the samples required to split an internal node, and a minimum number of 1 sample required to be at a leaf node. In order to create an informative decision tree that can be interpreted, we use a maximum depth of 4.

\bibliographystyle{plain}
\bibliography{references}

\begin{thebibliography}{10}

\bibitem{aghdam2015cn}
Rosa Aghdam, Mojtaba Ganjali, Xiujun Zhang, and Changiz Eslahchi.
\newblock {CN}: a consensus algorithm for inferring gene regulatory networks
  using the sorder algorithm and conditional mutual information test.
\newblock {\em Molecular BioSystems}, 11(3):942--949, 2015.

\bibitem{aitchison1982statistical}
John Aitchison.
\newblock The statistical analysis of compositional data.
\newblock {\em Journal of the Royal Statistical Society: Series B
  (Methodological)}, 44(2):139--160, 1982.

\bibitem{Anahtar2021}
Melis~N. Anahtar, Jason~H. Yang, and Sanjat Kanjilal.
\newblock Applications of machine learning to the problem of antimicrobial
  resistance: an emerging model for translational research.
\newblock {\em Journal of Clinical Microbiology}, 59(7):e01260--20, 2021.

\bibitem{badri2020shrinkage}
Michelle Badri, Zachary~D Kurtz, Richard Bonneau, and Christian~L M{\"u}ller.
\newblock Shrinkage improves estimation of microbial associations under
  different normalization methods.
\newblock {\em NAR genomics and bioinformatics}, 2(4):lqaa100, 2020.

\bibitem{belgiu2016random}
Mariana Belgiu and Lucian Dr{\u{a}}gu{\c{t}}.
\newblock Random forest in remote sensing: A review of applications and future
  directions.
\newblock {\em ISPRS journal of photogrammetry and remote sensing}, 114:24--31,
  2016.

\bibitem{Berendsen2012microbiome}
Roeland~L. Berendsen, Corné~M.J. Pieterse, and Peter~A.H.M. Bakker.
\newblock The rhizosphere microbiome and plant health.
\newblock {\em Trends in Plant Science}, 17(8):478--486, 2012.

\bibitem{callahan2017exact}
Benjamin~J Callahan, Paul~J McMurdie, and Susan~P Holmes.
\newblock Exact sequence variants should replace operational taxonomic units in
  marker-gene data analysis.
\newblock {\em The ISME journal}, 11(12):2639--2643, 2017.

\bibitem{chan1983algorithms}
Tony~F Chan, Gene~H Golub, and Randall~J LeVeque.
\newblock Algorithms for computing the sample variance: Analysis and
  recommendations.
\newblock {\em The American Statistician}, 37(3):242--247, 1983.

\bibitem{soumya2019bnn}
Soumya Ghosh, Jiayu Yao, and Finale Doshi-Velez.
\newblock Model selection in bayesian neural networks via horseshoe priors.
\newblock {\em Journal of Machine Learning Research}, 20(182):1--46, 2019.

\bibitem{gotelli2001quantifying}
Nicholas~J Gotelli and Robert~K Colwell.
\newblock Quantifying biodiversity: procedures and pitfalls in the measurement
  and comparison of species richness.
\newblock {\em Ecology letters}, 4(4):379--391, 2001.

\bibitem{guseva2022diversity}
Ksenia Guseva, Sean Darcy, Eva Simon, Lauren~V Alteio, Alicia
  Montesinos-Navarro, and Christina Kaiser.
\newblock From diversity to complexity: Microbial networks in soils.
\newblock {\em Soil Biology and Biochemistry}, 169:108604, 2022.

\bibitem{Hernandez2015BP}
Jos{\'e}~Miguel Hern{\'a}ndez-Lobato and Ryan~P. Adams.
\newblock {Probabilistic backpropagation for scalable learning of Bayesian
  Neural Networks}.
\newblock {\em ICML'15: Proceedings of the 32nd International Conference on
  International Conference on Machine Learning}, 37:1861--1869, July 2015.

\bibitem{Khodabakhsh2023}
Athar Khodabakhsh, Tobias~P. Loka, S{\'e}bastien Boutin, Dennis Nurjadi, and
  Bernhard~Y. Renard.
\newblock Predicting decision-making time for diagnosis over ngs cycles: An
  interpretable machine learning approach.
\newblock {\em bioRxiv}, 2023.

\bibitem{kurtz2015sparse}
Zachary~D Kurtz, Christian~L M{\"u}ller, Emily~R Miraldi, Dan~R Littman,
  Martin~J Blaser, and Richard~A Bonneau.
\newblock Sparse and compositionally robust inference of microbial ecological
  networks.
\newblock {\em PLoS computational biology}, 11(5):e1004226, 2015.

\bibitem{Lim2021}
Seongmin Lim, Jin-Hyung Kim, and Hae-Dong Kim.
\newblock Strategy for on-orbit space object classification using deep
  learning.
\newblock {\em Proceedings of the Institution of Mechanical Engineers, Part G:
  Journal of Aerospace Engineering}, 235(15):2326--2341, 2021.

\bibitem{martin2003dealing}
Josep~A Mart{\'\i}n-Fern{\'a}ndez, Carles Barcel{\'o}-Vidal, and Vera
  Pawlowsky-Glahn.
\newblock Dealing with zeros and missing values in compositional data sets
  using nonparametric imputation.
\newblock {\em Mathematical Geology}, 35(3):253--278, 2003.

\bibitem{martin2015bayesian}
Josep-Antoni Mart{\'\i}n-Fern{\'a}ndez, Karel Hron, Matthias Templ, Peter
  Filzmoser, and Javier Palarea-Albaladejo.
\newblock Bayesian-multiplicative treatment of count zeros in compositional
  data sets.
\newblock {\em Statistical Modelling}, 15(2):134--158, 2015.

\bibitem{mcmurdie2014waste}
Paul~J McMurdie and Susan Holmes.
\newblock Waste not, want not: why rarefying microbiome data is inadmissible.
\newblock {\em PLoS computational biology}, 10(4):e1003531, 2014.

\bibitem{neal2011hmc}
Radford~M. Neal.
\newblock {\em MCMC Using Hamiltonian Dynamics}.
\newblock CRC Press, 2011.

\bibitem{neal2012bayesian}
Radford~M. Neal.
\newblock {\em Bayesian learning for neural networks}, volume 118.
\newblock Springer Science \& Business Media, 2012.

\bibitem{scikit-learn}
F.~Pedregosa, G.~Varoquaux, A.~Gramfort, V.~Michel, B.~Thirion, O.~Grisel,
  M.~Blondel, P.~Prettenhofer, R.~Weiss, V.~Dubourg, J.~Vanderplas, A.~Passos,
  D.~Cournapeau, M.~Brucher, M.~Perrot, and E.~Duchesnay.
\newblock Scikit-learn: Machine learning in {P}ython.
\newblock {\em Journal of Machine Learning Research}, 12:2825--2830, 2011.

\bibitem{Pereira2018}
Mariana~Buongermino Pereira, Mikael Wallroth, Viktor Jonsson, and Erik
  Kristiansson.
\newblock Comparison of normalization methods for the analysis of metagenomic
  gene abundance data.
\newblock {\em BMC Genomics}, 19(1):274, 2018.

\bibitem{peschel2021netcomi}
Stefanie Peschel, Christian~L M{\"u}ller, Erika von Mutius, Anne-Laure
  Boulesteix, and Martin Depner.
\newblock Netcomi: network construction and comparison for microbiome data in
  r.
\newblock {\em Briefings in bioinformatics}, 22(4):bbaa290, 2021.

\bibitem{reyes2021full}
JF~Mandujano Reyes, E~Walleser, S~Hachenberg, S~Gruber, M~Kammer,
  C~Baumgartner, R~Mansfeld, K~Anklam, and D~D{\"o}pfer.
\newblock Full model selection using regression trees for numeric predictions
  of biomarkers for metabolic challenges in dairy cows.
\newblock {\em Preventive Veterinary Medicine}, 193:105422, 2021.

\bibitem{sun2012full}
Quan Sun, Bernhard Pfahringer, and Michael Mayo.
\newblock Full model selection in the space of data mining operators.
\newblock In {\em Proceedings of the 14th annual conference companion on
  genetic and evolutionary computation}, pages 1503--1504, 2012.

\bibitem{sun2013towards}
Quan Sun, Bernhard Pfahringer, and Michael Mayo.
\newblock Towards a framework for designing full model selection and
  optimization systems.
\newblock In {\em International Workshop on Multiple Classifier Systems}, pages
  259--270. Springer, 2013.

\bibitem{Trivedi2016Indicators}
Pankaj Trivedi, Manuel Delgado-Baquerizo, Ian~C. Anderson, and Brajesh~K.
  Singh.
\newblock Response of soil properties and microbial communities to agriculture:
  Implications for primary productivity and soil health indicators.
\newblock {\em Frontiers in Plant Science}, 7, 2016.

\bibitem{wagg2019fungal}
Cameron Wagg, Klaus Schlaeppi, Samiran Banerjee, Eiko~E Kuramae, and Marcel~GA
  van~der Heijden.
\newblock Fungal-bacterial diversity and microbiome complexity predict
  ecosystem functioning.
\newblock {\em Nature communications}, 10(1):4841, 2019.

\bibitem{xia2018statistical}
Yinglin Xia, Jun Sun, Ding-Geng Chen, et~al.
\newblock {\em Statistical analysis of microbiome data with R}, volume 847.
\newblock Springer, 2018.

\bibitem{yan2022prediction}
Tao Yan, Shui-Long Shen, Annan Zhou, and Xiangsheng Chen.
\newblock Prediction of geological characteristics from shield operational
  parameters by integrating grid search and k-fold cross validation into
  stacking classification algorithm.
\newblock {\em Journal of Rock Mechanics and Geotechnical Engineering}, 2022.

\bibitem{yao2019BNNcoomparison}
Jiayu Yao, Weiwei Pan, Soumya Ghosh, and Finale Doshi-Velez.
\newblock Quality of uncertainty quantification for bayesian neural network
  inference.
\newblock {\em Proceedings at the International Conference on Machine Learning:
  Workshop on Uncertainty \& Robustness in Deep Learning}, June 2019.

\bibitem{yeo2000new}
In-Kwon Yeo and Richard~A Johnson.
\newblock A new family of power transformations to improve normality or
  symmetry.
\newblock {\em Biometrika}, 87(4):954--959, 2000.

\bibitem{yoon2019microbial}
Grace Yoon, Irina Gaynanova, and Christian~L M{\"u}ller.
\newblock Microbial networks in spring-semi-parametric rank-based correlation
  and partial correlation estimation for quantitative microbiome data.
\newblock {\em Frontiers in genetics}, 10:516, 2019.

\end{thebibliography}

\section*{Data availability}
Data are available upon request to Richard Lankau (lankau@wisc.edu).

\section*{Code Availability}
All reproducible scripts are open source and publicly available in \url{https://github.com/solislemuslab/soil-microbiome-nn}.

\section*{Acknowledgements}
The authors thank Linda Kinkel and the Kinkel group. This work was supported by the Department of Energy [DE-SC0021016 to CSL]. The work was also supported by USDA Specialty Crop Multi-State Grant Program award SCMP1701.

\section*{Author contributions}
CSL and RL developed the idea.
RL and SS collected the data. XT and RL led all statistical analyses from data preprocessing to fitting of machine learning models, as well as summarizing the results by the creation of figures.
CSL, XT and RA wrote the initial complete draft of the manuscript. RL and SS contributed in biological interpretations and edition of the
manuscript. All authors read and approved the final manuscript.
\section*{Additional information}


\subsection*{Competing interests}
The authors declare that they have no competing interests.

\newpage
\appendix
\begin{center}
\LARGE Supplementary Material: \\Human limits in Machine Learning: Prediction of plant phenotypes using soil microbiome data
\end{center}

\setcounter{equation}{0}
\setcounter{figure}{0}
\setcounter{table}{0}
\setcounter{page}{1}

\listofappendixtables
\listofappendixfigures
\newpage

\begin{table}[!h]
\centering
\begin{tabular}{l|l}
\hline
\textbf{Soil health}&\\
\hline
Physicochemical (Soil)& pH \\
& Texture \\
& Organic matter content \\
& Cation/Anion exchange capacity \\
& Carbon fractions \\
& Organic nitrogen \\
& Available macronutrients \\
& Available micronutrients \\
& Potential carbon mineralization rate\\
Microbiome (OTU and alpha diversity) & Bacterial community composition, structure, diversity\\
& Fungal community composition, structure, diversity\\
& Bacterial and fungal population abundance\\
Pathogen & Verticillium dahliae, Pathogenic Streptomyces population\\ & Disease suppressiveness (DS) \\
& lab-estimated soil disease suppressiveness \\
\hline
\textbf{Potato crop assessment} \\
\hline
Yield & Tuber Yield, yield by size class \\
Disease severity &  \\
& Verticillium dahlia incidence \\
& Common scab disease incidence and severity \\
& Other diseases including hollow heart, silver scurf and black\\ & scurf\\
\hline
\textbf{Management practices}\\
\hline
Field history & Crop rotation \\
Practices around growing season & Potato cultivar selection, fumigation, \\ 
&cover crop, fertilization, pesticide, irrigation\\
\hline
\end{tabular}
\caption[Description of variables in Soil]{Description of   variables included in the soil dataset.} 
\label{DS1}
\end{table}

\begin{table}[!h]
\centering
\begin{tabular}{lcc}
\hline
Level & \# OTUs & \# OTUs after filtering \\\hline
Phylum    & 57         & 42        \\
Class     & 152        & 108          \\
Order     & 378        & 224          \\
Family    & 476        & 255          \\
Genus     & 1319       & 485    \\
\hline
\end{tabular}
\caption{Number of OTUs per taxonomic level (first column) in the original data (second column) and after filtering out OTUs that do not appear in at least 15 samples (third column).} 
\label{NOTU}
\end{table}

\begin{table}[!h]
\centering
\begin{tabular}{lccc}
\hline
 Model & Taxonomic levels & Normalization/zero replacement strategies & Number of predictors\\
    \hline
ALL-OTU & 5 & 20 & 42--485\\
OTU-S0 & 5 & 1 & 7--54 \\
OTU-S1 & 5 & 1 & 5--134\\
OTU-S2 & 5 & 1 & 5--134\\
OTU-S3 & 5 & 1 & 7--54\\
Alpha & 5 & 6 & 9\\
Soil & -- & 6 & 12\\
DS & --& 6 & 4\\
\hline
\end{tabular}
\caption{Types of predictors per model, number of taxonomic levels, normalization/zero replacement strategies, and range/number of predictors for each type.}\label{tablenumpredictor}
\end{table}

\definecolor{blue}{HTML}{99CCFF}

\tikzstyle{startstop} = [rectangle, rounded corners, minimum width=3cm, minimum height=0.8cm,text centered, draw=black, fill=blue]
\tikzstyle{io} = [trapezium, trapezium left angle=70, trapezium right angle=110, minimum width=3cm, minimum height=0.8cm, text centered, draw=black, fill=blue]
\tikzstyle{process} = [rectangle, minimum width=3cm, minimum height=0.8cm, text centered, draw=black, fill=blue]
\tikzstyle{decision} = [diamond, aspect=3, minimum width=3cm, minimum height=0.8cm, text centered, draw=black, fill=blue]
\tikzstyle{arrow} = [thick,->,>=stealth]

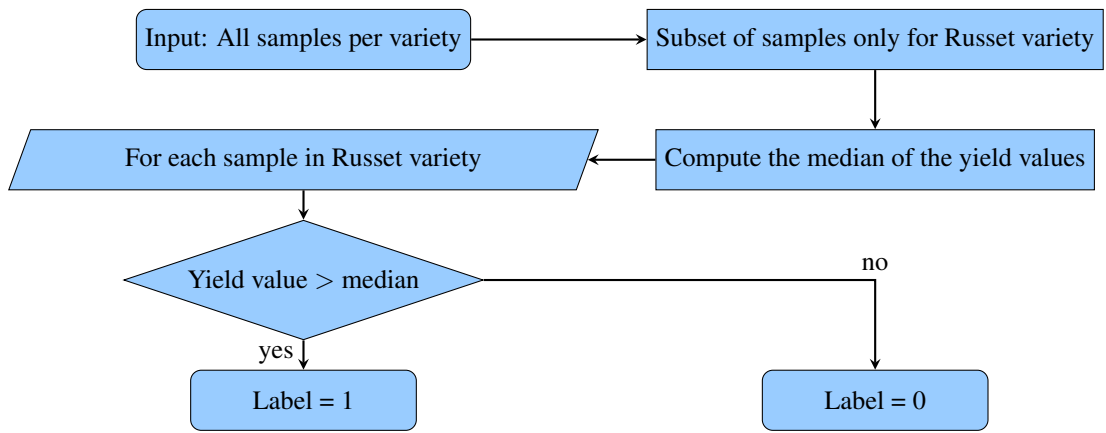
\begin{figure}[h]
    \centering
        \begin{tikzpicture}[node distance=1.6cm]
        \node (start) [startstop] {Input: All samples per variety};
        \node (proc1) [process, right of=start, xshift=6cm] {Subset of samples only for Russet variety};
        \node (proc2) [process, below of=proc1] {Compute the median of the yield values};
        \node (in1) [io, left of=proc2, xshift=-6cm] {For each sample in Russet variety};
        \node (dec1) [decision, below of=in1] {Yield value $>$ median};
        \node (out1) [startstop, below of=dec1] {Label = 1};
        \node (out2) [startstop, right of=out1, xshift=6cm] {Label = 0};
        
        \draw [arrow] (start) -- (proc1);
        \draw [arrow] (proc1) -- (proc2);
        \draw [arrow] (proc2) -- (in1);
        \draw [arrow] (in1) -- (dec1);
        \draw [arrow] (dec1) -- node[anchor=east] {yes} (out1);
        \draw [arrow] (dec1) -| node[anchor=south] {no} (out2);
        \end{tikzpicture}
    \caption[Flowchart for binarizing the continuous yield response]{Flowchart for binarizing the continuous yield response into binary labels for the Russet variety. The same procedure is used for every variety in the dataset.}
    \label{flowchart-label}
\end{figure}

\definecolor{blue}{HTML}{99CCFF}

\tikzstyle{startstop} = [rectangle, rounded corners, minimum width=3cm, minimum height=0.8cm,text centered, draw=black, fill=blue]
\tikzstyle{io} = [trapezium, trapezium left angle=70, trapezium right angle=110, minimum width=3cm, minimum height=0.8cm, text centered, draw=black, fill=blue]
\tikzstyle{process} = [rectangle, minimum width=3cm, minimum height=0.8cm, text centered, draw=black, fill=blue]
\tikzstyle{decision} = [diamond, aspect=2, minimum width=3cm, minimum height=0.8cm, text centered, draw=black, fill=blue]
\tikzstyle{arrow} = [thick,->,>=stealth]

\begin{figure}[h]
    \centering
    \begin{tikzpicture}[node distance=1.4cm]
        \node (start) [startstop] {Input: training set of original OTUs \& all labels};
        \node (dec1) [decision, below of=start] {\# Label $0$ $<$ 400};
        \node (pro0a) [process, below of=dec1, yshift=-0.5cm] {Select Label $0$};
        \node (dec2) [decision, right of=dec1, xshift=3cm] {\# Label $1$ $<$ 400};
        \node (pro0b) [process, below of=dec2, yshift=-0.5cm] {Select Label $1$};
        \node (pro1b) [process, below of=pro0a] {For each potato variety};
        \node (in1) [io, below of=pro1b] {Select the subset of samples for that variety};
        \node (pro2) [process, below of=in1] {Calculate the mean and variance of each OTU in the subset};
        \node (pro3) [process, below of=pro2] {Generate noises $\mathcal{N}\sim (\mu/100, \sigma/100)$};
        \node (pro4) [process, below of=pro3] {Augmented value = max(0, original+noise)};
        \node (in2) [io, below of=pro4] {Add this augmented sample};
        \node (end) [startstop, right of=in2, xshift=7cm] {Output: 800 samples};
        
        \draw [arrow] (start) -- (dec1);
        \draw [arrow] (dec1) -- node[anchor=east] {yes} (pro0a);
        \draw [arrow] (dec1) -- node[anchor=south] {no} (dec2);
        \draw [arrow] (dec2) -- node[anchor=east] {yes} (pro0b);
        \draw [arrow] (dec2) -| node[anchor=south] {no} (end);
        \draw [arrow] (pro0a) -- (pro1b);
        \draw [arrow] (pro0b) |- (pro1b);
        \draw [arrow] (pro1b) -- (in1);
        \draw [arrow] (in1) -- (pro2);
        \draw [arrow] (pro2) -- (pro3);
        \draw [arrow] (pro3) -- (pro4);
        \draw [arrow] (pro4) -- (in2);
        \draw [arrow] (in2.west) -- ++(-1.5,0)  |-(start);
    \end{tikzpicture}
    \caption[Flowchart for data augmentation algorithm]{Flowchart with the data augmentation algorithm. The target sample size for the training set is 800 with 400 samples for each label. The noise that we artificially generate needs to be variety-specific before adding to the original samples so that the biological implications of the original samples would be preserved.}
    \label{flowchart-aug}
\end{figure}
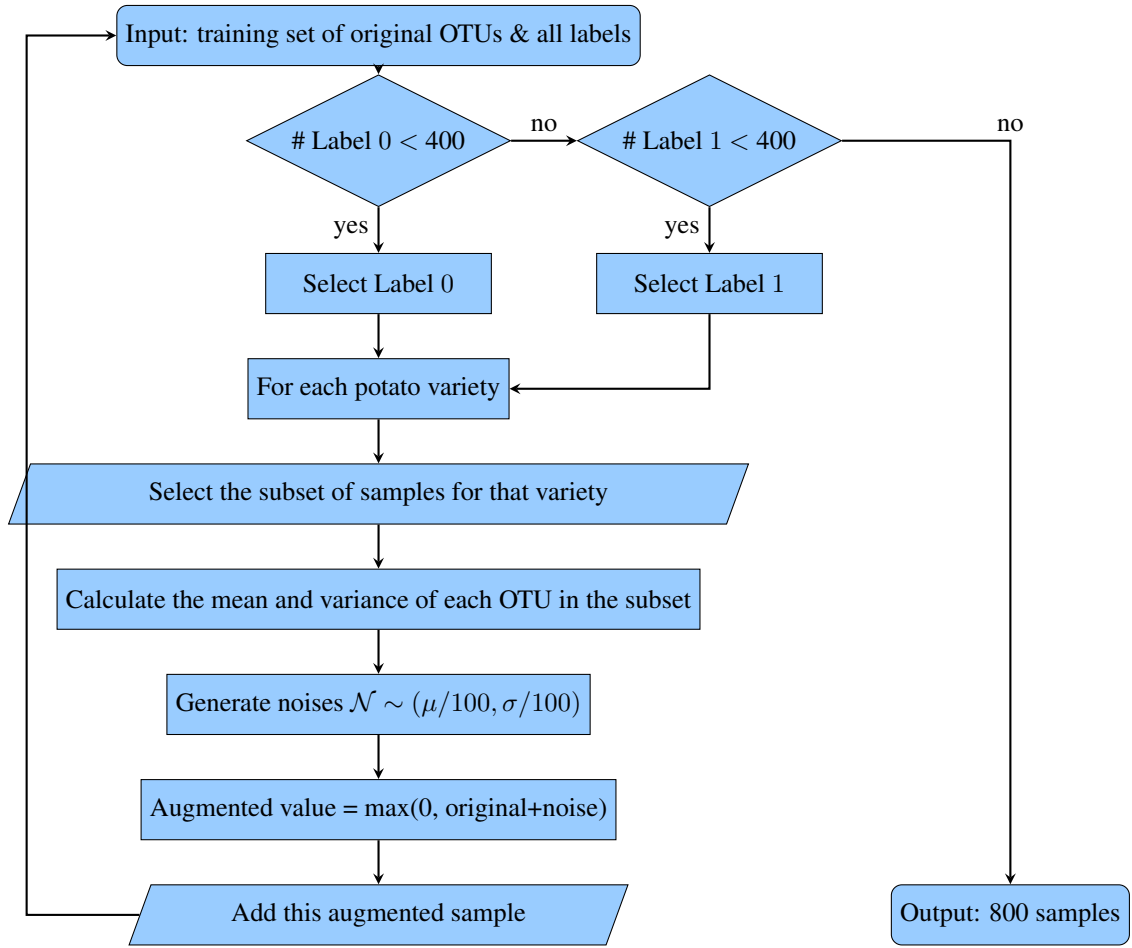

\begin{figure}[h]
\centering
\includegraphics[scale=0.35]{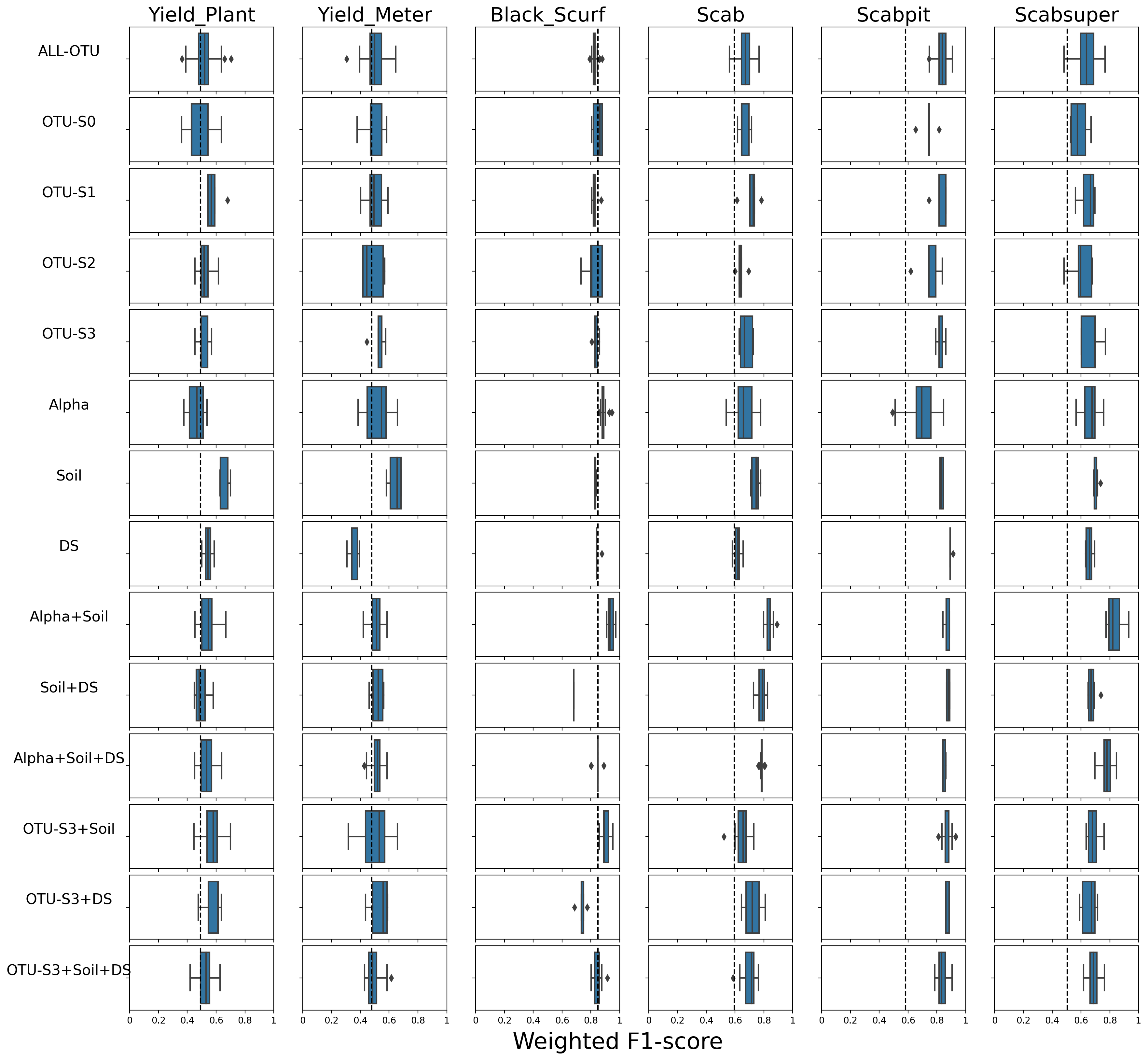}
\caption[Boxplots of weighted F1 scores for random forest models for different types of predictors (rows) and different yield or disease outcomes (columns)]{Boxplots of weighted F1 scores for random forest models for different types of predictors (rows) and different yield or disease outcomes (columns). For description of the rows, see Table \ref{tableNameMethods} in the main text. The range of each boxplot depicts the weighted F1 scores for datasets in different taxonomic levels and different normalization and zero replacement strategies. The dashed line corresponds to the weighted F1 score when fitting the model with random datasets(see Section \ref{sec-random}).}
\label{4-ALL-RF}
\end{figure}

\begin{figure}[h]
\centering
\includegraphics[scale=0.35]{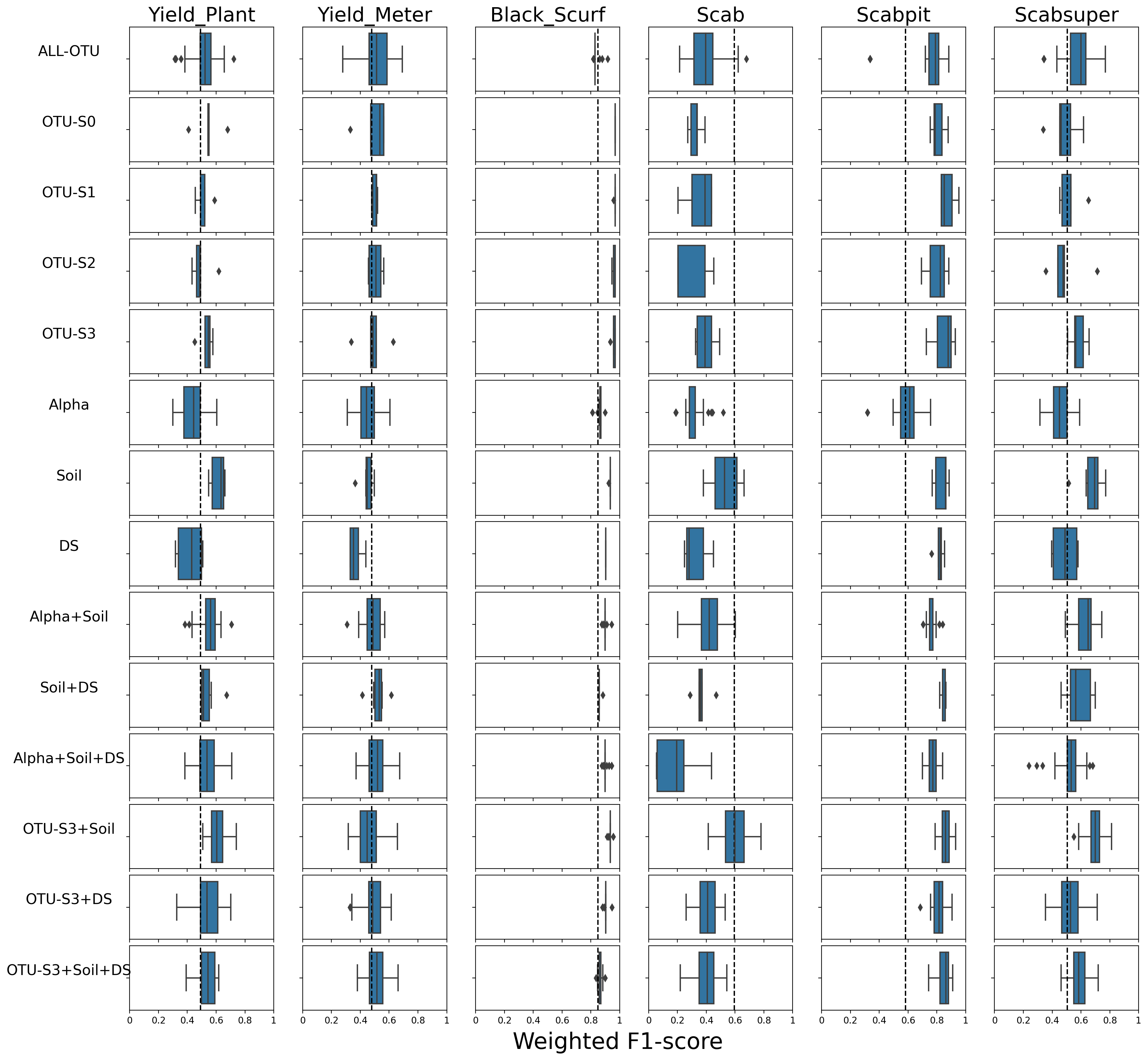}
\caption[Boxplots of weighted F1 scores for Bayesian neural network models for different types of predictors (rows) and different yield or disease outcomes (columns)]{Boxplots of weighted F1 scores for Bayesian neural network models for different types of predictors (rows) and different yield or disease outcomes (columns). For description of the rows, see Table \ref{tableNameMethods} in the main text. The range of each boxplot depicts the weighted F1 scores for datasets in different taxonomic levels and different normalization and zero replacement strategies. The dashed line corresponds to the weighted F1 score when fitting the model with random datasets (see Section \ref{sec-random}).}
\label{5-ALL-BNN}
\end{figure}

\begin{table}[ht]
\centering
\resizebox{0.7\textwidth}{!}{\begin{tabular}{llrrrrrrr}
\hline
                  & KBest & Mutual & LR & DT & GB & RF & Max & TOTAL \\
\hline
Firmicutes        & X     & X      & X  & X  & X  & X  & X   & 7     \\
Patescibacteria   & X     & X      & X  & X  & X  & X  & X   & 7     \\
Myxococcota       & -     & -      & X  & X  & X  & X  & X   & 5     \\
Methylomirabilota & X     & X      & X  & X  & -  & -  & X   & 5     \\
Acidobacteriota   & -     & -      & X  & X  & X  & X  & X   & 5     \\
Verrucomicrobiota & -     & -      & X  & X  & X  & -  & X   & 4     \\
Chloroflexi       & -     & -      & X  & X  & -  & X  & X   & 4     \\
Desulfobacterota  & X     & X      & X  & -  & -  & X  & -   & 4     \\
NB1.j             & X     & X      & -  & X  & -  & -  & -   & 3     \\
Thermoplasmatota  & X     & X      & -  & -  & X  & -  & -   & 3     \\
RCP2.54           & -     & -      & -  & X  & X  & X  & -   & 3     \\
WS2               & -     & X      & -  & X  & X  & -  & -   & 3     \\
Dependentiae      & -     & -      & -  & X  & X  & X  & -   & 3     \\
WPS.2             & X     & X      & -  & -  & -  & X  & -   & 3     \\
Cyanobacteria     & -     & -      & X  & -  & X  & -  & X   & 3     \\
Armatimonadota    & X     & X      & -  & X  & -  & -  & -   & 3     \\
Elusimicrobiota   & X     & -      & -  & -  & -  & X  & -   & 2     \\
Latescibacterota  & X     & -      & -  & -  & -  & X  & -   & 2     \\
Bdellovibrionota  & -     & -      & -  & -  & X  & X  & -   & 2     \\
Bacteroidota      & -     & -      & X  & -  & -  & -  & X   & 2     \\
Spirochaetota     & X     & X      & -  & -  & -  & -  & -   & 2     \\
Nitrospirota      & -     & -      & X  & -  & -  & -  & X   & 2     \\
Actinobacteriota  & -     & -      & X  & -  & -  & -  & X   & 2     \\
Planctomycetota   & -     & -      & -  & X  & -  & X  & -   & 2     \\
Proteobacteria    & -     & -      & X  & -  & -  & -  & X   & 2     \\
SAR324            & -     & X      & -  & -  & -  & -  & -   & 1     \\
Abditibacteriota  & -     & -      & -  & -  & X  & -  & -   & 1     \\
Nitrospinota      & -     & X      & -  & -  & -  & -  & -   & 1     \\
Nanoarchaeota     & -     & X      & -  & -  & -  & -  & -   & 1     \\
Gemmatimonadota   & -     & -      & -  & -  & -  & -  & X   & 1     \\
GAL15             & -     & -      & -  & -  & X  & -  & -   & 1     \\
Deinococcota      & X     & -      & -  & -  & -  & -  & -   & 1     \\
MBNT15            & X     & -      & -  & -  & -  & -  & -   & 1   
\end{tabular}}
\caption[Selected features with different Machine Learning methods]{The Operational Taxonomic Units (OTUs) based on their maximum values in samples are sorted and the features that their maximum values are among the top 30\% (Max column) are selected. The selected features by the SelectKBest method are marked in the KBest column.  Columns three to six are  the result of applying of logistic regression (LR), decision tree (DT), Gradient Boosting (GB), or Random Forrest (RF) as the choice of the algorithm using in the Recursive Feature Elimination method. The selected features are also considered based on mutual information statistics as shown in the Mutual column. We assigned a TOTAL value  to each OTU based on the number of times the OTU gets picked by any of the seven criteria (six ML models and the max OTU value) shown in the TOTAL column for the Phylum level associated with Scabpit disease. The OTUs with scoring values higher than zero are shown. }
\label{FSmethod}
\end{table} 

\begin{table}[ht]
\centering
\resizebox{0.5\textwidth}{!}{\begin{tabular}{lccc}
  \hline
 & Degree 0 & Degree 1 & Degree difference\\
  \hline
 MBNT15     & 1 & 9 & 8  \\
 Actinobacteriota          & 8 & 3 & 5   \\
 NB1.j       & 3 & 8 & 5  \\
 Sva0485    & 6 & 2 & 4   \\
 Bacteroidota         & 4 & 7 & 3  \\
 Halobacterota         & 4 & 1 & 3  \\
 Methylomirabilota         & 5 & 2 & 3  \\
 Proteobacteria        & 3 & 6 & 3  \\
 Chloroflexi          & 4 & 6 & 2   \\
 Cyanobacteria          & 0 & 2 & 2  \\
 Entotheonellaeota        & 3 & 1 & 2  \\
 GAL15      & 1 & 3 & 2  \\
 Hydrogenedentes        & 2 & 0 & 2 \\
 Patescibacteria       & 0 & 2 & 2  \\
 Planctomycetota       & 0 & 2 & 2 \\
 RCP2.54     & 2 & 4 & 2  \\
 WPS.2       & 2 & 0 & 2  \\
 Zixibacteria           & 1 & 3 & 2  \\
 Abditibacteriota          & 0 & 1 & 1   \\
 Desulfobacterota         & 4 & 5 & 1  \\
 Elusimicrobiota       & 0 & 1 & 1  \\
 Fibrobacterota        & 2 & 3 & 1  \\
 Gemmatimonadota        & 2 & 1 & 1   \\
 Myxococcota           & 3 & 2 & 1 \\
 Nanoarchaeota          & 2 & 1 & 1   \\
 Nitrospirota           & 3 & 2 & 1 \\
 Spirochaetota        & 3 & 4 & 1 \\
 Sumerlaeota            & 1 & 0 & 1  \\
 Thermoplasmatota         & 1 & 0 & 1  \\
 Verrucomicrobiota        & 0 & 1 & 1  \\
 Acidobacteriota        & 3 & 3 & 0   \\
 Armatimonadota         & 0 & 0 & 0     \\
 Bdellovibrionota          & 2 & 2 & 0    \\
 Crenarchaeota         & 1 & 1 & 0  \\
 Deinococcota           & 0 & 0 & 0     \\
 Dependentiae           & 0 & 0 & 0    \\
 FCPU426     & 0 & 0 & 0 \\
 Firmicutes     & 1 & 1 & 0   \\
 Latescibacterota          & 2 & 2 & 0   \\
 Nitrospinota           & 4 & 4 & 0   \\
 SAR324 & 0 & 0 & 0    \\
 WS2        & 1 & 1 & 0      
\end{tabular}}
\caption[List of OTUs in microbial networks constructed by \texttt{SPRING} 
for two classes of scabpit]{List of OTUs in microbial networks constructed by \texttt{SPRING} 
for two classes of scabpit. Degree $0$ correspond to the degree of the OTU node in the network built with the samples with Label $0$.  Degree $1$ corresponds to the degree of the OTU node in the network built with samples with Label $1$}\label{S-FSNetcomi}
\end{table}
\begin{figure}[ht!]
\centering
\includegraphics[scale=0.3]{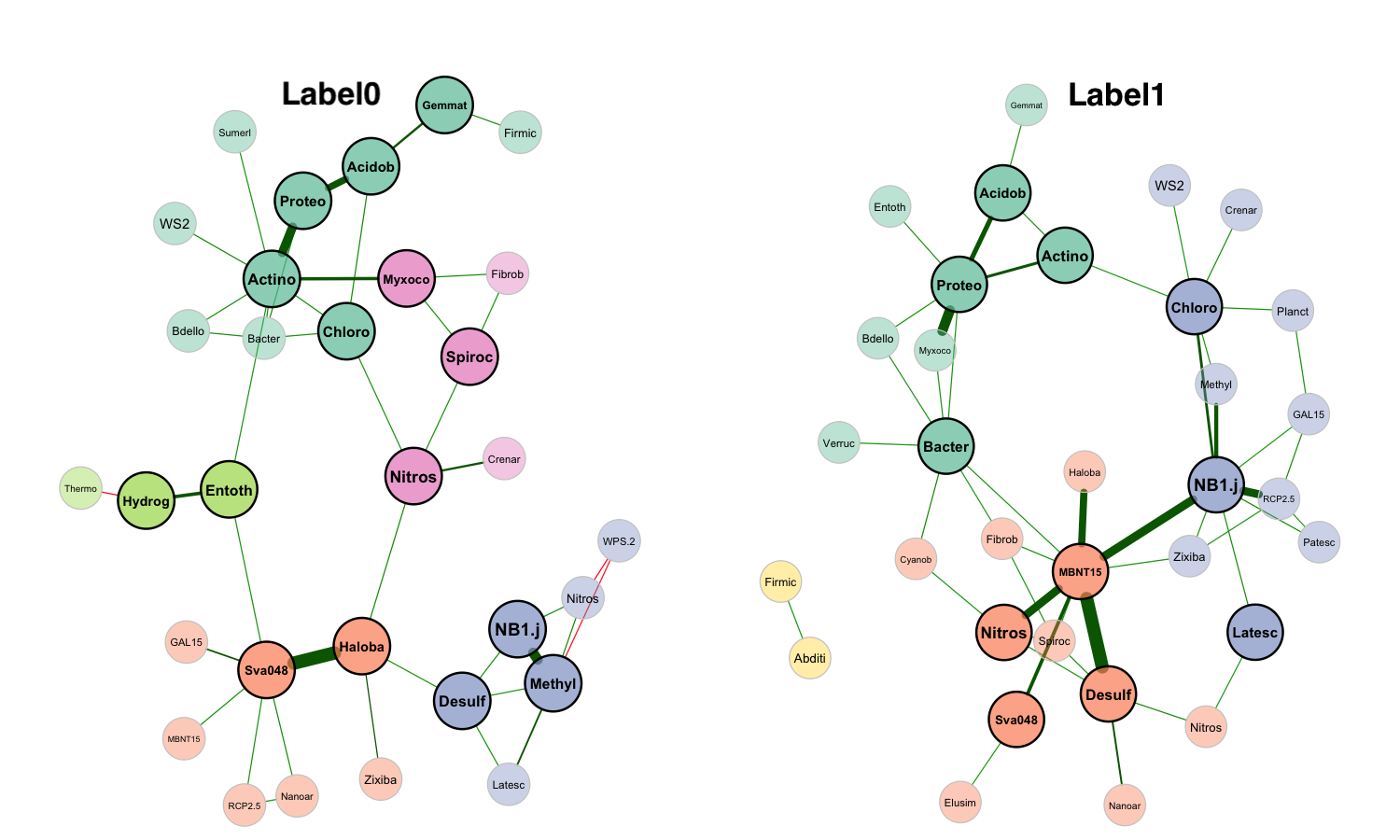}
\caption{Microbial networks at Phylum level for non-diseased (Label $0$) and diseased (Label $1$) samples for pitted scab (Scabpit). We use the \texttt{SPRING} method \cite{yoon2019microbial} to reconstruct these microbial networks.
Node colors represent clusters, which are determined using greedy modularity optimization.
Clusters have the same color in both networks if they share at least two OTUs. Green edges correspond to positive associations and red edges to negative ones.  Nodes that are unconnected in both groups are removed. OTUs names are abbreviated
(see Table S\ref{S-FSNetcomi} in the Appendix for the original names).}
\label{1-NetComi}
\end{figure}

\begin{figure}[ht!]
\centering
\includegraphics[scale=0.6]{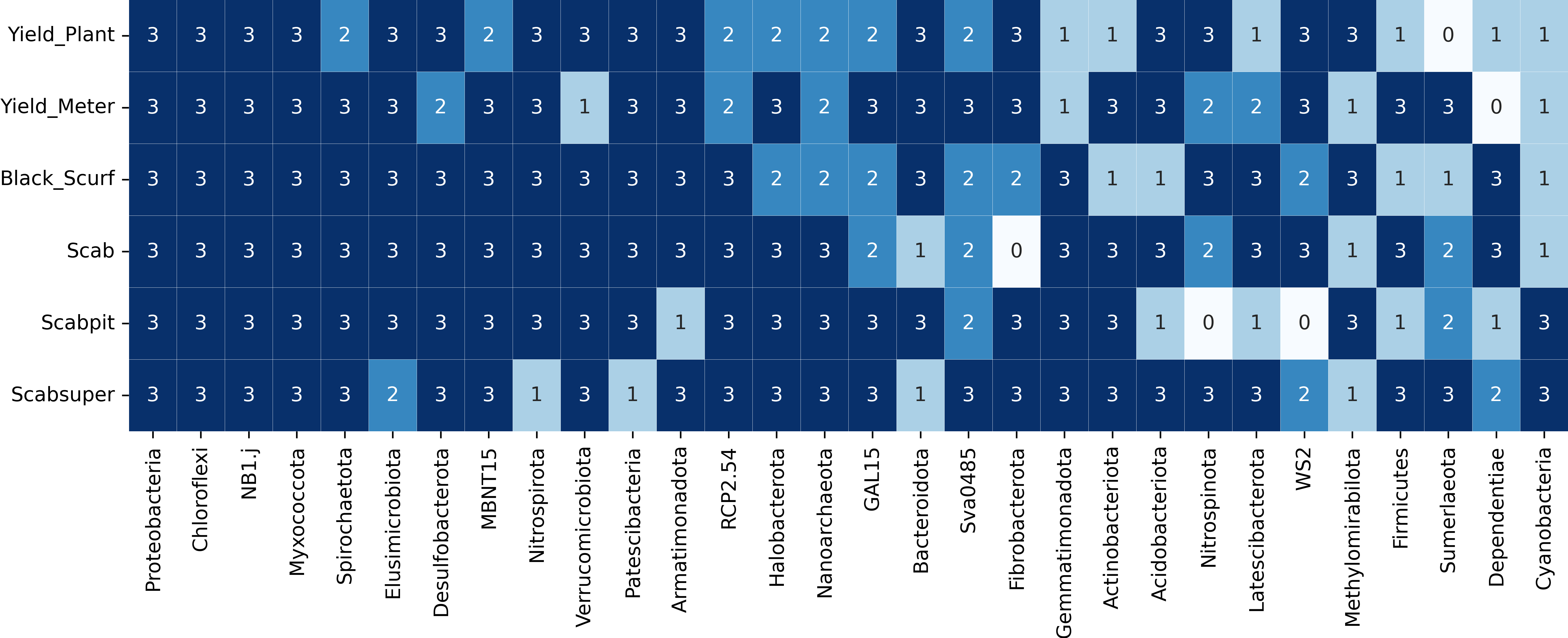}
\caption{Heatmap with scoring of OTUs (Phylum level) as important for each response. Y-axis shows the responses and x-axis shows the OTUs in Phylum level. We define a scoring value for each OTU based on the selection by ML strategy ($score=1$), network comparison strategy ($score=2$), both ($score=3$) or neither ($score=0$).}
\label{2-MLNetComi}
\end{figure} 

\begin{table}[]
\begin{tabular}{l|l|l}
Parameters& Description & Tuning Parameters \\
\hline
n\_estimators & The number of trees in the forest. & {[}100,200,500{]}          \\
min\_samples\_split                  & Minimum necessary number of samples to split an internal node                                                                                                                   & {[}8,10{]}                 \\
min\_samples\_leaf                   & The minimum needed the number of samples at a leaf node.                                                                                                                            & {[}3,4,5{]}                \\
max\_depth                           & The maximum depth of the tree                                                                                                                                                   & {[}80,90{]}                \\
criterion                            & The function used to evaluate a split's quality& ('gini','entropy')        
\end{tabular}
\caption[Parameters for Random Forest Model]{Different values for parameters for Random Forest Model.}\label{RFparameter}
\end{table}


\begin{figure}[h]
\centering
\includegraphics[scale=0.4]{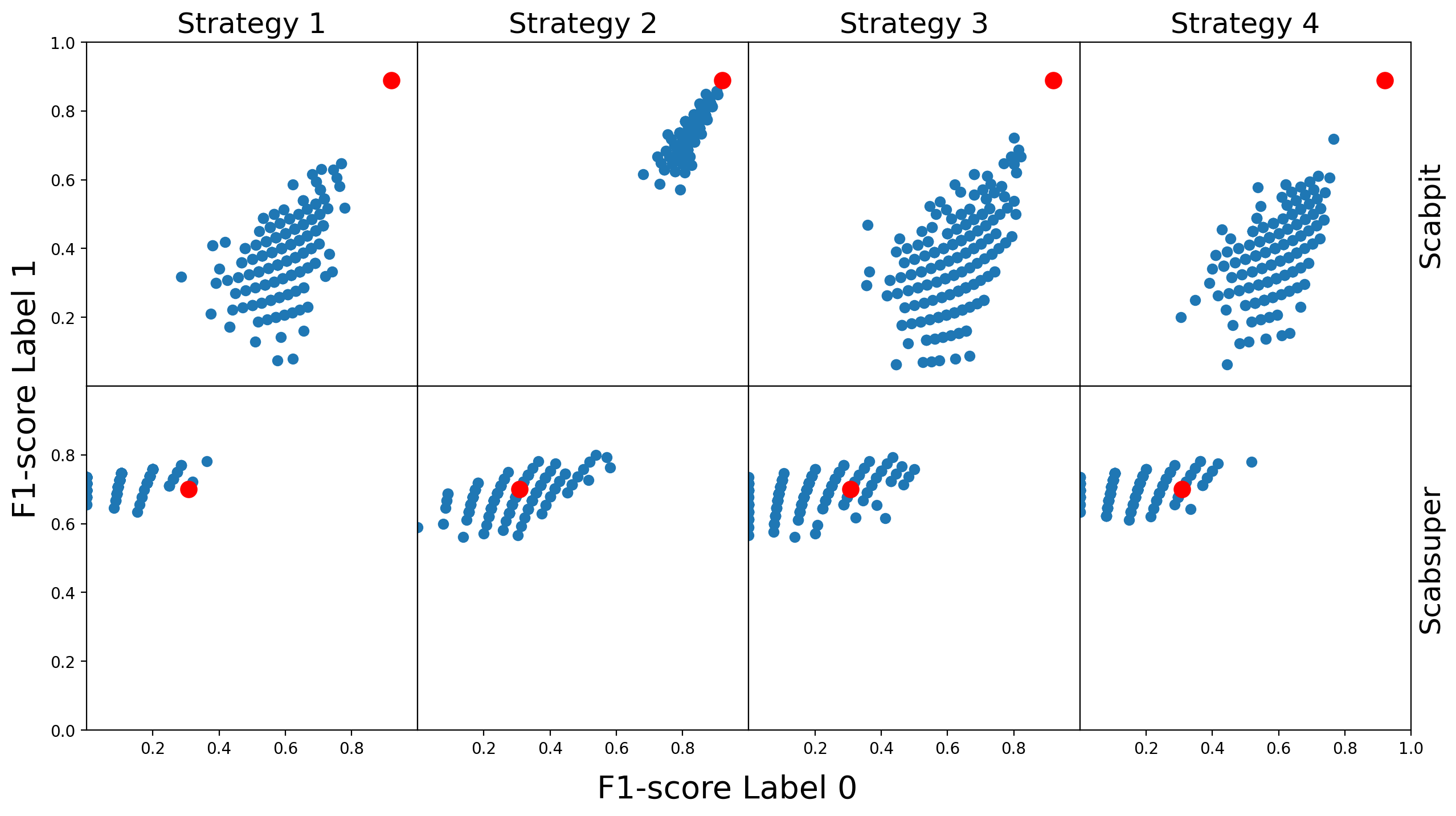}
\caption{Scatterplot for F1 scores in the non-diseased (Label $0$) group versus the F1 scores in the diseased (Label $1$) group by disease outcome (rows) and randomization strategy (columns). Each panel has 200 generated datasets (each blue point is a random dataset). The red point corresponds to the F1 scores when the model is fitted on the real data. These results are for OTUs at the Phylum level. There is clear predictive power for the pitted scab (Scabpit) as the red dot is in the upper right region of the panel, but none for the superficial scab (Scabsuper).}
\label{7-CompareToRandom}
\end{figure}

\begin{figure}[h]
\centering
\includegraphics[scale=0.4]{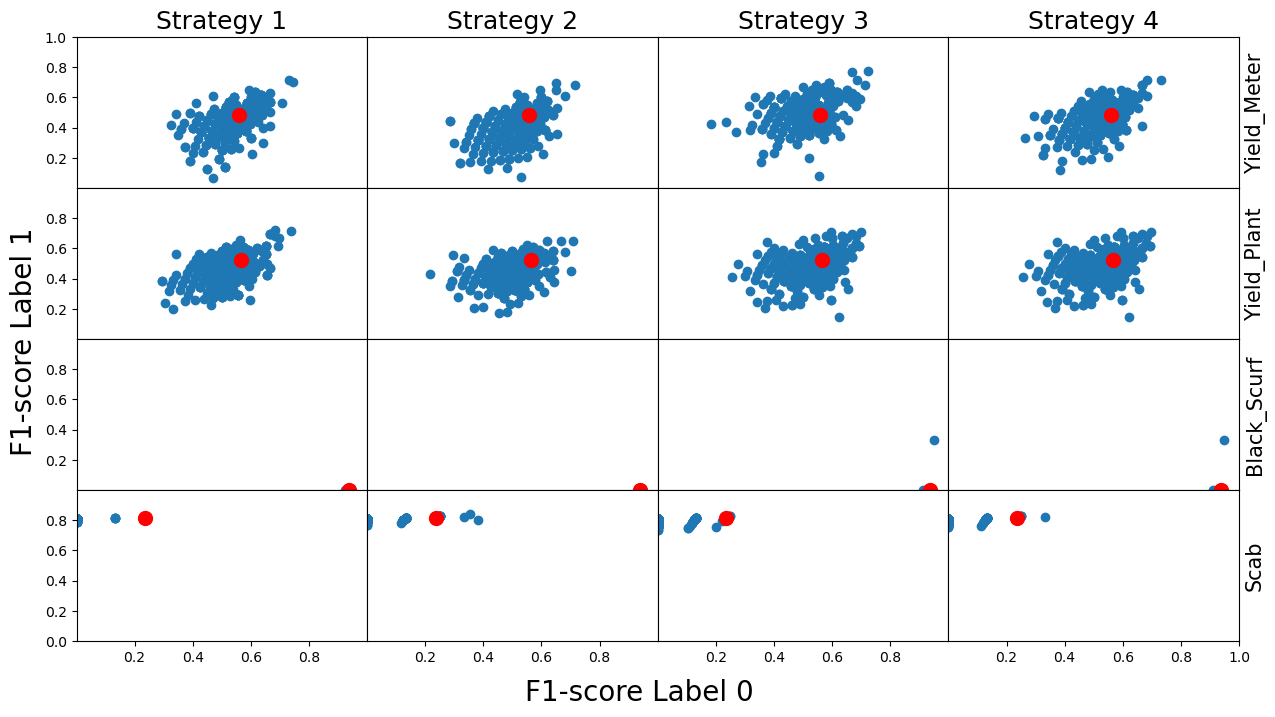}
\caption[Scatterplot for F1 scores for real and 200 generated datasets]{Scatterplot for F1 scores in the non-diseased (Label $0$) group versus the F1 scores in the diseased (Label $1$) group by disease outcome (rows) and randomization strategy (columns). Each panel has 200 generated datasets (each blue point is a random dataset), and in a red point, we should the results on the real data. These results are for data at the Phylum level. Fewer points in scab and black scurf diseases are due to extreme imbalance. For example, there are only 11 samples without disease (Label $0$) out of 46 testing samples in Scab which results in 11 label 0 samples and only 12 possible F1 values for both 0 and 1 classes, i.e, only 12 points. Since there are far more label 1 samples in Scab, the model is biased towards label 1 and tend to predict most testing samples as label 1, so the F1 values are clustered around 75-85\% for label 1. In this case, the points are more likely to stack on top of one another, thus leading to fewer points in the visualization. 
}
\label{S-CompareToRandom}
\end{figure}

\section{More power to predict pitted scab disease in microbiome data compared to random data}
\label{sec-random}

In order to study whether the microbiome data has predictive signal to classify samples corresponding to diseased or non-diseased potatoes, we can compare the performance of our models when fitted on the real data and when fitted on completely random data. Intuitively, if the data has predictive signal, models fitted on real data will dramatically outperform models fitted on random data.

Figure \ref{7-CompareToRandom} shows the scatterplot of F1 scores in the samples in one class (non-diseased potato: Label $0$) versus the F1 scores in the samples in the other class (diseased potato: Label $1$) for two disease outcomes (rows) and the four strategies to generate random data (columns). Strategies correspond to randomization of rows or columns, or generation of random abundance values (more details on the randomization strategies in Methods).
Each blue point in this scatterplot corresponds a random dataset, and the red point corresponds to the F1 scores when the model is fitted with the real data. Here, we only focus on microbiome data at the Phylum level.

For the pitted scab on the top row, the F1 scores for the original dataset (red dots) are on the upper right quadrant compared to points from random datasets (blue dots). This means that the F1 scores on the original data are higher than F1 scores on random data, and this, the original data has more predictive power than random data. However, for the superficial scab on the bottom row, many random  datasets show better performance (higher F1 score values) than the real data, and thus, the real data does not contain much information to predict this disease. 
Other diseases as well as yield outcomes also underperform compared to random data as shown in Figure \ref{S-CompareToRandom} in the Supplementary Material. We highlight that pitted scab is the most severe case of the potato disease common scab, and possibly a stronger signal of the soil microbial community composition and corresponding disease-suppressiveness than other measures of the disease such as superficial scab.

\subsection{Predictive Power of Microbiome Data: Comparison to Random Data}

One of the key questions we want to address in this study is whether the microbiome data has predictive power to infer disease or yield outcomes.
To address this question more formally, we devise 
a simulation scheme in which we simulate random microbiome data to train our RF model and compare the prediction potential with the trained RF model on real microbiome data. If there is indeed a signal to predict yield or disease outcomes within the real data, then we expect the model trained with the real data to dramatically outperform the model trained with random microbiome datasets.
We use four strategies to generate random datasets:
\begin{description}
   \item[Strategy 1:] Random matrix with values between 0 and 1, normalized so that each row has a sum of 1. We use the response vector from the original dataset. 
   \item[Strategy 2:] Real microbiome matrix where each entry that is greater than zero is replaced by a random number between 0 and 1. This strategy preserves the sparsity in the real data as zero entries remain zero. Rows are normalized so that the sum equals 1. We use the response vector from the original dataset.
    \item[Strategy 3:] Real microbiome matrix is used with a permuted response vector.
    \item[Strategy 4:] Rows in the real microbiome matrix are permuted, and the response vector is unchanged.
\end{description}
For each strategy, we generate $N=200$ random datasets to train the RF model.
Let $F_i$ for $i = 1, ..., N$, denote the weighted F1 scores for the randomized data and let $F_{original}$ be the weighted F1 score on the real data.
Let $X = \sum_{i=1}^N \mathbb{I}(F_i>F_{original})$
be the number of random datasets that perform better than the read data. We use the exceeding value (EV)  
$EV=\frac{X}{N}$ as test statistic to test the null hypothesis that the real data performs just as random data in the prediction on disease outcomes. The EV denotes the percentage of random datasets that perform better than the real microbiome data out of $N$ random datasets. If $EV<\alpha$, we reject the null hypothesis and conclude that there is predictive power (beyond random) on the real microbiome data. Here, we set $\alpha=0.05$.
\begin{figure}[ht!]
\centering
\includegraphics[scale=0.35]{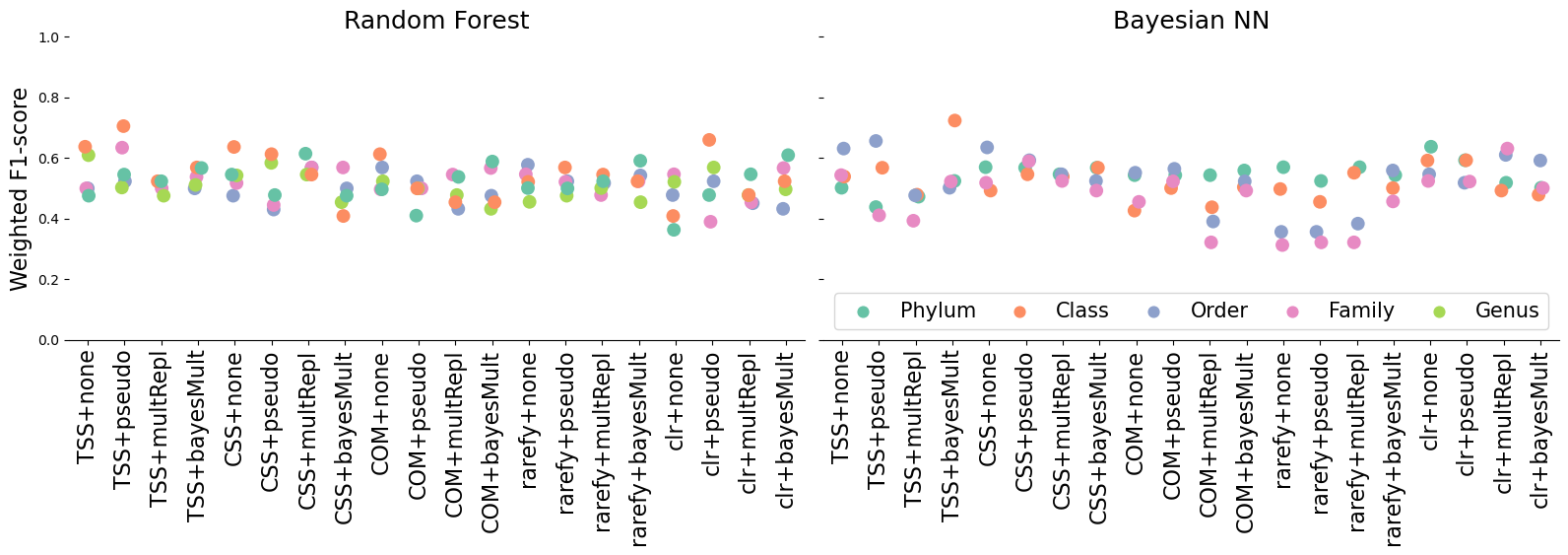}
\caption[Weighted F1 scores for yield by plant under the 20 normalization/zero replacement strategies]{Weighted F1 scores (y-axis) for Random Forest and Bayesian Neural Network (NN) models for yield by plant under the 20 normalization/zero replacement strategies (x-axis). The lack of pattern prevents us from making recommendations of optimal strategies for microbiome OTU data. We can conclude, however, that taxonomic levels, normalization and zero replacement strategies have an effect on the prediction accuracy of the models as evidenced by the broad range displayed by the points.}
\label{S-RF-BNN-Normalized-Yield-Plant}
\end{figure}

\begin{figure}[ht!]
\centering
\includegraphics[scale=0.35]{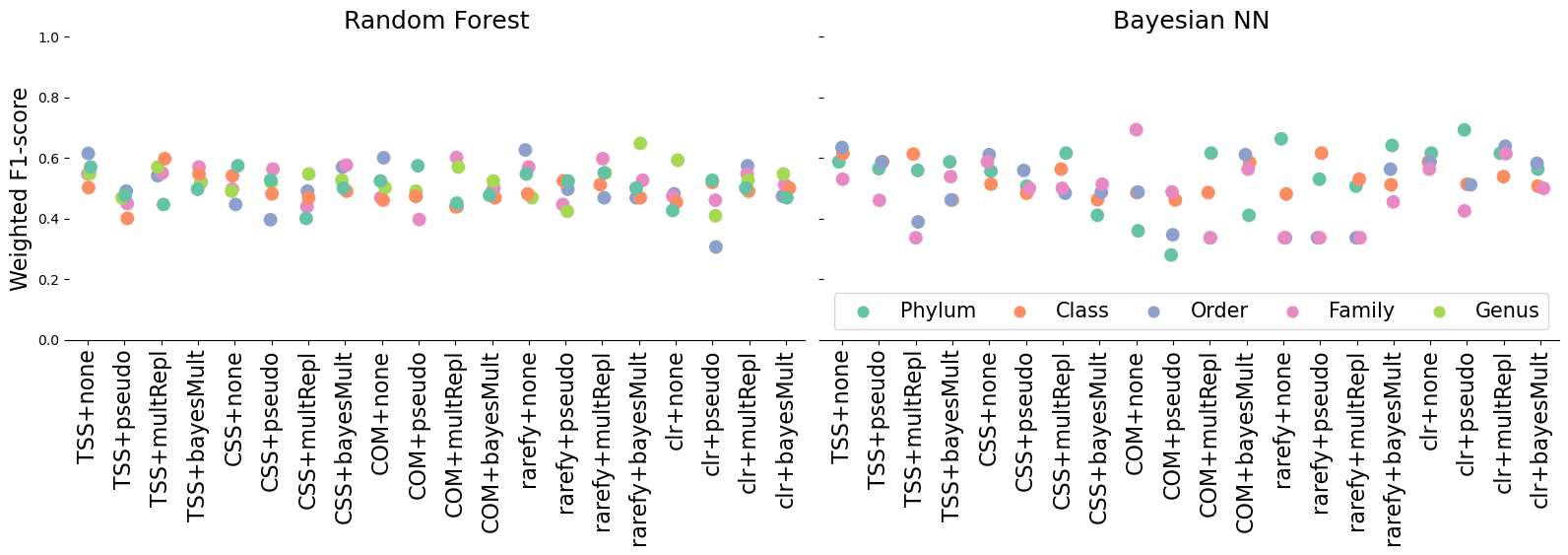}
\caption[Weighted F1 scores for yield by meter  under the 20 normalization/zero replacement strategies]{Weighted F1 scores (y-axis) for Random Forest and Bayesian Neural Network (NN) models for yield by meter  under the 20 normalization/zero replacement strategies (x-axis). The lack of pattern prevents us from making recommendations of optimal strategies for microbiome OTU data. We can conclude, however, that taxonomic levels, normalization and zero replacement strategies have an effect on the prediction accuracy of the models as evidenced by the broad range displayed by the points.}
\label{S-RF-BNN-Normalized-Yield-Meter}
\end{figure}
\begin{figure}[ht!]
\centering
\includegraphics[scale=0.35]{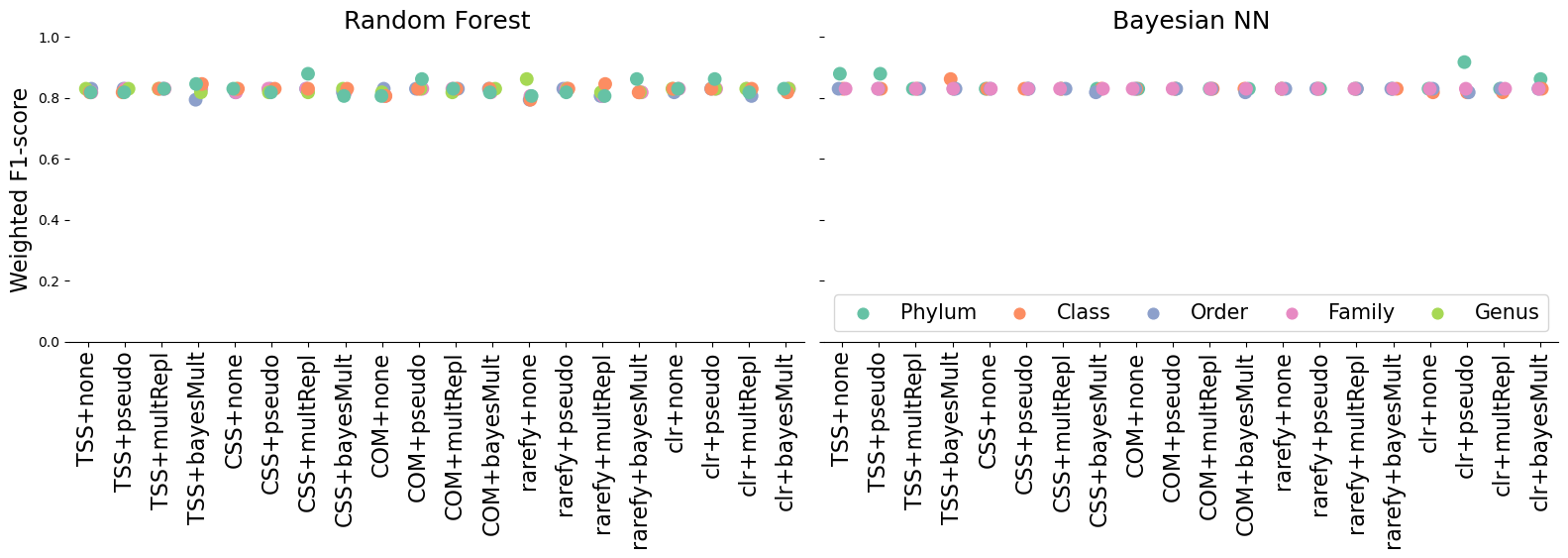}
\caption[Weighted F1 scores for black scurf disease under the 20 normalization/zero replacement strategies]{Weighted F1 scores (y-axis) for Random Forest and Bayesian Neural Network (NN) models for black scurf disease under the 20 normalization/zero replacement strategies (x-axis). The lack of pattern prevents us from making recommendations of optimal strategies for microbiome OTU data. We can conclude, however, that taxonomic levels, normalization and zero replacement strategies have an effect on the prediction accuracy of the models as evidenced by the broad range displayed by the points.}
\label{S-RF-BNN-Normalized-Black}
\end{figure}

\begin{figure}[ht!]
\centering
\includegraphics[scale=0.35]{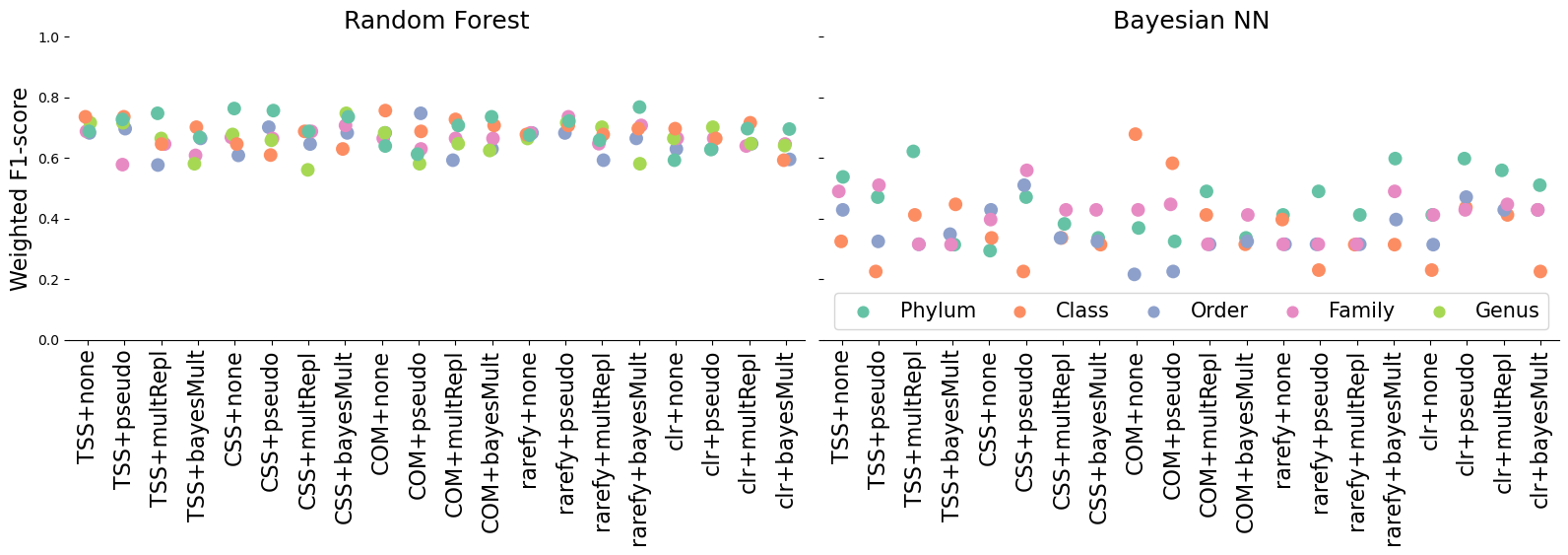}
\caption[Weighted F1 scores for scab disease under the 20 normalization/zero replacement strategies]{Weighted F1 scores (y-axis) for Random Forest and Bayesian Neural Network (NN) models for scab disease under the 20 normalization/zero replacement strategies (x-axis). The lack of pattern prevents us from making recommendations of optimal strategies for microbiome OTU data. We can conclude, however, that taxonomic levels, normalization and zero replacement strategies have an effect on the prediction accuracy of the models as evidenced by the broad range displayed by the points.}
\label{S-RF-BNN-Normalized-Scab}
\end{figure}
\begin{figure}[ht!]
\centering
\includegraphics[scale=0.35]{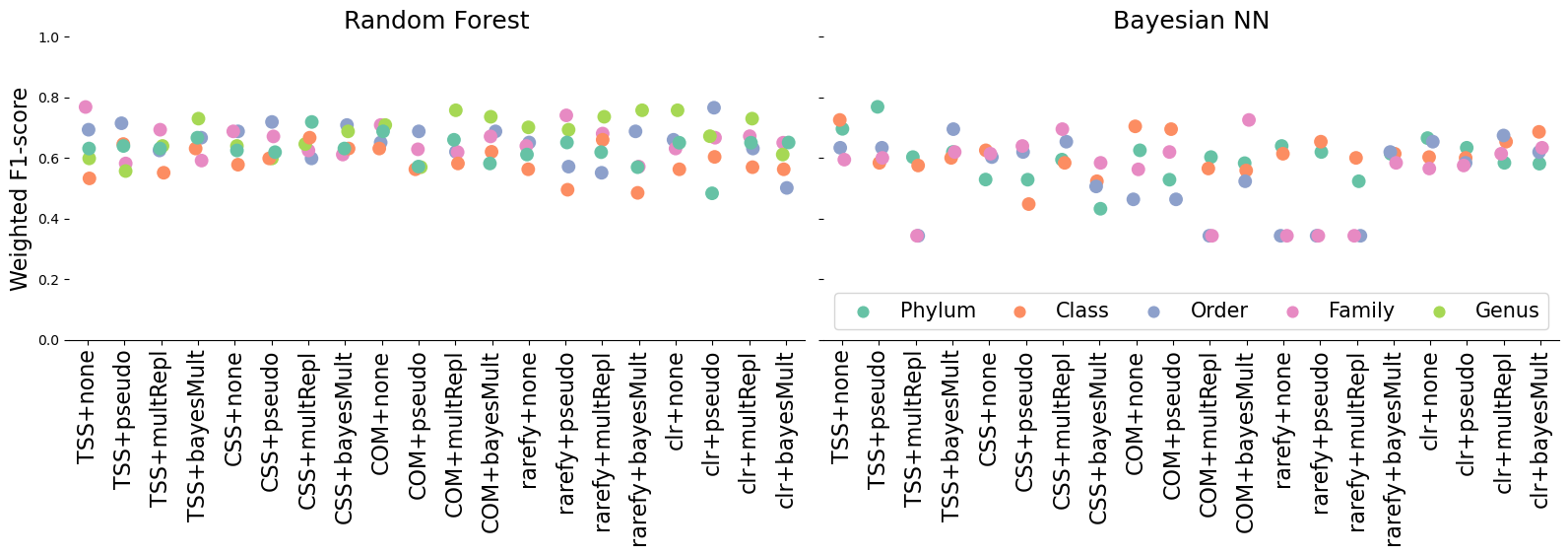}
\caption[Weighted F1 scores for superficial scab (Scabsuper) disease under the 20 normalization/zero replacement strategies]{Weighted F1 scores (y-axis) for Random Forest and Bayesian Neural Network (NN) models for superficial scab (Scabsuper) disease under the 20 normalization/zero replacement strategies (x-axis). The lack of pattern prevents us from making recommendations of optimal strategies for microbiome OTU data. We can conclude, however, that taxonomic levels, normalization and zero replacement strategies have an effect on the prediction accuracy of the models as evidenced by the broad range displayed by the points.}
\label{S-RF-BNN-Normalized-Scabsuper}
\end{figure}

\begin{figure}[ht!]
\centering
\includegraphics[scale=0.35]{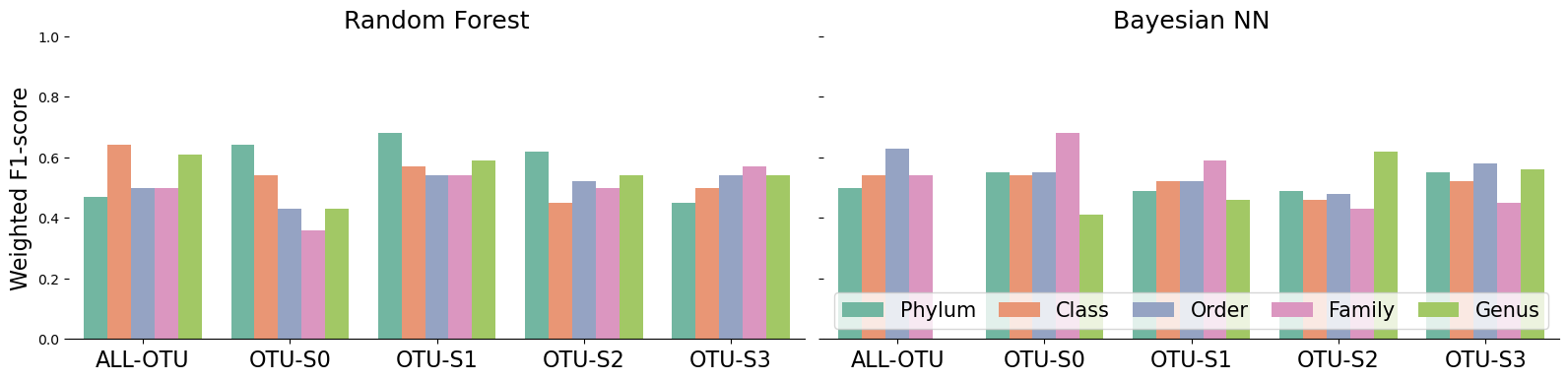}
\caption[Weighted F1 scores for yield by plant and selected features by different strategies]{Weighted F1 scores (y-axis) by Random Forest and Bayesian Neural Network (NN) models for yield by plant by feature selection strategy (x-axis) including all OTUs (All-OTU), OTUs selected by the ML method (OTU-S1), the network comparison method (OTU-S2), both methods (OTU-S3), or neither method (OTU-S0).}
\label{S-RF-BNN-SelectedFeatures-Yield-Plant}
\end{figure}
\begin{figure}[ht!]
\centering
\includegraphics[scale=0.35]{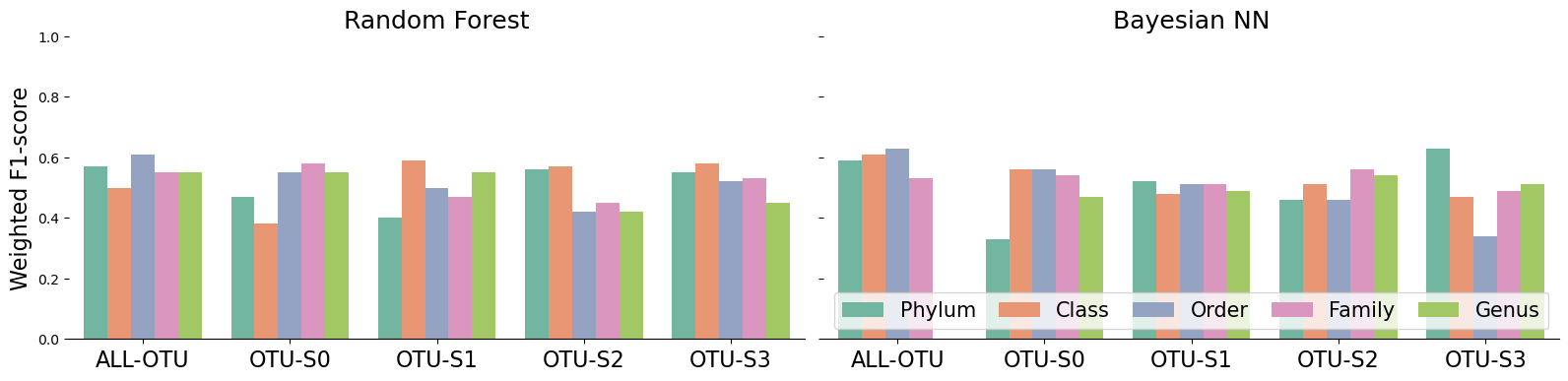}
\caption[Weighted F1 scores for yield by meter and selected features by different strategies]{Weighted F1 scores (y-axis) by Random Forest and Bayesian Neural Network (NN) models for yield by meter  by feature selection strategy (x-axis) including all OTUs (All-OTU), OTUs selected by the ML method (OTU-S1), the network comparison method (OTU-S2), both methods (OTU-S3), or neither method (OTU-S0).}
\label{S-RF-BNN-SelectedFeatures-Yield-Meter}
\end{figure}
\begin{figure}[ht!]
\centering
\includegraphics[scale=0.35]{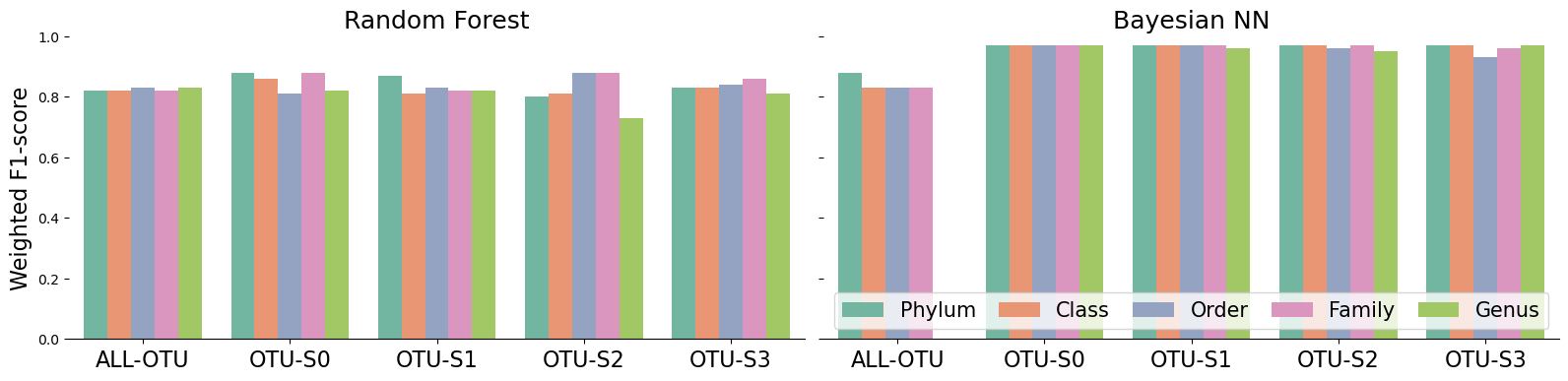}
\caption[Weighted F1 scores for black scurf and selected features by different strategies]{Weighted F1 scores (y-axis) by Random Forest and Bayesian Neural Network (NN) models for black scurf disease by feature selection strategy (x-axis) including all OTUs (All-OTU), OTUs selected by the ML method (OTU-S1), the network comparison method (OTU-S2), both methods (OTU-S3), or neither method (OTU-S0).}
\label{S-RF-BNN-SelectedFeatures-Black}
\end{figure}
\begin{figure}[ht!]
\centering
\includegraphics[scale=0.35]{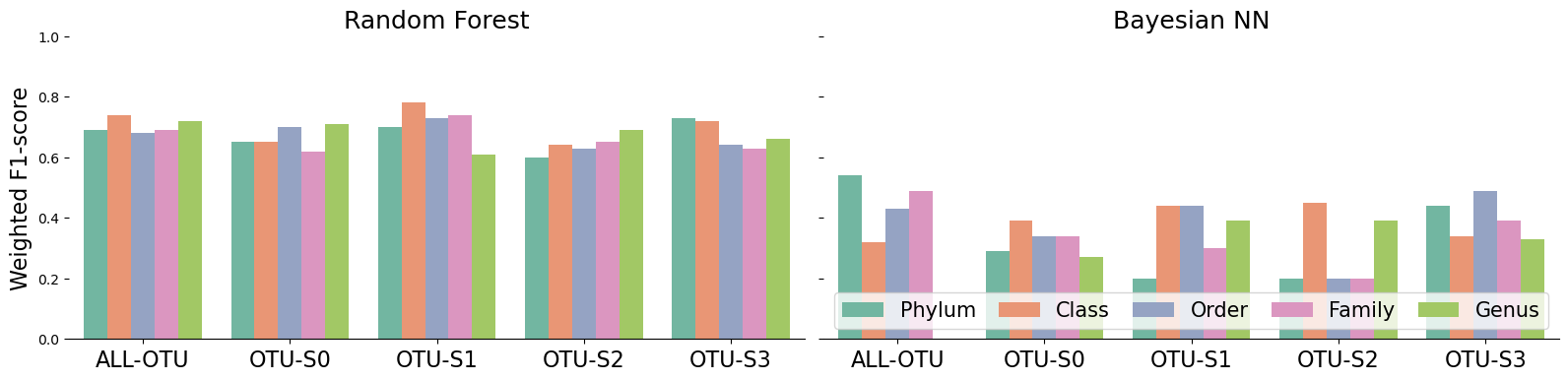}
\caption[Weighted F1 scores for scab disease and selected features by different strategies]{Weighted F1 scores (y-axis) by Random Forest and Bayesian Neural Network (NN) models for scab disease by feature selection strategy (x-axis) including all OTUs (All-OTU), OTUs selected by the ML method (OTU-S1), the network comparison method (OTU-S2), both methods (OTU-S3), or neither method (OTU-S0).}
\label{S-RF-BNN-SelectedFeatures-Scab}
\end{figure}
\begin{figure}[ht!]
\centering
\includegraphics[scale=0.35]{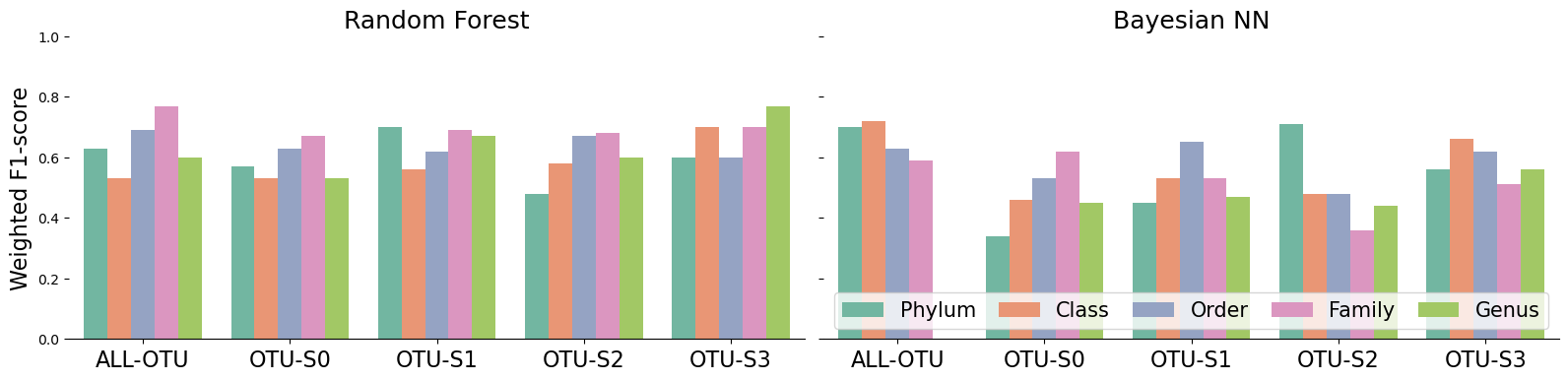}
\caption[Weighted F1 scores for superficial scab (Scabsuper) disease and selected features by different strategies]{Weighted F1 scores (y-axis) by Random Forest and Bayesian Neural Network (NN) models for superficial scab (Scabsuper) disease by feature selection strategy (x-axis) including all OTUs (All-OTU), OTUs selected by the ML method (OTU-S1), the network comparison method (OTU-S2), both methods (OTU-S3), or neither method (OTU-S0).}
\label{S-RF-BNN-SelectedFeatures-Scabsuper}
\end{figure}

\begin{figure}[ht!]
\centering
\includegraphics[scale=0.21]{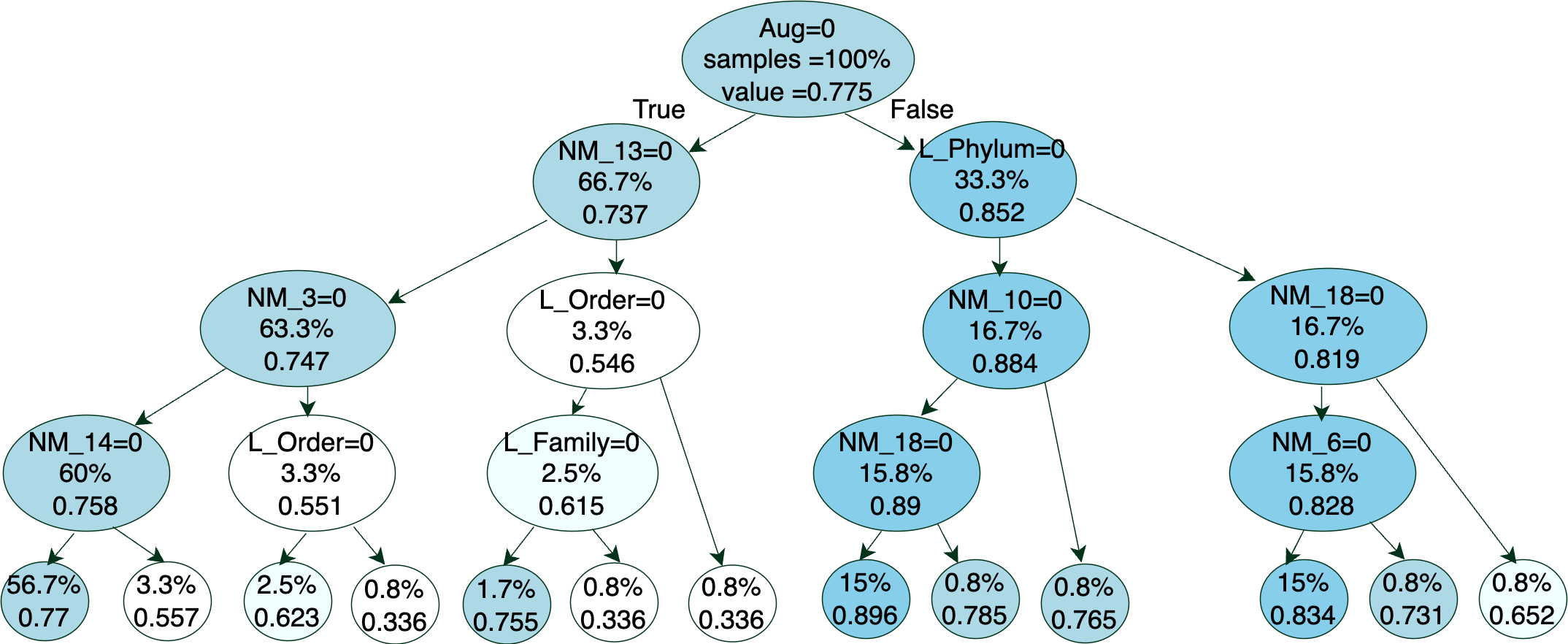}
\caption{Full model selection  decision tree with a maximum depth of 4 summarized the results of Bayesian neural network models on pitted scab disease (Scabpit) disease. When the condition at a node is true, we follow the branch on the left, and when the condition is false, we follow the branch on the right. The percentage of the data preprocessing options and mean of weighted F1 scores are shown in each nodes. The darker the red, the higher the weighted F1 score.}
\label{13-DT-BNN-Scabpit}
\end{figure}

\begin{figure}[ht!]
\centering
\includegraphics[scale=0.3]{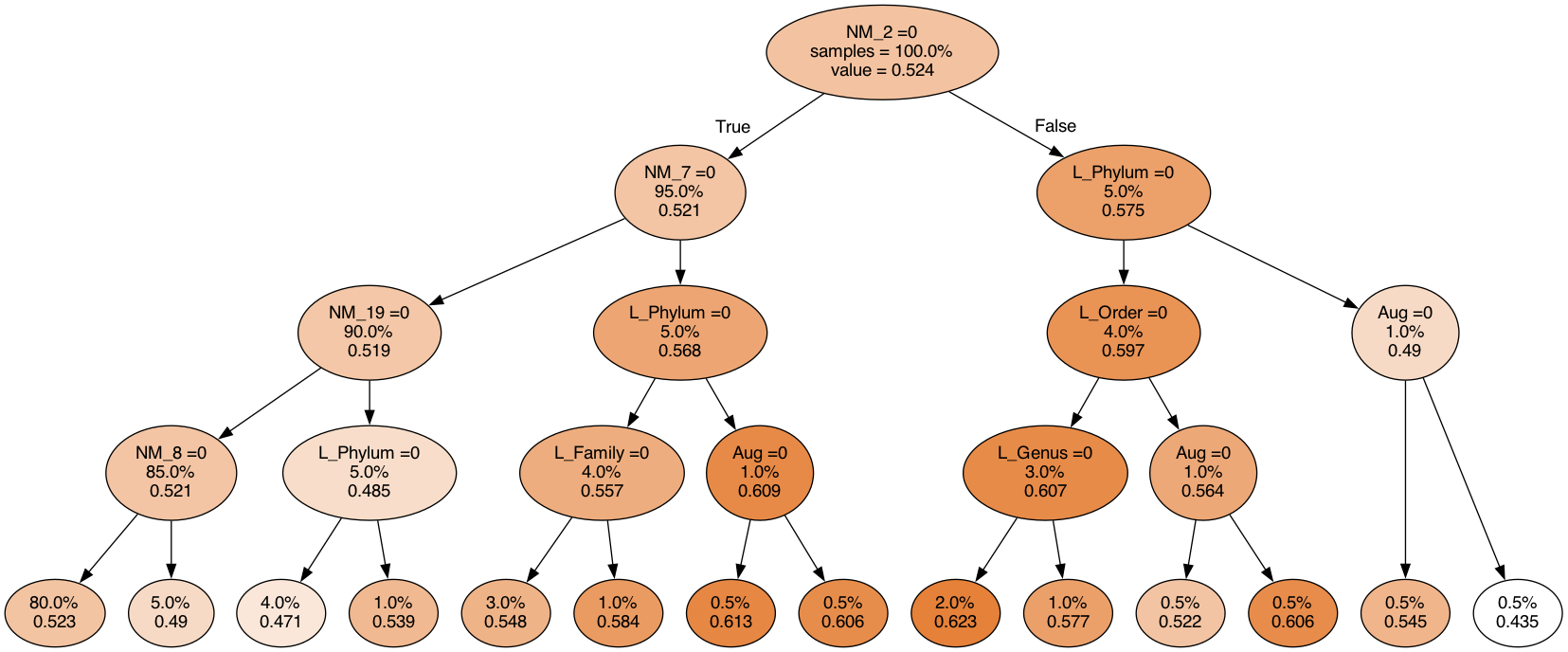}
\caption[Full model selection decision tree summarizing the results of Random Forest models on yield by plant]{Full model selection decision tree with a maximum depth of 4 summarizing the results of Random Forest models on yield by plant. When the condition at a node is true, we follow the branch on the left, and when the condition is false, we follow the branch on the right. The percentage of data preprocessing options and mean of weighted F1 scores are shown in each nodes.}
\label{S-DT-RF-Yield-Plant}
\end{figure}
\begin{figure}[ht!]
\centering
\includegraphics[scale=0.3]{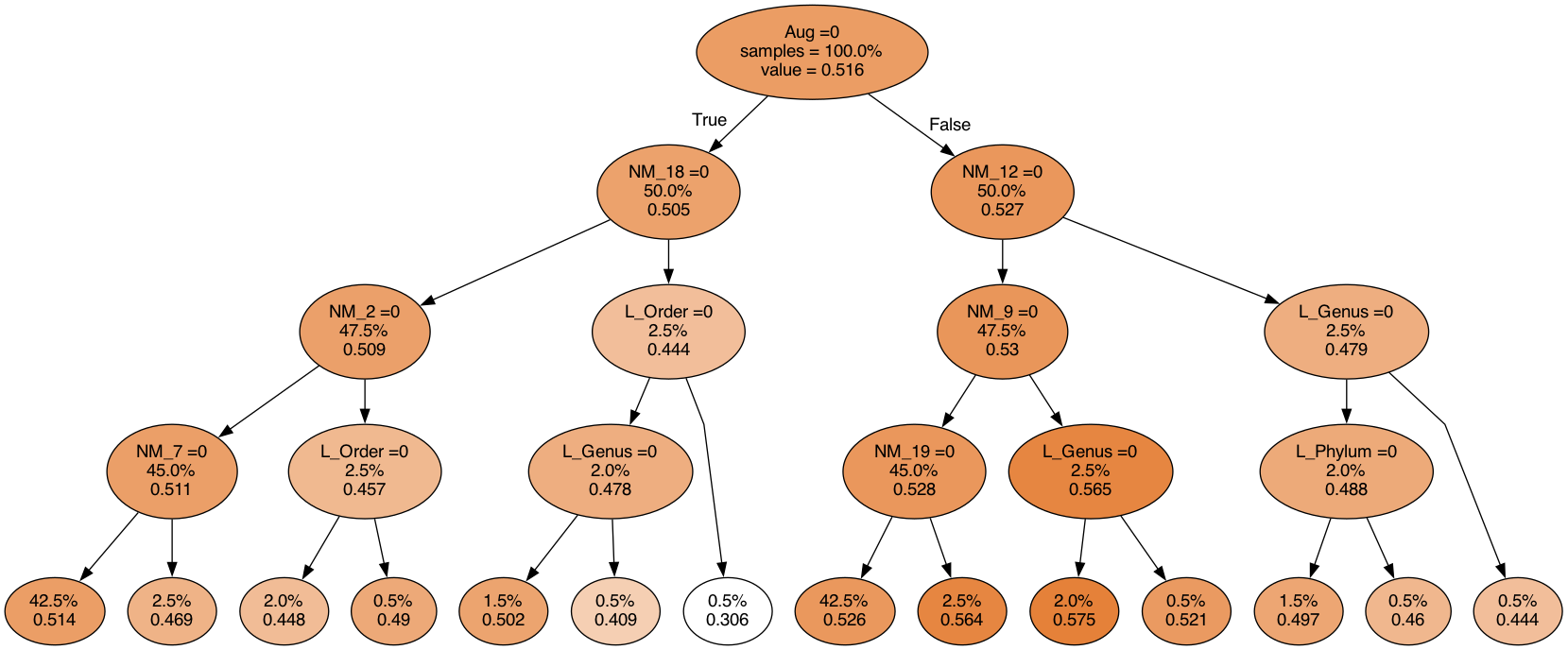}
\caption[Full model selection decision tree  summarizing the results of Random Forest models on yield by meter]{Full model selection  decision tree with a maximum depth of 4 summarizing the results of Random Forest models on yield by meter. When the condition at a node is true, we follow the branch on the left, and when the condition is false, we follow the branch on the right. The percentage of data preprocessing options and mean of weighted F1 scores are shown in each nodes.}
\label{S-DT-RF-Yield-Meter}
\end{figure}
\begin{figure}[ht!]
\centering
\includegraphics[scale=0.3]{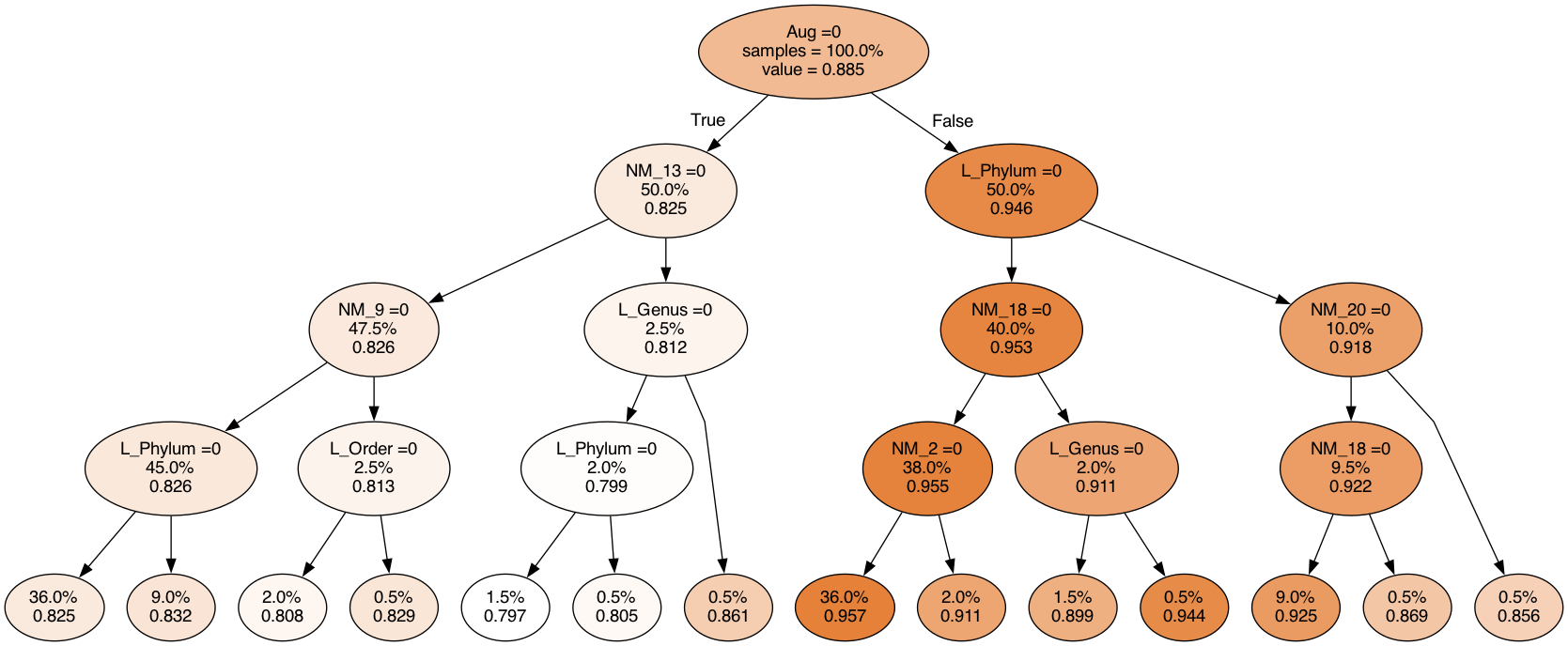}
\caption[Full model selection decision tree  summarizing the results of Random Forest models on black scurf]{Full model selection  decision tree with a maximum depth of 4 summarizing the results of Random Forest models on black scurf disease. When the condition at a node is true, we follow the branch on the left, and when the condition is false, we follow the branch on the right. The percentage of data preprocessing options and mean of weighted F1 scores are shown in each nodes.}
\label{S-DT-RF-Black}
\end{figure}
\begin{figure}[ht!]
\centering
\includegraphics[scale=0.3]{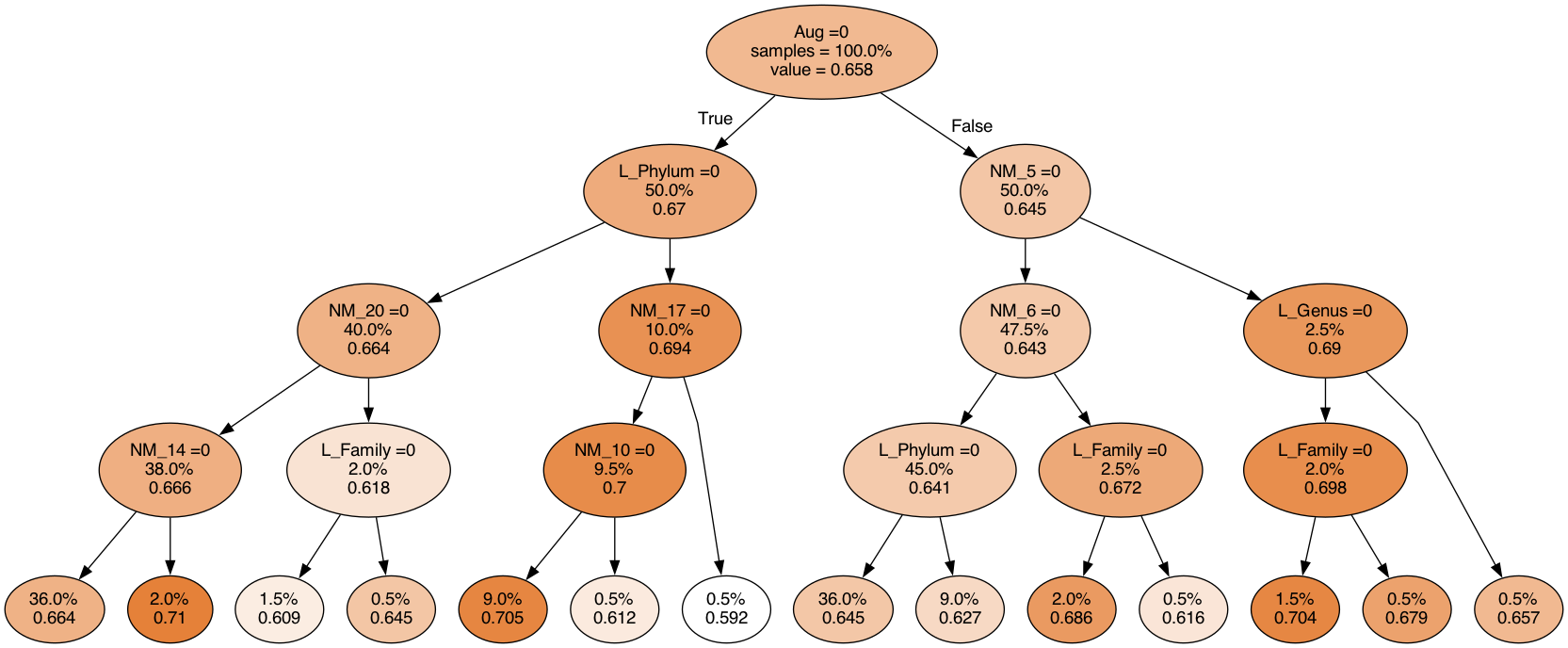}
\caption[Full model selection decision tree summarizing the results of Random Forest models on scab disease]{Full model selection  decision tree with a maximum depth of 4 summarizing the results of Random Forest models on scab disease. When the condition at a node is true, we follow the branch on the left, and when the condition is false, we follow the branch on the right. The percentage of data preprocessing options and mean of weighted F1 scores are shown in each nodes.}
\label{S-DT-RF-Scab}
\end{figure}
\begin{figure}[ht!]
\centering
\includegraphics[scale=0.3]{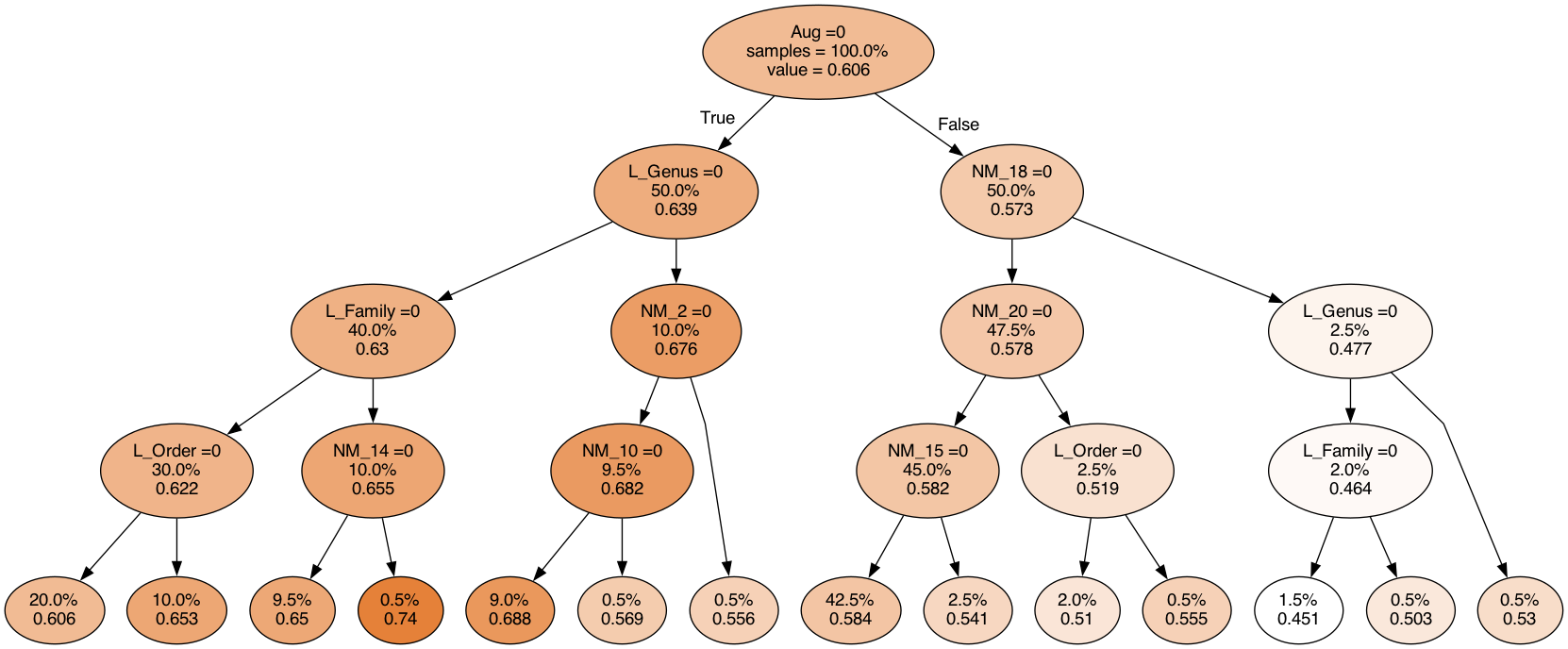}
\caption[Full model selection decision tree summarizing the results of Random Forest models on superficial scab (Scabsuper) disease]{Full model selection  decision tree with a maximum depth of 4 summarizing the results of Random Forest models on superficial scab (Scabsuper) disease. When the condition at a node is true, we follow the branch on the left, and when the condition is false, we follow the branch on the right. The percentage of data preprocessing options and mean of weighted F1 scores are shown in each nodes.}
\label{S-DT-RF-Scabsuper}
\end{figure}

\begin{figure}[ht!]
\centering
\includegraphics[scale=0.3]{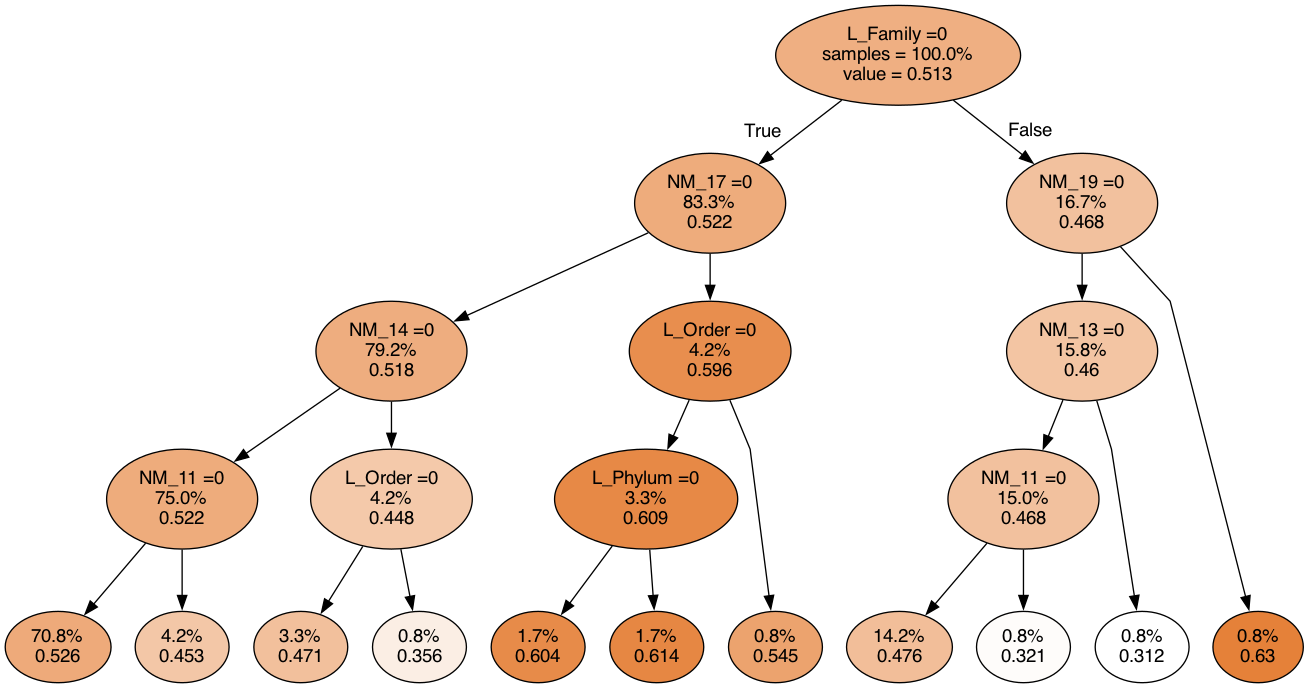}
\caption[Full model selection decision tree summarizing the results of Bayesian Neural Network on yield plant ]{Full model selection  decision tree with a maximum depth of 4 summarized the results of Bayesian Neural Network models on yield by plant. When the condition at a node is true, we follow the branch on the left, and when the condition is false, we follow the branch on the right. The percentage of data preprocessing options and mean of weighted F1 scores are shown in each nodes.}
\label{S-DT-BNN-Yield-Plant}
\end{figure}
\begin{figure}[ht!]
\centering
\includegraphics[scale=0.3]{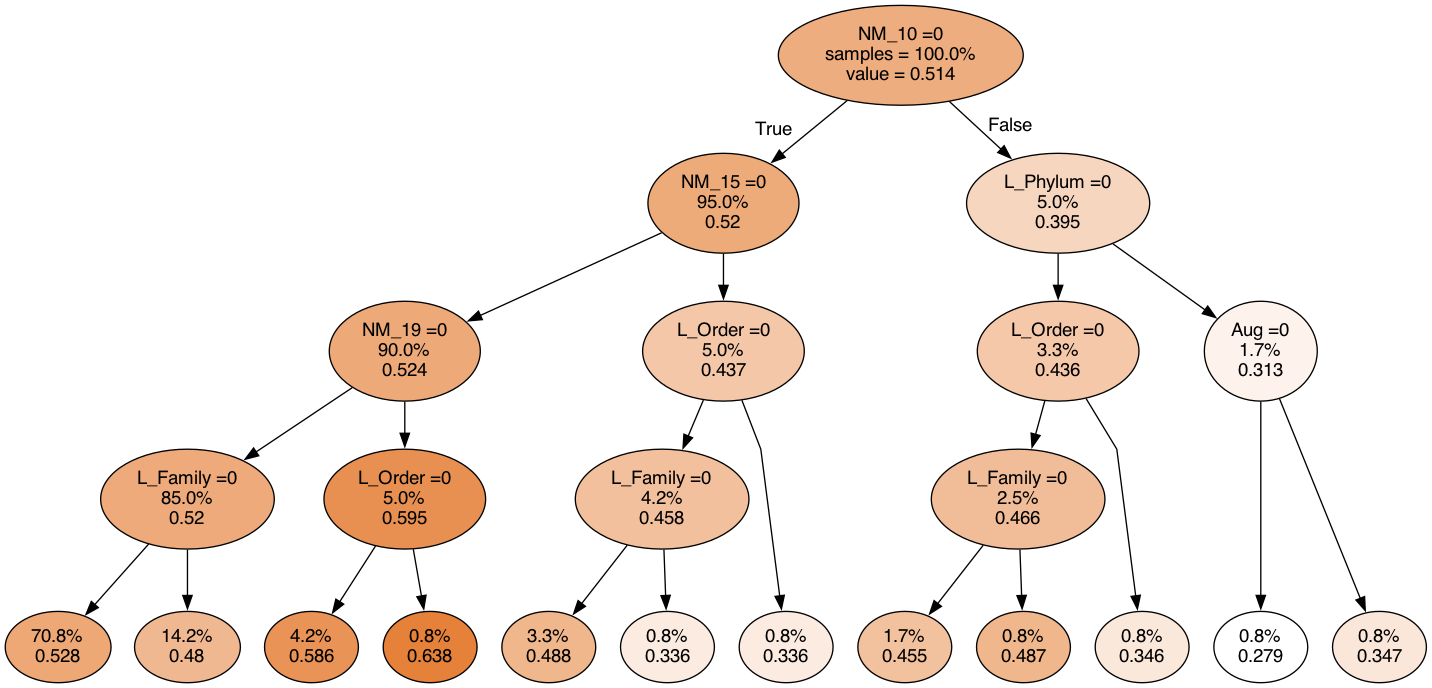}
\caption[Full model selection decision tree  summarizing the results of Bayesian Neural Network on yield by meter]{Full model selection  decision tree with a maximum depth of 4 summarized the results of Bayesian Neural Network models on yield by meter. When the condition at a node is true, we follow the branch on the left, and when the condition is false, we follow the branch on the right. The percentage of data preprocessing options and mean of weighted F1 scores are shown in each nodes.}
\label{S-DT-BNN-Yield-Meter}
\end{figure}
\begin{figure}[ht!]
\centering
\includegraphics[scale=0.3]{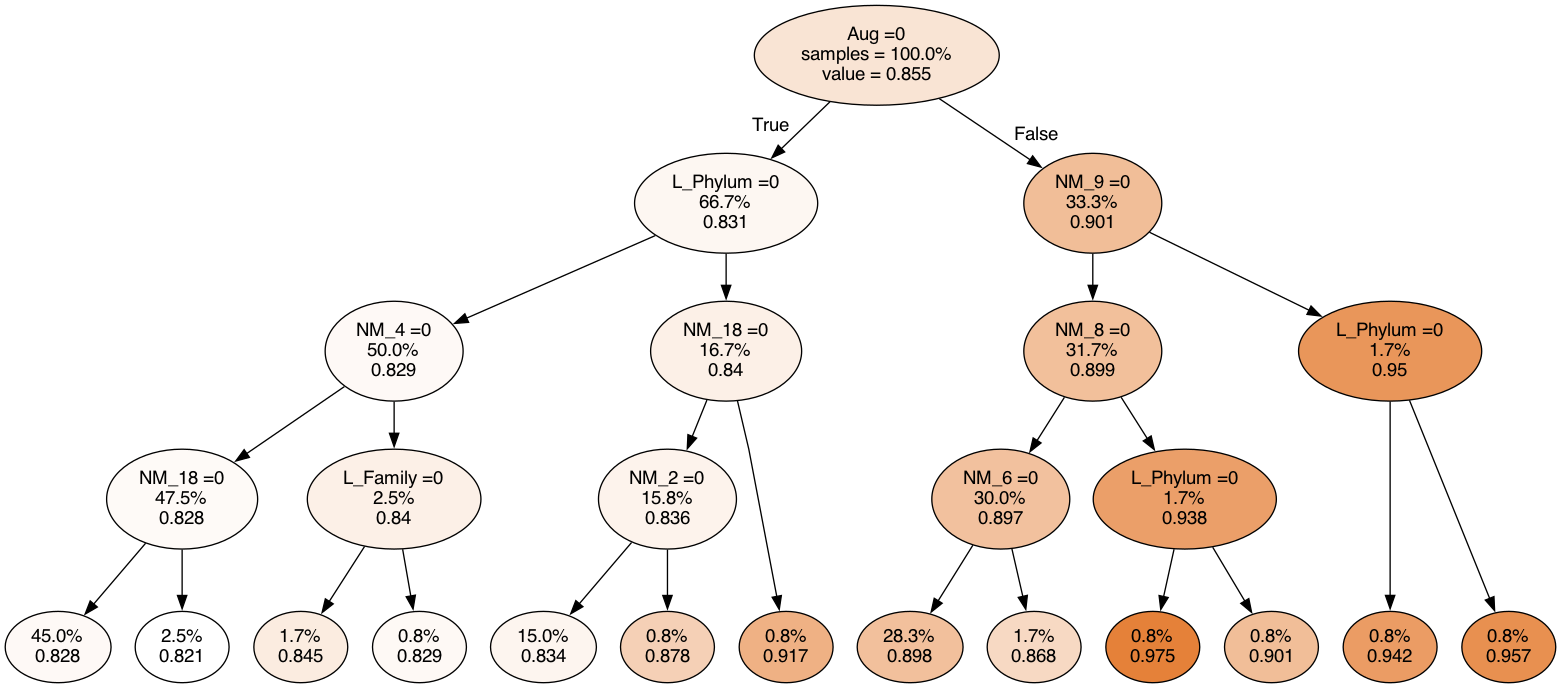}
\caption[Full model selection decision tree summarizing the results of Bayesian Neural Network on black scurf]{Full model selection  decision tree with a maximum depth of 4 summarized the results of Bayesian Neural Network models on black scurf disease. When the condition at a node is true, we follow the branch on the left, and when the condition is false, we follow the branch on the right. The percentage of data preprocessing options and mean of weighted F1 scores are shown in each nodes.}
\label{S-DT-BNN-Black}
\end{figure}
\begin{figure}[ht!]
\centering
\includegraphics[scale=0.3]{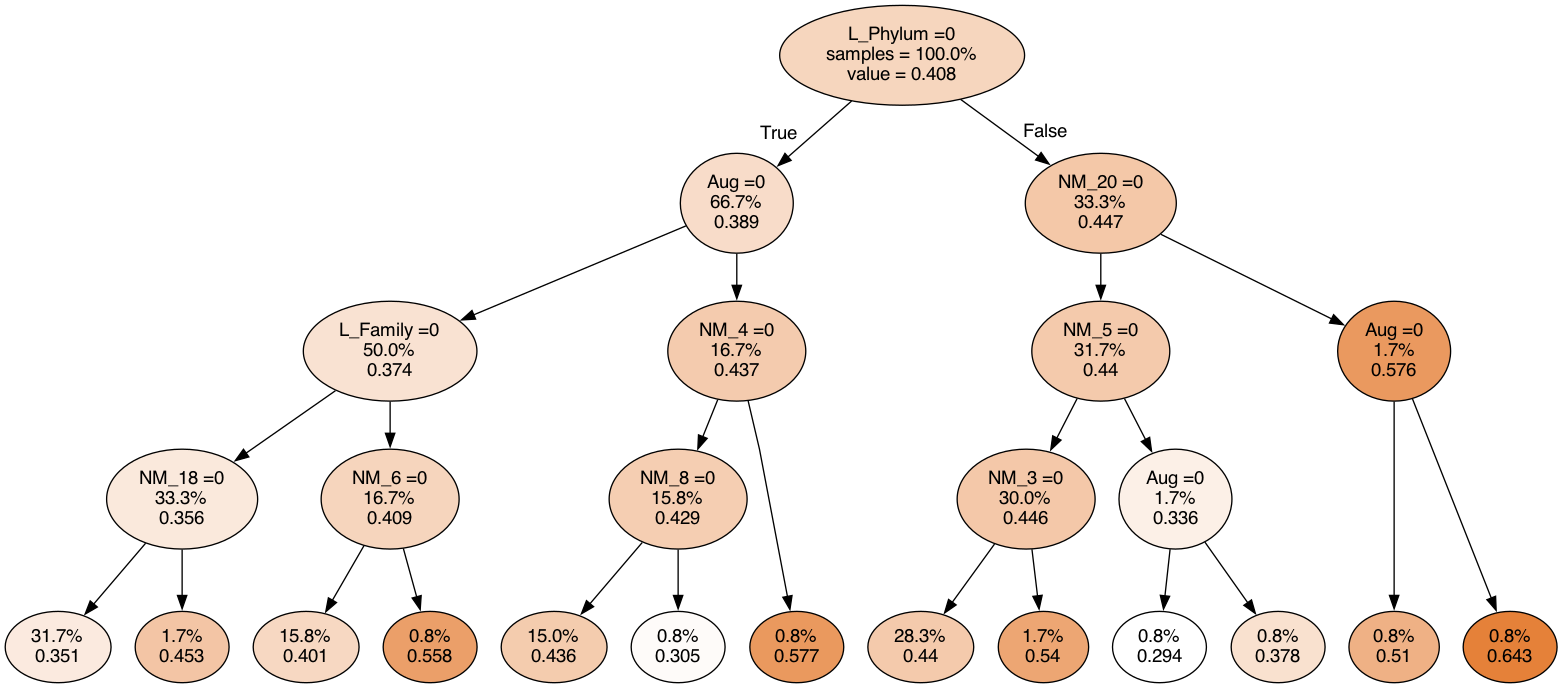}
\caption[Full model selection decision tree  summarizing the results of Bayesian Neural Network on scab disease]{Full model selection  decision tree with a maximum depth of 4 summarized the results of Bayesian Neural Network models on scab disease. When the condition at a node is true, we follow the branch on the left, and when the condition is false, we follow the branch on the right. The percentage of data preprocessing options and mean of weighted F1 scores are shown in each nodes.}
\label{S-DT-BNN-Scab}
\end{figure}
\begin{figure}[ht!]
\centering
\includegraphics[scale=0.3]{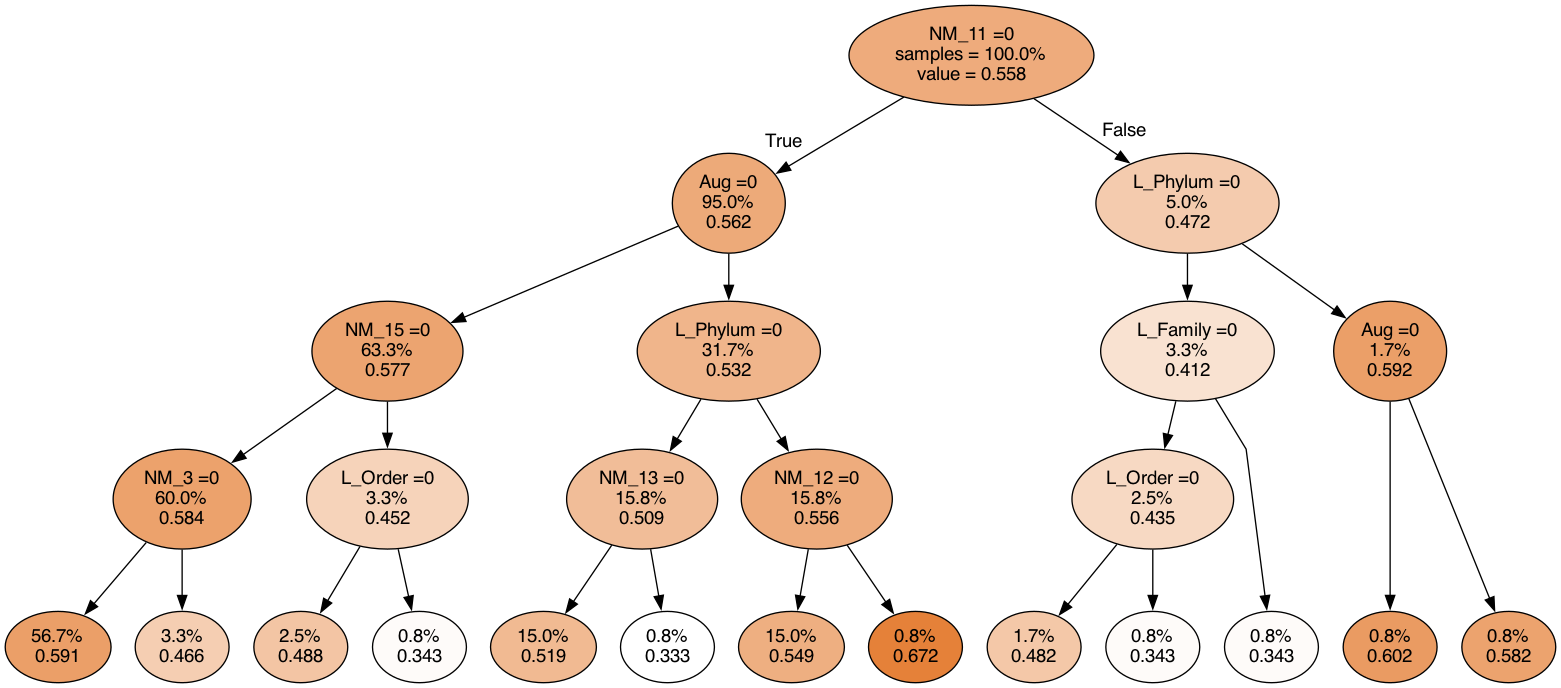}
\caption[Full model selection decision tree summarizing the results of Bayesian Neural Network on superficial scab (Scabsuper) disease]{Full model selection  decision tree with a maximum depth of 4 summarized the results of Bayesian Neural Network models on superficial scab (Scabsuper) disease. When the condition at a node is true, we follow the branch on the left, and when the condition is false, we follow the branch on the right. The percentage of data preprocessing options and mean of weighted F1 scores are shown in each nodes.}
\label{S-DT-BNN-Scabsuper}
\end{figure}

\begin{figure}[h]
\centering
\includegraphics[scale=0.5]{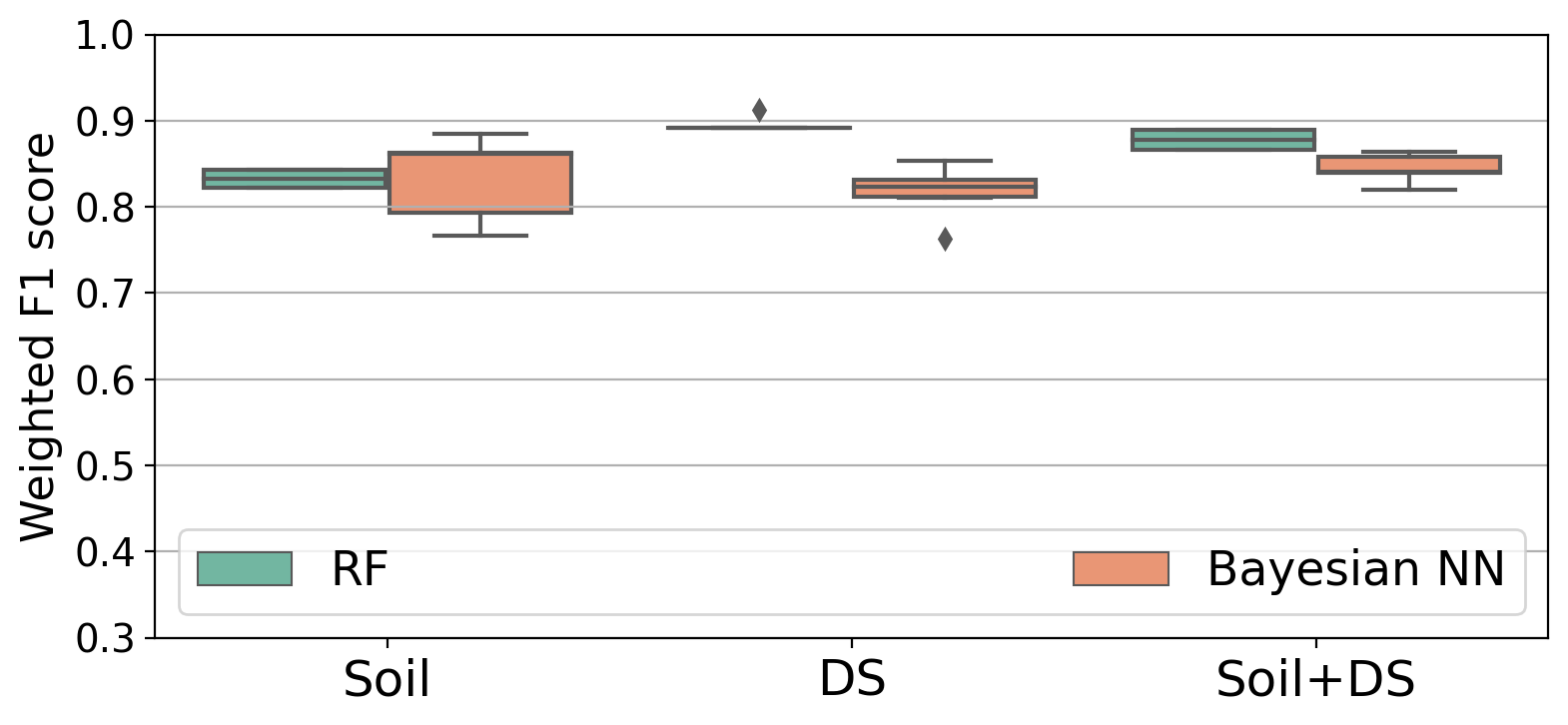}
\caption{\small{Boxplots of the weighted F1 scores by random forest  (RF) and Bayesian neural network (Bayesian NN) models for pitted scab disease using environmental predictors (Soil: Soil chemistry, DS: Microbial population density in soil).}}
\label{14-RF-BNN-Env}
\end{figure}

\begin{figure}[h]
\centering
\includegraphics[scale=0.5]{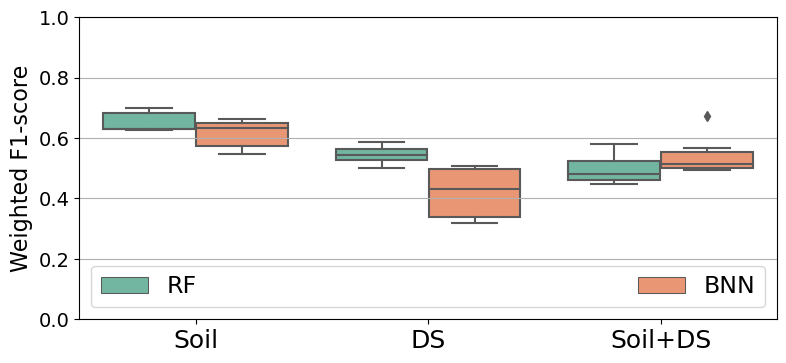}
\caption[Boxplots of the weighted F1 scores  for yield by plant using environmental predictors]{Boxplots of the weighted F1 scores by Random Forest (RF) and Bayesian Neural Network (BNN) models for yield by plant using environmental predictors (Soil: Soil chemistry, DS: Microbial population density in soil).}
\label{S-RF-BNN-Env-Yield-Plant}
\end{figure}
\begin{figure}[h]
\centering
\includegraphics[scale=0.5]{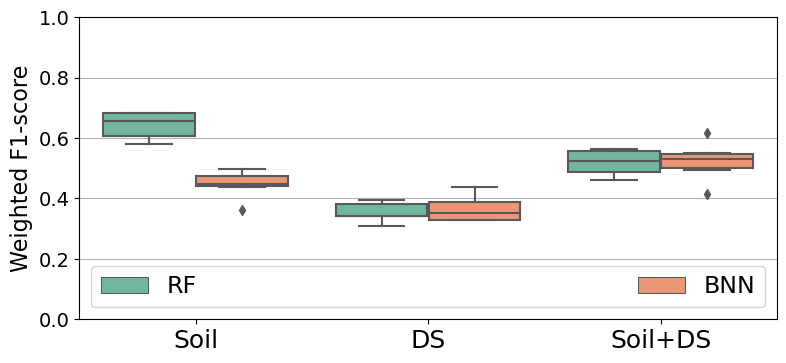}
\caption[Boxplots of the weighted F1 scores  for yield by meter using environmental predictors]{Boxplots of the weighted F1 scores by Random Forest (RF) and Bayesian Neural Network (BNN) models for yield by meter using environmental predictors (Soil: Soil chemistry, DS: Microbial population density in soil).}
\label{S-RF-BNN-Env-Yield-Meter}
\end{figure}
\begin{figure}[h]
\centering
\includegraphics[scale=0.5]{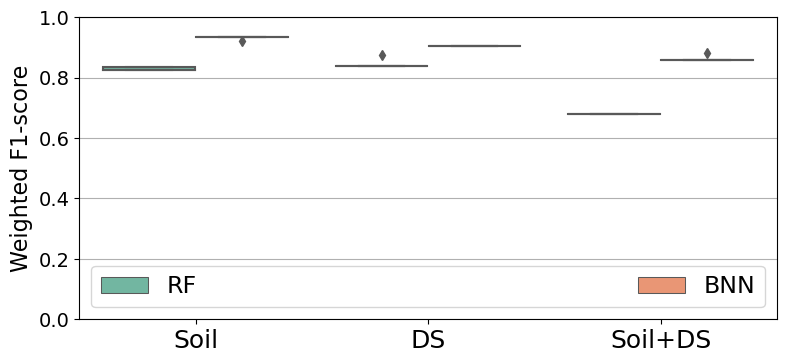}
\caption[Boxplots of the weighted F1 scores for black scurf disease  using environmental predictors]{Boxplots of the weighted F1 scores by Random Forest (RF) and Bayesian Neural Network (BNN) models for black scurf disease using environmental predictors (Soil: Soil chemistry, DS: Microbial population density in soil).}
\label{S-RF-BNN-Env-Black}
\end{figure}
\begin{figure}[h]
\centering
\includegraphics[scale=0.5]{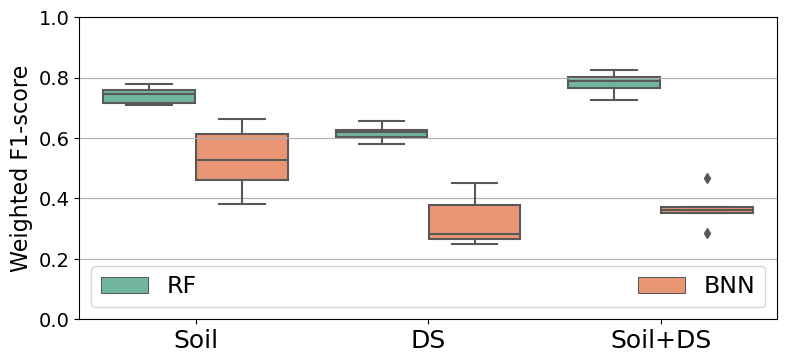}
\caption[Boxplots of the weighted F1 scores for scab disease using environmental predictors]{Boxplots of the weighted F1 scores by Random Forest (RF) and Bayesian Neural Network (BNN) models for scab disease outcome using environmental predictors (Soil: Soil chemistry, DS: Microbial population density in soil).}
\label{S-RF-BNN-Env-Scab}
\end{figure}
\begin{figure}[h]
\centering
\includegraphics[scale=0.5]{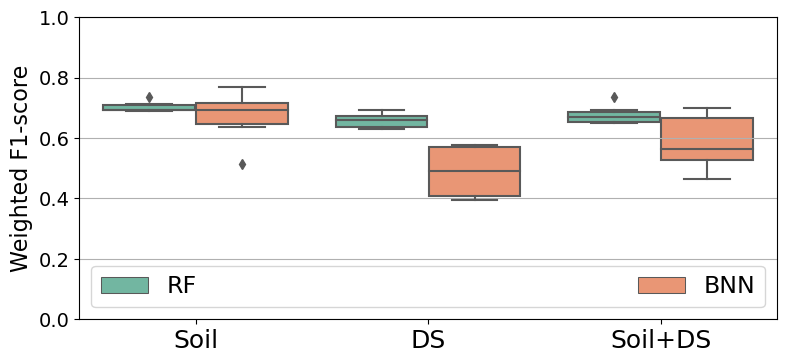}
\caption[Boxplots of the weighted F1 scores for superficial scab (Scabsuper) disease  using environmental predictors]{Boxplots of the weighted F1 scores by Random Forest (RF) and Bayesian Neural Network (BNN) models for superficial scab (Scabsuper) disease using environmental predictors (Soil: Soil chemistry, DS: Microbial population density in soil).}
\label{S-RF-BNN-Env-Scabsuper}
\end{figure}

\begin{figure}[ht!]
\centering
\includegraphics[scale=0.4]{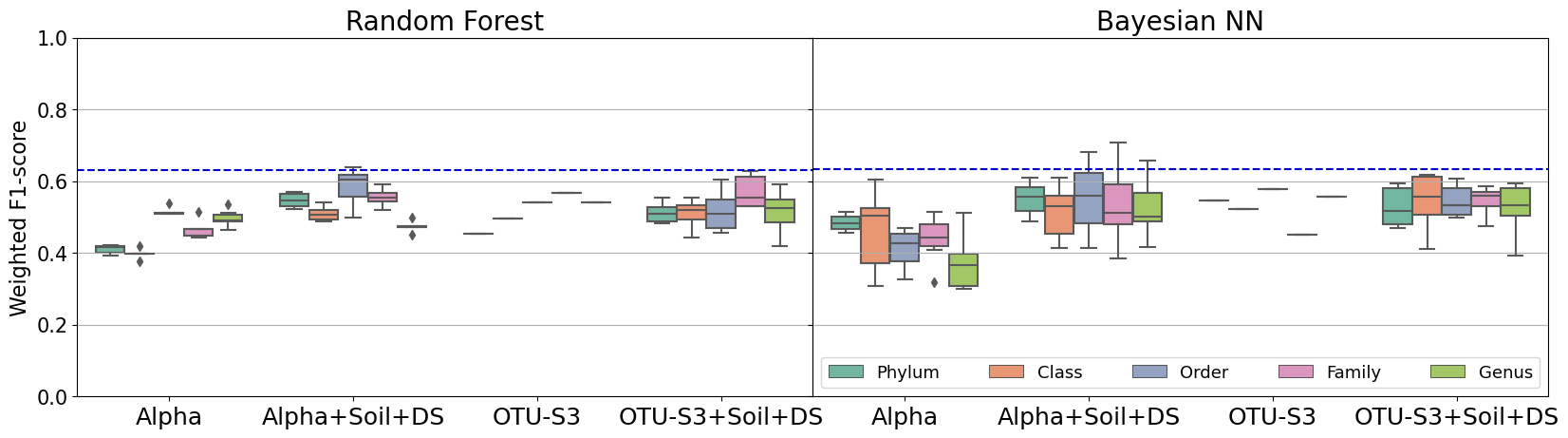}
\caption[Boxplots of the weighted F1 scores for yield per plant based on Alpha, Alpha+Soil+DS, Soil+DS, OTU-S3, and OTU-S3+Soil+DS predictors]{Boxplots with the weighted F1 scores (y-axis) by Random Forest and Bayesian Neural Network (NN) models  for yield per plant.  The models including both types of predictors outperform other models, yet models including OTU data alone (OTU-S3) are comparable which suggests that the microbial information indeed contains signal to predict the disease outcome on its own. However, OTU data is more expensive to collect, and perhaps not necessary, given that the model without OTU data (Soil) performs just as accurately (blue dashed line).}
\label{S-RF-BNN-Com-Yield-Plant}
\end{figure}
\begin{figure}[ht!]
\centering
\includegraphics[scale=0.4]{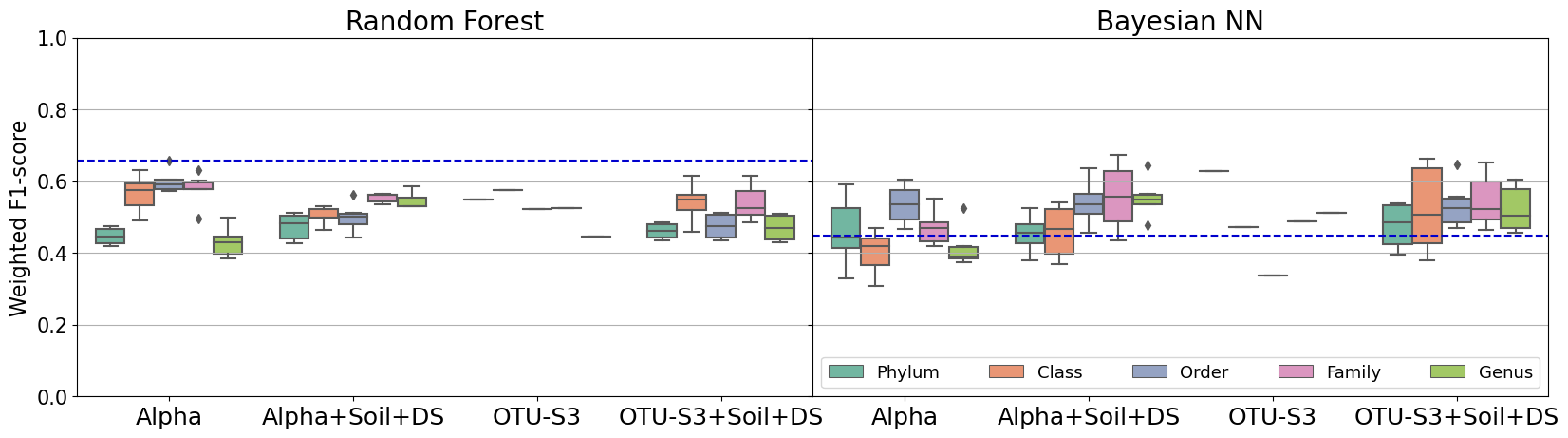}
\caption[Boxplots of the weighted F1 scores for yield per meter based on Alpha, Alpha+Soil+DS, Soil+DS, OTU-S3, and OTU-S3+Soil+DS predictors]{Boxplots with the weighted F1 scores (y-axis) by Random Forest and Bayesian Neural Network (NN) models for yield per meter.  The models including both types of predictors outperform other models, yet models including OTU data alone (OTU-S3) are comparable which suggests that the microbial information indeed contains signal to predict the disease outcome on its own. However, OTU data is more expensive to collect, and perhaps not necessary, given that the model without OTU data (Soil) performs just as accurately (blue dashed line).}
\label{S-RF-BNN-Com-Yield-Meter}
\end{figure}
\begin{figure}[ht!]
\centering
\includegraphics[scale=0.4]{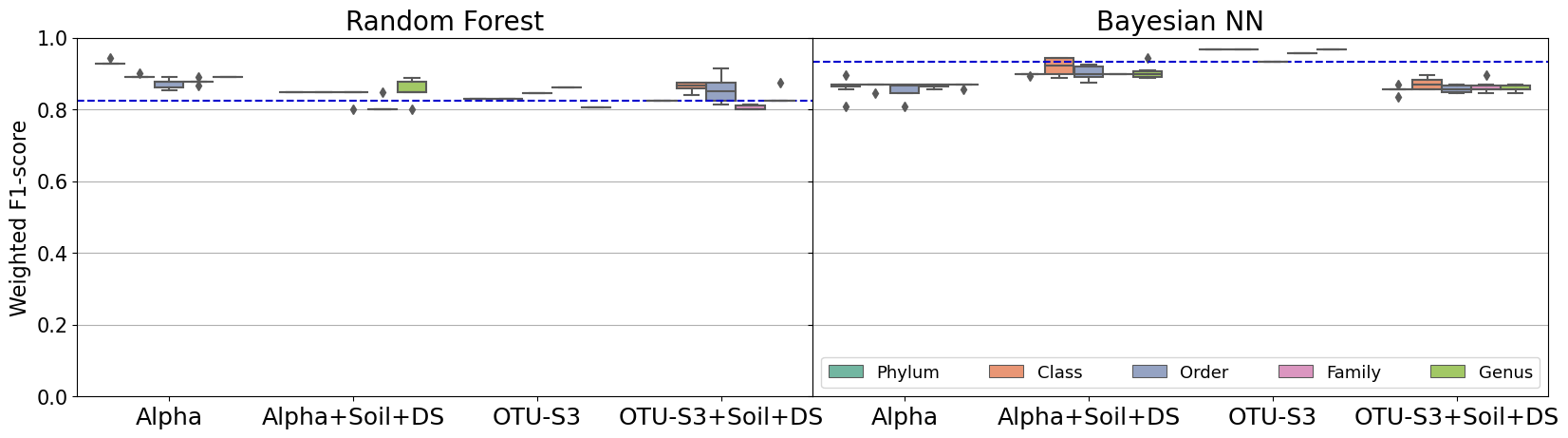}
\caption[Boxplots of the weighted F1 scores for black scurf disease based on Alpha, Alpha+Soil+DS, Soil+DS, OTU-S3, and OTU-S3+Soil+DS predictors]{Boxplots with the weighted F1 scores (y-axis) by Random Forest and Bayesian Neural Network (NN) models for black scurf disease.  The models including both types of predictors outperform other models, yet models including OTU data alone (OTU-S3) are comparable which suggests that the microbial information indeed contains signal to predict the disease outcome on its own. However, OTU data is more expensive to collect, and perhaps not necessary, given that the model without OTU data (Soil) performs just as accurately (blue dashed line).}
\label{S-RF-BNN-Com-Black}
\end{figure}
\begin{figure}[ht!]
\centering
\includegraphics[scale=0.4]{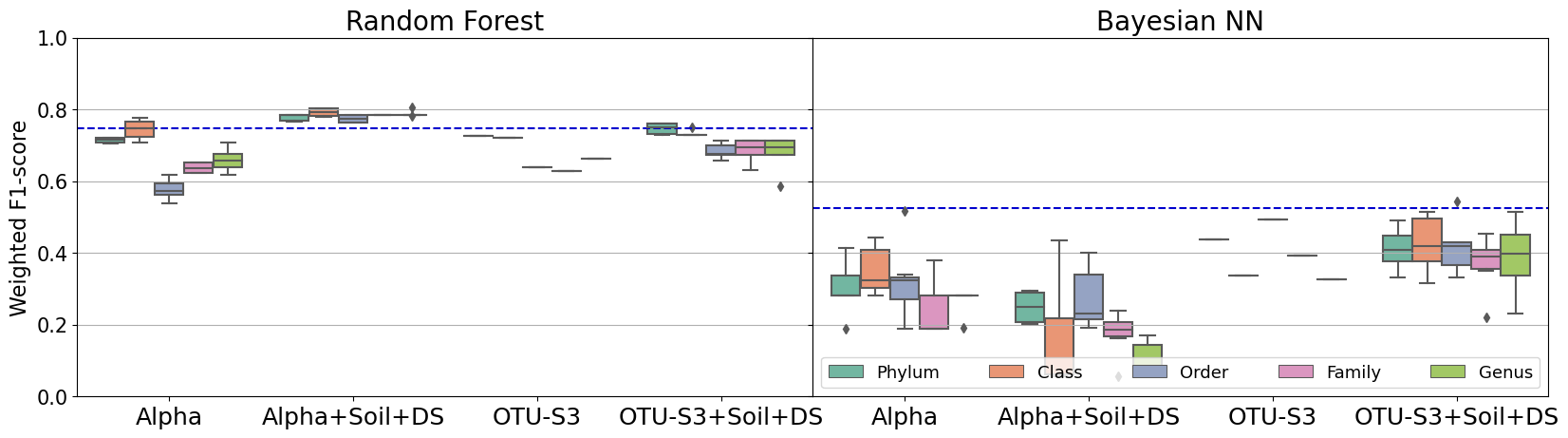}
\caption[Boxplots of the weighted F1 scores for scab disease based on Alpha, Alpha+Soil+DS, Soil+DS, OTU-S3, and OTU-S3+Soil+DS predictors]{Boxplots with the weighted F1 scores (y-axis) by Random Forest and Bayesian Neural Network (NN) models for scab disease.  The models including both types of predictors outperform other models, yet models including OTU data alone (OTU-S3) are comparable which suggests that the microbial information indeed contains signal to predict the disease outcome on its own. However, OTU data is more expensive to collect, and perhaps not necessary, given that the model without OTU data (Soil) performs just as accurately (blue dashed line).}
\label{S-RF-BNN-Com-Scab}
\end{figure}
\begin{figure}[ht!]
\centering
\includegraphics[scale=0.4]{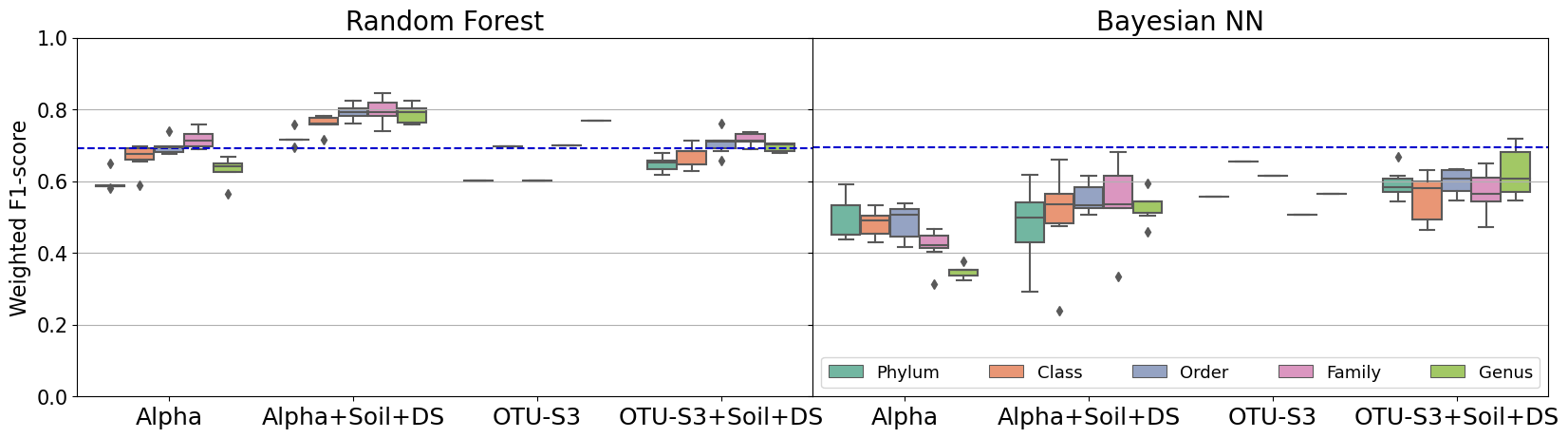}
\caption[Boxplots of the weighted F1 scores for superficial scab (Scabsuper) disease based on Alpha, Alpha+Soil+DS, Soil+DS, OTU-S3, and OTU-S3+Soil+DS predictors]{Boxplots with the weighted F1 scores (y-axis) by Random Forest and Bayesian Neural Network (NN) models for superficial scab (Scabsuper) disease.  The models including both types of predictors outperform other models, yet models including OTU data alone (OTU-S3) are comparable which suggests that the microbial information indeed contains signal to predict the disease outcome on its own. However, OTU data is more expensive to collect, and perhaps not necessary, given that the model without OTU data (Soil) performs just as accurately (blue dashed line).}
\label{S-RF-BNN-Com-Scabsuper}
\end{figure}

\end{document}